\documentclass{article}

\PassOptionsToPackage{numbers}{natbib}
\usepackage[preprint]{neurips_2026}

\usepackage[utf8]{inputenc} % allow utf-8 input
\usepackage[T1]{fontenc}    % use 8-bit T1 fonts
\usepackage[hidelinks]{hyperref}       % hyperlinks
\usepackage{url}            % simple URL typesetting
\usepackage{booktabs}       % professional-quality tables
\usepackage{amsfonts}       % blackboard math symbols
\usepackage{nicefrac}       % compact symbols for 1/2, etc.
\usepackage{microtype}      % microtypography
\usepackage{xcolor}         % colors

\usepackage{amsmath}
\usepackage{amssymb}
\usepackage{mathtools}
\usepackage{amsthm}
\usepackage{graphicx} 
\usepackage{cancel} 
\usepackage{multirow}
\usepackage{subcaption}
\usepackage[capitalise]{cleveref}
\usepackage[percent]{overpic}
\usepackage{tikz}
\usepackage{placeins}
\usepackage{array}
\usepackage{enumitem}

\crefname{appendix}{App.}{Apps.}
\Crefname{appendix}{App.}{Apps.}

\title{Mirror Learning}

\author{%
  \begin{tabular}{c}
    \bfseries
    Yunpeng Liu\textsuperscript{\rm 1,\rm 2} \quad
    Matthew Niedoba\textsuperscript{\rm 1,\rm 2} \quad
    Oluwanifemi A. Adekanye\textsuperscript{*,\rm 1} \quad
    Jason Yoo\textsuperscript{*,\rm 1} \\
    \bfseries
    Yingchen He\textsuperscript{\rm 1} \quad
    Berend Zwartsenberg\textsuperscript{\rm 2} \quad
    Frank Wood\textsuperscript{\rm 1,\rm 2,\rm 3} \\
    \normalfont
    \textsuperscript{\rm 1}University of British Columbia \quad
    \textsuperscript{\rm 2}Inverted AI \quad
    \textsuperscript{\rm 3}Amii
  \end{tabular}
}

\begin{document}

\maketitle

\begingroup
\renewcommand{\thefootnote}{*}
\footnotetext{Equal contribution.}
\endgroup
\begin{abstract}
We investigate imitation learning through the lens of third-person observation and propose a framework for mirror learning: acquiring actionable policies from passive observation. While behavior cloning (BC) excels under dense, well-aligned first-person data, it fundamentally fails to leverage the rich observational signals arising from third-person demonstrations that humans and animals routinely exploit.  We introduce a method that composes (i) a learned perspective transformation that places learners in  demonstrators' shoes using a fine-tuned video diffusion model and (ii) an inverse dynamics model that infers action trajectories in the learners' control space. This enables the synthesis of mirror data, pseudo first-person expert data generated from third-person observations of demonstrator behavior. Empirically, we show that mirror data alone can train effective policies, and that augmenting first-person BC training with mirror data further improves downstream policy performance.  Our results suggest that modern generative world models implicitly encode sufficient structure to enable a scalable and safe alternative to teleoperation-heavy data collection.

\end{abstract}

\section{Introduction}

% Behavior cloning (BC) \citep{pomerleau1989alvinn,bojarski2016end}—which we define as supervised learning from first-person observations and interactions—is conceptually straightforward and performs well when demonstrations are dense, diverse, low-noise, and near-optimal, with sufficient state coverage. 
% Recent industrial successes, particularly in autonomous driving \citep{bojarski2016end,wayve2023foundation}, have reinforced BC as a dominant paradigm, further amplified by increasing computational scale \citep{brown2020language}. 
% Indeed, large language model alignment can be viewed as a form of behavior cloning over human demonstrations \citep{ouyang2022training}.

Behavior cloning (BC) \citep{pomerleau1989alvinn,bojarski2016end} is supervised learning of policies from first-person observation-action trajectories. It is conceptually straightforward and performs well when demonstrations are dense, diverse, low-noise, and near-optimal with sufficient state coverage. Recent industrial successes, particularly in autonomous driving \citep{bojarski2016end,gao2025survey}, have reinforced BC as a dominant paradigm, further amplified by increasing computational scale \citep{brown2020language}. 
% wayve2023foundation is not found so replaced with a survey paper 
% Indeed, large language model alignment can be viewed as a form of behavior cloning over human demonstrations \citep{ouyang2022training}.
Why behavior cloning? First, supervised learning is robust and scalable \citep{lecun2015deep}. 
Second, reinforcement learning (RL), the principal alternative, suffers from sample inefficiency \citep{mnih2015human} and requires reward specification, which can be ill-posed in real-world settings \citep{ng1999policy}. 
Moreover, RL training is inherently first-person and experiential, raising safety and feasibility concerns in physical systems \citep{garcia2015comprehensive}.

% \begin{figure}[bt]
%     \centering
%     \includegraphics[width=\linewidth]{figures/multi_pngs/005_012.png}
%     \caption{The learner}
%     \label{fig:model_diagram}
% \end{figure}

\begin{figure*}[th]
    \centering
    \begin{tikzpicture}
        \node[anchor=south west, inner sep=0] (img) at (0,0)
            {\includegraphics[width=0.95\textwidth]{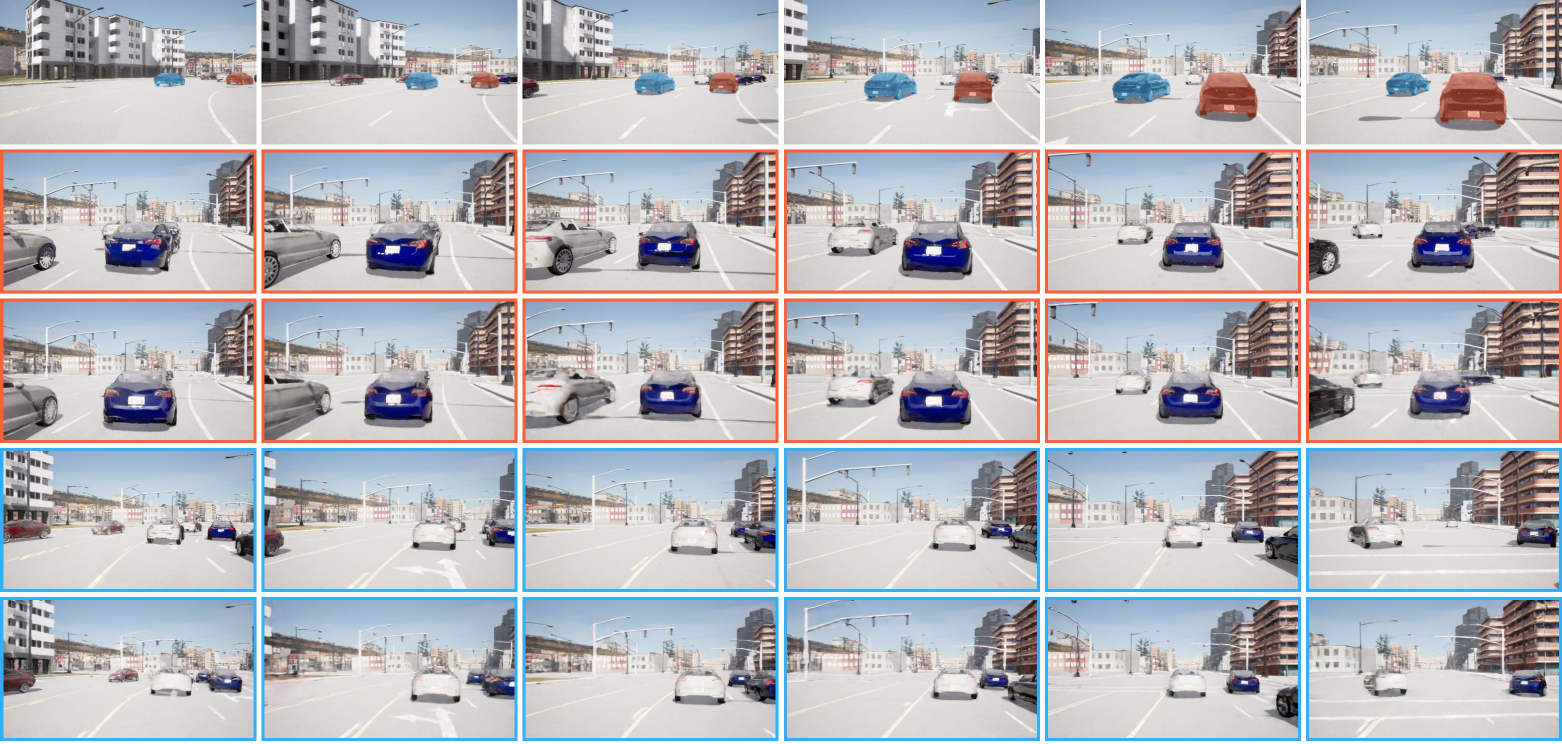}};
        \node[anchor=south, font=, font=\fontsize{5}{6}\selectfont, rotate=90, align=center] at (-0.05, 5.70) {Observation\\(masks overlaid)};
        \node[anchor=south, font=, font=\fontsize{5}{6}\selectfont, rotate=90, align=center] at (-0.05, 4.430) {Ground Truth};
        \node[anchor=south, font=, font=\fontsize{5}{6}\selectfont, rotate=90, align=center] at (-0.05, 3.20) {Mirrored};
        \node[anchor=south, font=, font=\fontsize{5}{6}\selectfont, rotate=90, align=center] at (-0.05, 1.925) {Ground Truth};
        \node[anchor=south, font=, font=\fontsize{5}{6}\selectfont, rotate=90, align=center] at (-0.05, 0.675) {Mirrored};
    \end{tikzpicture}
    \caption{Mirror video model: test environment video (CARLA town excluded from training).  Top row: learner egocentric view with two demonstrator attention masks overlaid in orange and blue. For each demonstrator, we show the ground-truth egocentric video and one sample from our mirror video model. These mirror video model results demonstrate target-specific egocentric generation and temporal consistency that leverages context from throughout the learner video. Specifically the tan building on the right is not visible to the learner until the fifth shown frame, yet the model begins rendering it in the first mirrored frame.  Car types, colors, light states, and lane markings are faithfully, though imperfectly, rendered.  We show a single sample here because the learner is  following both demonstrators and there is high overlap in their viewing frustums over time and therefore little variability between samples.  Examples with greater sample variability arising from the learner and demonstrator crossing or opposing each other in motion appear in~\cref{sec:appendix:harder-test-town}.}
    \label{fig:mvm-carla-test-town-first-page-banner}
\end{figure*}
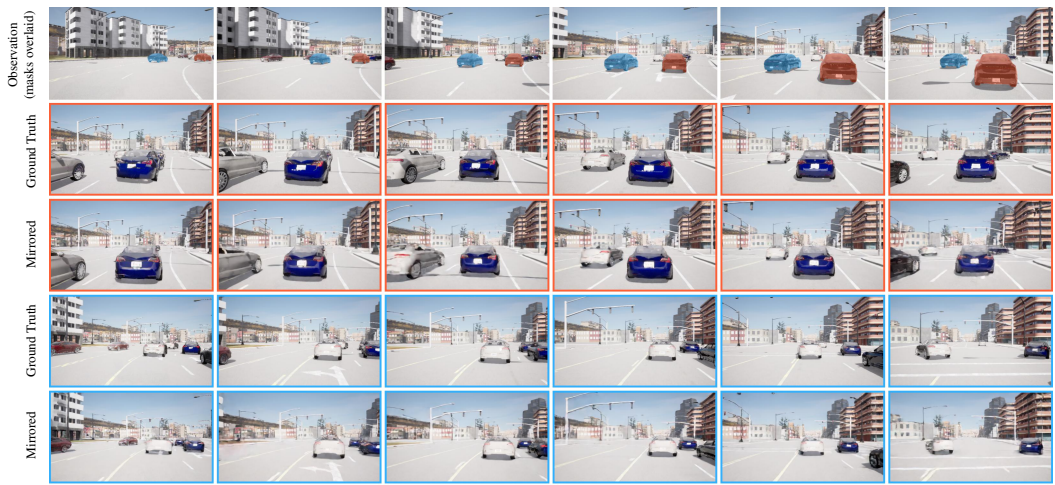

In contrast, humans and animals learn from observing others. 
Extensive evidence from cognitive science and neuroscience shows that demonstration  accelerates skill acquisition \citep{bandura1977social,heyes2011automatic}. 
Observational learning provides strong priors, particularly in tasks with high failure costs, such as locomotion or tool use \citep{winstein1991knowledge}. 
However, this creates a fundamental challenge: how can agents learn from demonstrations when only their own embodiment is available for interaction?
Neuroscience offers partial insights through the discovery of mirror neurons, which respond during both action execution and observation \citep{rizzolatti1996premotor,iacoboni1999cortical}. 
These findings suggest a shared representation between perception and action, though their computational role remains debated \citep{hickok2009eight}. 
Related phenomena, such as vicarious responses to pain and touch, further support the existence of cross-agent representational alignment \citep{keysers2010somatosensation}.

In machine learning, practical solutions have focused on scaling first-person data collection via teleoperation or simulation \citep{levine2016end}. 
However, teleoperation is often infeasible for complex embodiments or tasks. %, and mapping between demonstrator and learner embodiments introduces significant challenges \citep{argall2009survey}. 
  %Inferring latent states, including proprioception, from demonstrations remains an open problem.
Prior work on third-person imitation learning attempts to address this gap by learning domain-invariant representations \citep{sermanet2018time} or performing explicit viewpoint transformations \citep{liu2018imitation}. 
These approaches often rely on known camera geometry or constrained environments.
In principle, combining perspective transformation with inverse dynamics models (IDMs) allows recovery of action labels from observation \citep{pathak2017curiosity}. 
IDMs have been applied to infer missing actions in first-person data \citep{agrawal2016learning}, but their application to cross-perspective learning remains limited.

Recent advances in generative video modeling suggest a new direction. 
Video diffusion and world models learn rich latent representations of physical environments \citep{ho2022video,brooks2024video}. 
These models enable planning \citep{hafner2019learning}, inpainting \citep{lugmayr2022repaint}, and representation learning~\citep{velez2025image}, indicating strong implicit world modeling capabilities.
 Motivated by this, we develop a novel approach to adapting pretrained video diffusion to perform demonstrator-to-learner perspective transformation, resulting in what we call a {\em mirror video model}.  
%our framework fine-tunes such models to perform this transformation without requiring explicit camera parameters, pose estimation, or models of the world per se.
%Concretely, the model takes the form of a video diffusion architecture \citep{azzolini2025cosmos} with attention-based conditioning over instance-segmented demonstrator identities.
% Our approach is instantiated using a standard video diffusion architecture with attention-based conditioning over segmented demonstrator identities. 
% Empirically, we find that the model produces high-fidelity first-person renderings and when those are combined with an IDM, this in turn enables the synthesis of pseudo-first-person trajectories from passive observation.   We also find not only that augmenting behavior cloning with mirror data improves policy performance but that effective policies can be learned from only mirror data.

Moreover, we find that our mirror video model enables {\em mirror learning} which we define to be a method for policy improvement or learning directly from demonstrator observations.
%The central finding of this paper is that  powerful pre-trained video generative models have internal representations of the world that are sufficiently rich to enable mirror learning.  %The first step towards this, the central methodological contribution of this paper, was the development of a concrete recipe for adapting and fine-tuning pre-trained video generative models to enable demonstrator-to-learner perspective transformation.  This result, which we call a {\em mirror video model}, was achieved with only minor architectural modifications to an existing pre-trained video generative model and a straightforward fine-tuning task. % It stands to reason that similar or better results can probably be achieved with different video diffusion models and modifications thereto.
%Qualitative and some quantitative evaluation of this mirror video model was performed (e.g. \cref{fig:mvm-carla-test-town-first-page-banner, fig:may-dataset-oneshot, fig:plaicraft_val_main_paper} and many more in the appendix). 
Mirror learning requires an inverse dynamics model so in one domain we develop one, compose it with our mirror video model and show that learning a policy from mirror data alone is possible.
%Our results demonstrate that passive observational learning can be operationalized in modern ML systems. 
The name mirror learning reflects its conceptual connection to mirror neuron systems and suggests that it can enable safe, scalable learning without direct interaction, offering a promising alternative to traditional imitation learning pipelines.

\begin{figure}[th]
    \centering
    \includegraphics[width=1.0\linewidth]{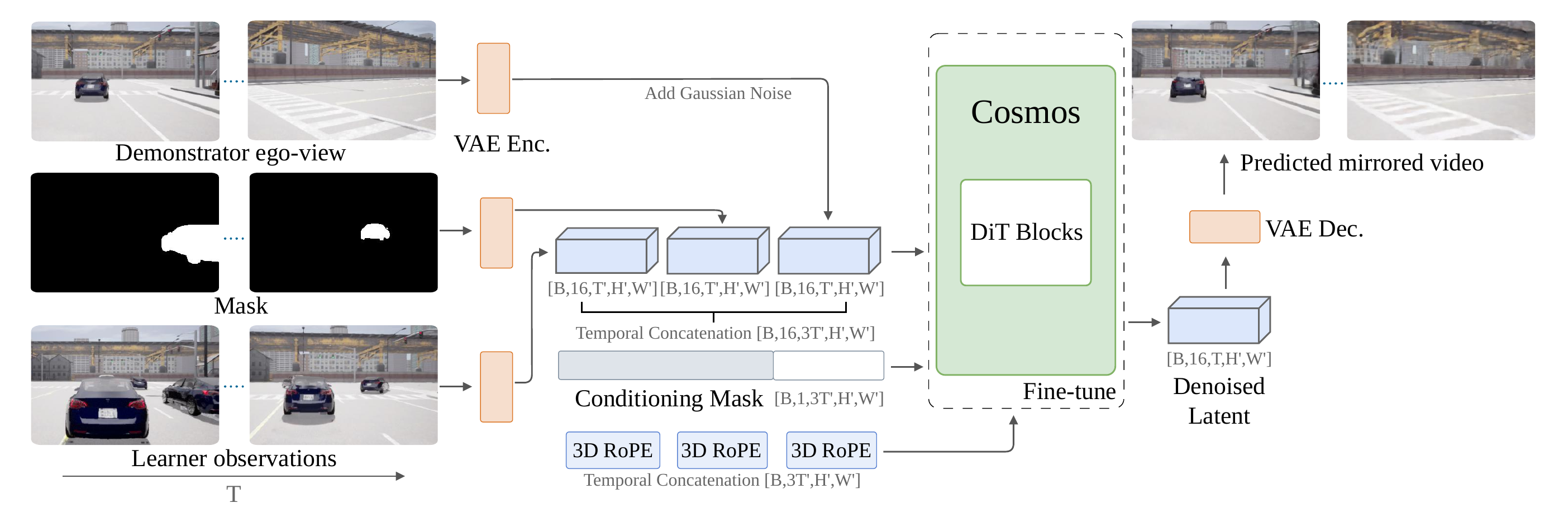}
\caption{\textbf{Architecture of our mirror video model (MVM).}
Given a learner ego-view video and a binary attention-mask video that identifies a demonstrator, the MVM predicts the corresponding demonstrator ego-view video. During training, we fine-tune \textsc{Cosmos-Predict 2.5-2B} to denoise a noised latent encoding of the target demonstrator ego-view conditioned on the learner video and the demonstrator mask. The temporally aligned learner video, demonstrator mask, and noised target videos are encoded with a spatio-temporal VAE encoder, and the resulting latents are concatenated along the temporal dimension. To adapt the pretrained video backbone to this multi-stream input, we assign the same 3D RoPE to tokens with aligned spatiotemporal positions across the three encoded videos and use a conditioning mask to distinguish observed from denoised tokens. The denoised latent is then decoded to produce the mirrored video, i.e.~the inferred first-person video of the demonstrator.}
    \label{fig:model_diagram}
\end{figure}

\section{Methods}
% The first part of mirror learning requires converting observations of a demonstrator agent into perceptual data in the learner's own observation space -- putting the learner ``in the shoes'' of the demonstrator.  The second part of mirror learning requires synthesizing actions that correspond to  demonstrators' observed behavior.  The combination, a perceptual observation sequence and a corresponding sequence of actions, if sufficiently noise-free, can then be used by a learner to behavior clone demonstrators.  

The first part of mirror learning requires converting observations of a demonstrator agent into perceptual data in the learner's own observation space -- putting the learner ``in the shoes'' of the demonstrator. The second part requires synthesizing actions corresponding to the demonstrator's observed behavior. If the combination of the perceptual observation sequence and corresponding inferred action sequence is sufficiently noise-free, they can be used by a learner to behavior clone demonstrators. While there are many possible approaches to solving this problem, we specifically restrict ourselves to those that require minimal external supervision and leverage pre-trained artifacts. %that could, in theory, be evolutionarily conserved or learned directly by the learner itself using only first person observations.  

% The first aspect of this problem is identifying which of potentially many demonstrators to attend to.  In realistic settings there will be a number of potential demonstrators within view.  We use SAM3~\citep{carion2025sam3segmentconcepts} with a text prompt to segment the relevant object class across the video sequence, and then manually identify the demonstrator among the resulting candidate tracks using the temporally consistent object IDs produced by SAM3. The specific demonstrator attention representation we employ is a binary ``attention'' mask video which consists of the instance segmentation of an identified demonstrator in the learner video.  

% \jy{Stylistically, what do you think about having a shorter "intro" to the section and diving deeper into each component in subsections? For example, the paragraph below should be folded into section 2.1, and second paragraph to section 2.2.}

A subproblem of this is to which of potentially many demonstrators to attend.  In realistic settings there will be a number of potential demonstrators within view. We use SAM3~\citep{carion2025sam3segmentconcepts} with a text prompt to segment the relevant object class across the video, and then manually select the demonstrator from the resulting candidate tracks using SAM3's temporally consistent object IDs. The specific demonstrator attention representation we employ is a binary ``attention'' mask video which consists of the instance segmentation of an identified demonstrator in the learner video.  

Let us assume for the moment that we have a sufficiently capable mechanism for inferring the demonstrator first person observation from learner observations.  Armed with such a capability the last aspect of the mirror learning problem is to infer the actions the demonstrator took while being observed.  An inverse dynamics model (IDM) that is specific to both the application and the learner's action space can be learned by the learner itself using local data, requiring only observation pairs and the action that the learner took to effect those transitions.

% Note that learning such a model only requires data local to the learner, in particular we use learner ego experience to train an IDM.

%while the specific choices of domain also allow us to experiment with policy learning
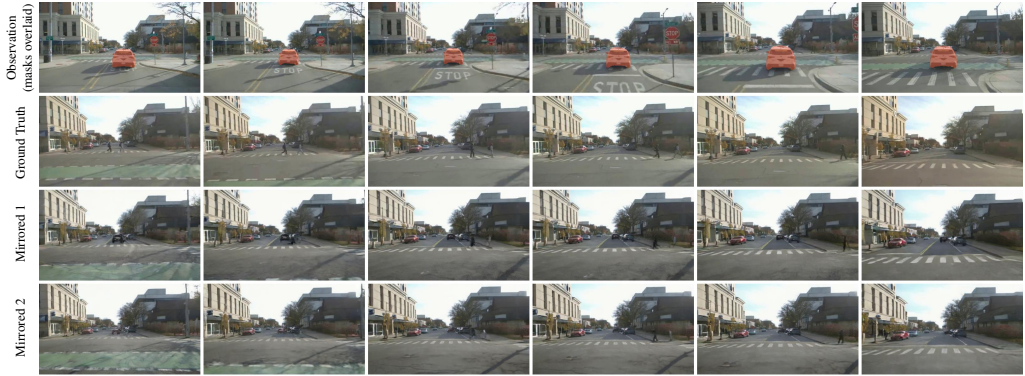
\begin{figure*}[t]
    \centering
    \begin{tikzpicture}
        \node[anchor=south west, inner sep=0] (img) at (0,0)
            {\includegraphics[width=0.93\textwidth]{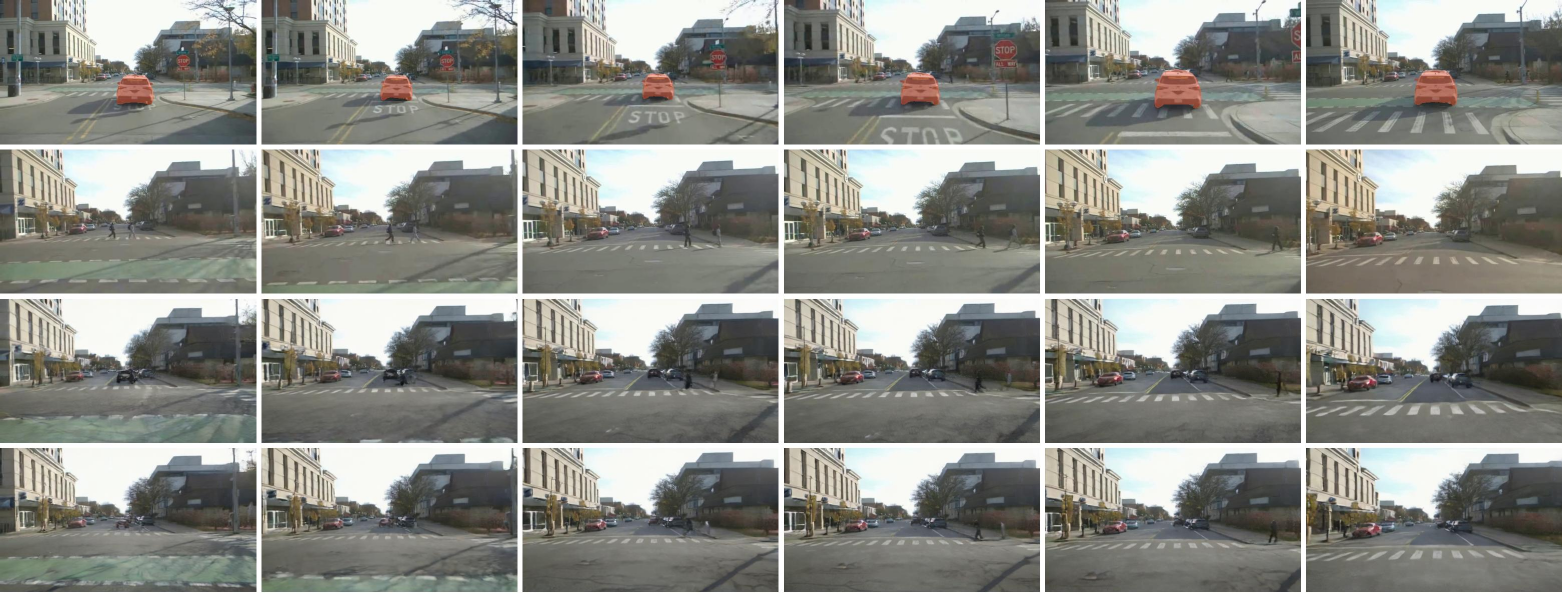}};
        \node[anchor=center, font=\fontsize{5}{6}\selectfont,align=center, rotate=90] at (-0.25, 4.35) {Observation\\(masks overlaid)};
        \node[anchor=center, font=\fontsize{5}{6}\selectfont, align=center, rotate=90] at (-0.25, 3.10) {Ground Truth};
        \node[anchor=center, font=\fontsize{5}{6}\selectfont, align=center, rotate=90] at (-0.25, 1.90) {Mirrored 1};
        \node[anchor=center, font=\fontsize{5}{6}\selectfont, align=center, rotate=90] at (-0.25, 0.65) {Mirrored 2};
    \end{tikzpicture}
    \caption{Zero-shot mirror video model predictions on real-world data (MVM trained only on CARLA synthetic data w/o pedestrians).  Ground truth extracted from  data \citep{li2024multiagent} in which a ``learner'' May robotaxi follows and observes a ``demonstrator'' May robotaxi.}% for which ground truth is thus known.  }
    \label{fig:may-dataset-oneshot}
\end{figure*}

% Frank wrote this
% What remains is the problem of view transformation, namely the task of ``putting the learner in the demonstrator's shoes.''  There are many elegant, infer the latent structure of the world, approaches to doing this, each coming with a variety of trade-offs -- inverse graphics~\citep{yao20183d}, NERFs~\citep{mildenhall2021nerf} and Gaussian splatting~\citep{kerbl20233d}. They all require knowing or inferring at least the relative camera pose of the demonstrator and sometimes the latent scene geometry itself~\citep{Fu_2024_CVPR}.  All are sensitive to scene coverage in the sense of needing to extrapolate geometry beyond that which is directly observed by the learner~\citep{shih2024extranerf}.  

What remains is the core problem of view transformation, namely the task of ``putting the learner in the demonstrator's shoes.'' We address this by fine-tuning a video diffusion model without assuming known relative camera pose between the learner and demonstrator, or explicit scene geometry. This avoids explicitly extrapolating latent geometry beyond the scene coverage directly observed by the learner, a common challenge for geometry-based novel-view synthesis methods~\citep{Fu_2024_CVPR, shih2024extranerf}.

\subsection{Specification}
 %We first define our perspective transformation task then describe how we modified and fine-tuned the DiT-based Cosmos video diffusion model to this task.

% to support policy learning. The core of our approach is a mirror network, which we develop by adapting a large foundation video prediction model to predict the egocentric observations of a target agent from another agent's viewpoint. This formulation is loosely inspired by the mirror neuron hypothesis, which suggests that observed behavior can be internally related to the observer's own sensorimotor representations. To emulate this idea, we formulate two coupled tasks, mirroring, which predicts a target agent's egocentric observations, and inverse dynamics, which recovers the actions associated with those observations. We first define the mirroring task in a synchronized multi-agent setting, then describe the modifications that adapt the DiT-based Cosmos video diffusion model to this task, and finally introduce the inverse dynamics model used in the downstream autonomous driving setting.

% \paragraph{Mirroring task}
% \begin{figure}[t]
%     \centering
%     \includegraphics[width=\linewidth]{figures/video_token_conditioned_dit_1105.pdf}
%     \caption{Model overview.}
%     \label{fig:model_diagram}
% \end{figure}

%We study mirroring in a synchronized multi-agent setting, where multiple embodied agents interact in the same dynamic environment and each agent observes the scene from its own egocentric viewpoint. 

\paragraph{Mirror Video Model} Let $o^{(i)}_{1:T}$ denote the ego-view video of learner $i$ with $T$ frames. Let $m^{(j \leftarrow i)}_{1:T}$ denote a sequence of binary attention masks that correspond to an instance segmentation of demonstrator $j$ in  agent $i$'s video.  Our mirror video model (MVM) learns the conditional distribution of demonstrator $j$'s ego-view video given learner $i$'s video and the mask $m^{(j \leftarrow i)}_{1:T}$ that identifies demonstrator $j$,
$
p_\theta\!\left(o^{(j)}_{1:T} \mid o^{(i)}_{1:T}, m^{(j \leftarrow i)}_{1:T}\right). \label{eq:mvm-dist}
$
% from which the mirror network generates a prediction
% \[
% \hat{o}^{(j)}_{1:T} \sim p_\theta\!\left(\cdot \mid o^{(i)}_{1:T}, m^{(j \leftarrow i)}_{1:T}\right).
% \]
%
Unlike standard conditional video generation and prediction \citep{ho2022video,harvey2022flexible,blattmann2023stable,wan2025wan}, this task requires transforming from one viewpoint to another which includes, implicitly, relative camera pose inference, scene geometry inference, and in-filling or in-painting of scene geometries or pixels corresponding to environment elements that are not directly viewed from the learners perspective, all while preserving temporal consistency of shared scene occupants and dynamics, with particular sensitivity to the motion of demonstrator.  Concretely, when the temporal and spatial overlaps between agent $i$ and  $j$'s viewing frustums are limited, the model must synthesize visually coherent content that is not directly observed and is not specified by an explicit inpainting mask. This is substantially more challenging than standard conditional video inpainting~\citep{green2024semantically,li2025diffueraser}.

% \citep{dylansinpaintingworkandothers}.

\paragraph{Inverse Dynamics Model} To turn the mirror video model outputs into a behavior cloning dataset, we additionally require the corresponding demonstrator actions.  To do this one can train an inverse dynamics model (IDM)~\citep{pathak2017curiosity,torabi2018behavioral,baker2022video} that predicts actions from egocentric observations. Given a video $o^{(i)}_{1:T}$ and action sequence $a^{(i)}_{1:T-1}$ from an agent $i$ (taken to the learner), an IDM models the conditional distribution 
$p_\phi(a^{(i)}_{1:T-1} \mid o^{(i)}_{1:T}).$
%where $a^{(i)}_{1:T-1}$ is the sequence of actions executed by agent $i$. 

\paragraph{Mirror Learning} 
Mirror learning composes an MVM with an IDM 
% Applying the IDM to a mirrored prediction $\hat{o}^{(j)}_{1:T}$ yields inferred actions
\vspace{-0.5em}
{
\setlength{\belowdisplayskip}{2pt}
\setlength{\belowdisplayshortskip}{2pt}
\begin{equation}
\hat{a}^{(i)}_{1:T-1} \sim p_\phi\!\left(\cdot \mid \hat{o}^{(j)}_{1:T}\right), \hat{o}^{(j)}_{1:T} \sim p_\theta\!\left(\cdot \mid o^{(i)}_{1:T}, m^{(j \leftarrow i)}_{1:T}\right).    
\end{equation}
}
The created mirror data, or synthetic behavior cloning trajectory $(\hat{o}^{(j)}_{1:T}, \hat{a}^{(i)}_{1:T-1})$ data, is the inferred ego-perspective video $\hat{o}^{(j)}_{1:T}$  of demonstrator $j$ and the corresponding action sequence $\hat{a}^{(i)}_{1:T-1}$ that the learner would need to have taken to generate it.  Tuples from this sequence are behavior cloning training data for learner policy learning.

% Mirror learning composes an MVM with an IDM 
% %Applying the IDM to a mirrored prediction $\hat{o}^{(j)}_{1:T}$ yields inferred actions
% \begin{equation}
% \hat{a}^{(i)}_{1:T-1} \sim p_\phi\!\left(\cdot \mid \hat{o}^{(j)}_{1:T}\right), \hat{o}^{(j)}_{1:T} \sim p_\theta\!\left(\cdot \mid o^{(i)}_{1:T}, m^{(j \leftarrow i)}_{1:T}\right).    
% \end{equation}
% We refer to these generated sequences as ``mirror data''. Tuples from this sequence are used as synthetic behavior cloning training data for the learner policy.

% 
% Frank picked this from preivous run figures/plaicraft_pdfs/val/mountain_stare.pdf
% we are doing this {figures/plaicraft_pdfs_new/val/selected_189235_snowsky_static.pdf}

\begin{figure*}[t]
    \centering
    \begin{tikzpicture}
        \node[anchor=south west, inner sep=0] (img) at (0,0)
            {\includegraphics[width=0.93\textwidth]{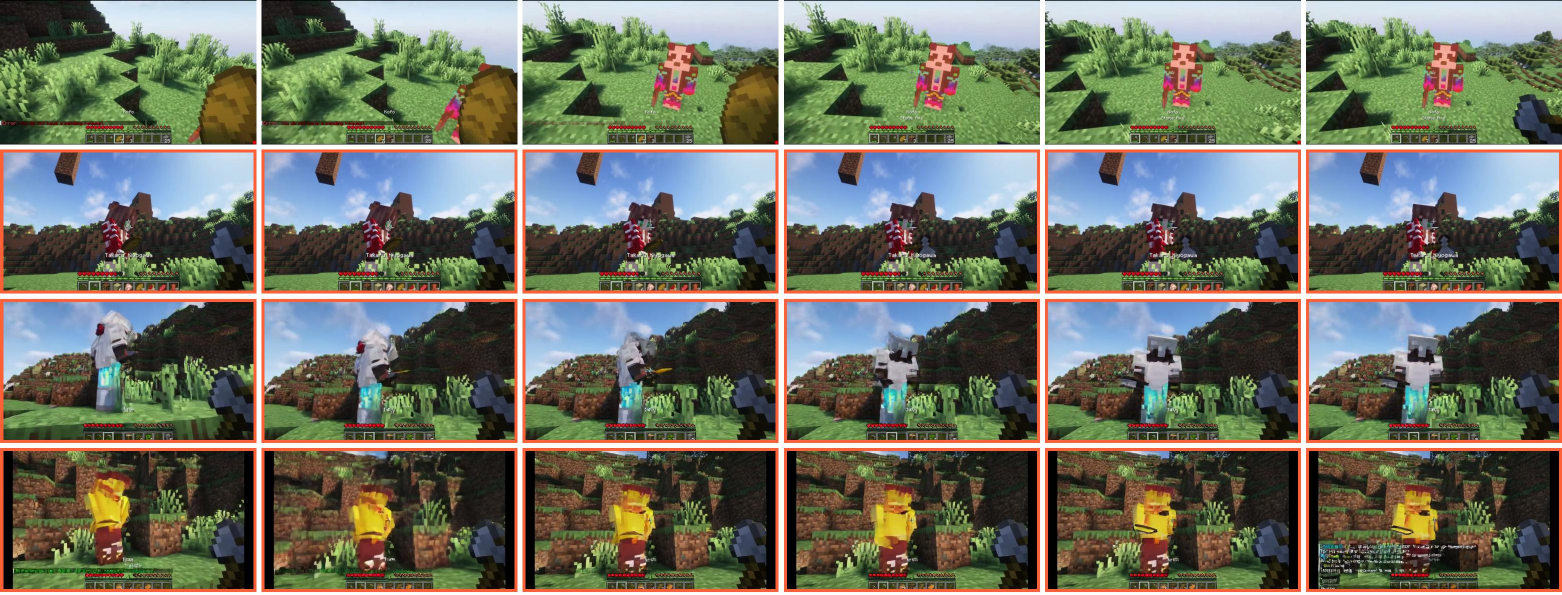}};
        \node[anchor=center, font=\fontsize{5}{6}\selectfont,align=center, rotate=90] at (-0.25, 4.35) {Learner \\(masks overlaid)};
        \node[anchor=center, font=\fontsize{5}{6}\selectfont, align=center, rotate=90] at (-0.25, 3.10) {Demonstrator};
        \node[anchor=center, font=\fontsize{5}{6}\selectfont, align=center, rotate=90] at (-0.25, 1.90) {Mirrored 1};
        \node[anchor=center, font=\fontsize{5}{6}\selectfont, align=center, rotate=90] at (-0.25, 0.65) {Mirrored 2};
    \end{tikzpicture}
    \caption{Minecraft mirror video model samples. Demonstrator instance segmentation mask indicated by orange highlighting. For the learner and demonstrator both we show time-aligned frames from their ground-truth egocentric videos.  The two samples are MVM predictions of the demonstrator view given the learner and instance segmentation mask videos.  The learner cannot see its own ``skin,'' so the MVM samples two different plausible skins.  The learner can see that the demonstrator is holding an axe; this is reflected in all samples.  The learner can see that the demonstrator is looking at and past the learner in a mountain biome; the MVM inpaints varied plausible mountain biome physical context.  Many more examples appear in~\cref{sec:appendix:plaicraft-examples}}
    \label{fig:plaicraft_val_main_paper}
\end{figure*}
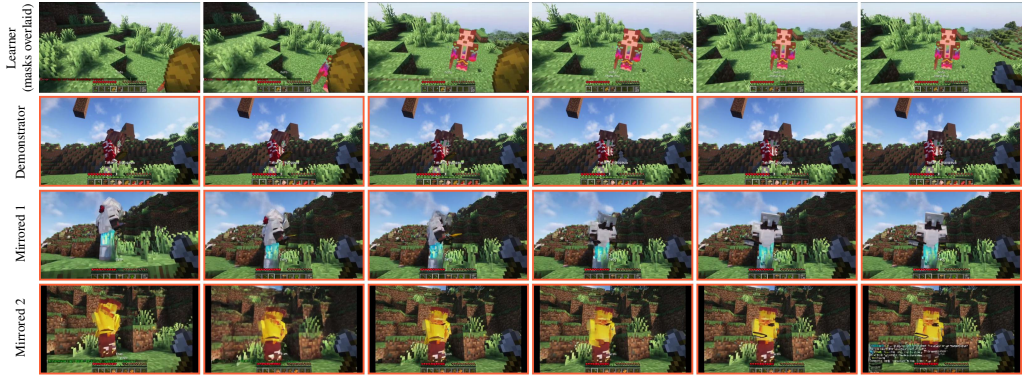

% \begin{figure*}[t]
%     \centering
%     \begin{tikzpicture}
%         \node[anchor=south west, inner sep=0] (img) at (-0.05,0)
%             {\includegraphics[width=0.925\textwidth]{figures/plaicraft_pdfs/val/mountain_stare.pdf}};

%         \node[anchor=south, font=\scriptsize, rotate=90, align=center] at (-0.125, 4.25) {Learner};
%         \node[anchor=south, font=\scriptsize, rotate=90, align=center] at (-0.125, 3.05) {Demonstrator};
%         \node[anchor=south, font=\scriptsize, rotate=90, align=center] at (-0.125, 1.85) {Sample 1};
%         \node[anchor=south, font=\scriptsize, rotate=90, align=center] at (-0.125, 0.65) {Sample 2};
%     \end{tikzpicture}
%     \caption{Minecraft mirror video model samples. Demonstrator instance segmentation mask indicated by orange highlighting. For the learner and demonstrator both we show time-aligned frames from their ground-truth egocentric videos.  The two samples are MVM predictions of the demonstrator view given the learner and instance segmentation mask videos.  The learner cannot see its own ``skin,'' so the MVM samples two different plausible skins.  The learner can see that the demonstrator is holding an axe; this is reflected in all samples.  The learner can see that the demonstrator is looking at and past the learner in a mountain biome; the MVM inpaints varied plausible mountain biome physical context.  Many more examples appear in~\cref{sec:appendix:plaicraft-examples}}
%     \label{fig:plaicraft_val_main_paper}
% \end{figure*}

\subsection{Implementation}

\paragraph{Mirror Video Model}
Foundation video prediction models are trained to generate future video frames conditioned on an observed video prefix and optional text input, i.e.~$p\!\left(o_{t+1:T} \mid o_{1:t}, c_{\text{text}}\right)$,
where $o_{1:t}$ denotes an observed video prefix and $c_{\text{text}}$ is text conditioning.  Such models implicitly continue videos from the viewpoint of the provided context.   %to the mirror video model task \cref{eq:mvm-dist}. 
%A strong video prediction backbone is particularly well suited for this setting because successful mirroring requires modeling scene dynamics, agent interactions, and temporal coherence over extended horizons. Our design aims to preserve the strengths of the pretrained model while adapting it to condition on agent $i$'s observations and an instance mask of agent $j$ in order to recover agent $j$'s egocentric perspective. In the following, we refer to agent $i$ as the observing agent and agent $j$ as the mirror-target agent.

We adapt one such model, Cosmos-Predict2.5 (Cosmos),  through its conditioning interface by combining encodings the observing agent's ego-view video $o^{(i)}_{1:T}$ with the corresponding binary mask sequence $m^{(j \leftarrow i)}_{1:T}$. We encode both through the pre-trained Wan2.1 VAE encoder~\citep{wan2025wan} used by Cosmos, replicating the binary mask three times across color channels to match video dimension before encoding. Three encoded conditioning video streams are concatenated temporally: the encoded mask sequence, the encoded learner sequence, and the noisy representation of the corresponding demonstrator first-person video used for flow matching.
% We utilize Cosmos' conditioning mask, which identifies which latent tokens correspond to conditional inputs and which correspond to denoising prediction targets, and mark both the encoded mask and learner videos as conditional inputs
% . 
Like Cosmos we use a binary conditioning mask to distinguish observed latents from target latents, and concatenate this mask channel-wise with the input latent sequence. We mark the encoded learner video sequence and encoded mask sequence as observed inputs, while treating the noised target latent sequence as the generation target.
The last modification we make is to the 3D Rotary Position Embedding (RoPE) used in Cosmos. Since $o^{(i)}_{1:T}$, $m^{(j \leftarrow i)}_{1:T}$, and $o^{(j)}_{1:T}$ are temporally aligned, we duplicate the RoPE embeddings across the three  video embeddings  so that temporally corresponding tokens share aligned positional encodings. During inference, the target stream is initialized from noise, and the model generates $\hat{o}^{(j)}_{1:T}$ conditioned on the learner's ego video and the demonstrator-identifying attention mask sequence.

Our fine-tuning objective is the same as Cosmos' flow-matching training objective \citep[Eq.~(3)]{ali2025world}.  Cosmos training infrastructure was used without major modification.  During fine-tuning, we allow all layers of the denoising transformer to adapt while freezing the text encoder and the spatial-temporal latent encoder and decoder.

\section{Experiments}
\label{sec:experiments}
%In this section, we evaluate the effectiveness of the mirror network through the downstream performance of an end-to-end imitation learning driving policy trained with synthetic data generated by the mirror network. We also investigate the generalization capability of our mirror network on a held-out CARLA town and demonstrate its potential to support autonomous fleet deployment in a new location.

In this section we present our findings and provide answers to three main questions:  
% \begin{enumerate}
%     \item Does a pre-trained video diffusion model possess a  sufficiently rich set of features to enable mirror learning?
%     \item Can a mirror video model fine-tuned only on simulated data zero-shot real data?
%     \item Can the quality of mirror video model output be good enough to be useful in downstream behavioral cloning?
%    % \item At what data scale do video mirroring capabilities emerge?
% \end{enumerate}

\begin{enumerate}[leftmargin=2em, itemsep=0pt, topsep=2pt, parsep=0pt, partopsep=0pt]
    \item Does a pretrained video diffusion model possess a sufficiently rich set of features to enable mirror learning?
    \item Can a mirror video model fine-tuned only on simulated data zero-shot real data?
    \item Can the quality of mirror video model output be good enough to be useful in downstream behavioral cloning?
\end{enumerate}

% Video diffusion model fine-tuning is computationally demanding, particularly if entirely general purpose functionality is desired.  For this reason our answer to question (1) is proof by existence -- we show that Cosmos is capable of performing this task in two different simulated domains -- autonomous vehicle and Minecraft.  Synthetic domains allow us to both compute ground truth viewpoints and metrics and investigate both relatively constrained camera movement environments whose feature representation in Cosmos is likely well developed -- cars on streets -- versus highly unconstrained, relatively uncommon environments like open-world, multi-player Minecraft where both head and body pose and the environment itself are nearly unconstrained.  We approach answering (2) also by demonstration.  We use a mirror video model trained in simulation to one-shot a real world robotaxi dataset~\citep{li2024multiagent} where occassionally one robotaxi sees the other and so we have access to ground truth data.  In the autonomous vehicle domain we answer (3) in the affirmative by showing how mirror learning can improve downstream visual policy performance.   

Video diffusion model fine-tuning is computationally demanding, particularly if one seeks entirely general-purpose functionality. For this reason, our answer to question (1) is a proof of existence, we show that Cosmos can perform this task in two simulated domains.%, autonomous driving and Minecraft. Synthetic domains let us compute ground-truth viewpoints and  metrics while also evaluating both relatively constrained environments, such as cars on streets, whose visual features are likely well represented in Cosmos, versus highly unconstrained, relatively uncommon environments like open-world, multi-player Minecraft where both head and body pose and the environment itself are nearly unconstrained. 
  We approach answering (2) also by demonstration.  We use a mirror video model trained in simulation to zero-shot a real world robotaxi (May) dataset~\citep{li2024multiagent} where occasionally one robotaxi sees the other and so we have access to ground truth data.  In the autonomous vehicle domain we answer (3) by showing that mirror learning can improve downstream visual policy performance.  

% ---------------------------------------------------------------
% Combined float: Figure 5 (scaling plot, LEFT) + Table 1 (RIGHT)
% Packages needed: booktabs, multirow, graphicx, subcaption (all in preamble)
% ---------------------------------------------------------------
\begin{figure}[t]
  \centering
  %--- left panel: scaling plot ---
  \begin{subfigure}[b]{0.46\linewidth}
    \centering
    \includegraphics[width=\linewidth]{%
      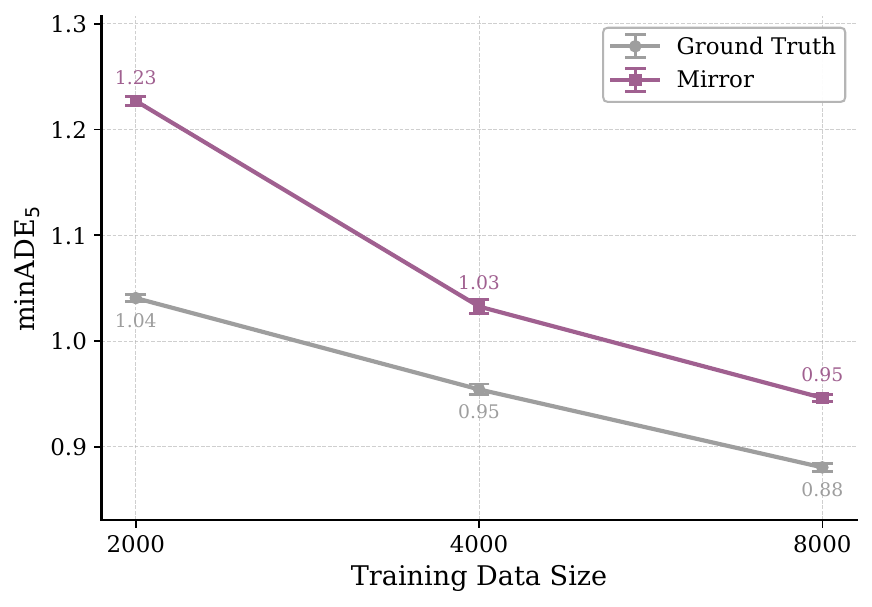}
    \caption{}
    \label{fig:mirror-vs-gt}
  \end{subfigure}
  \hfill
  %--- right panel: geographic expansion table ---
  \begin{subfigure}[b]{0.51\linewidth}
    \centering
    \small
    \setlength{\tabcolsep}{3pt}
    \begin{tabular}{lccr}
      \toprule
      GT data & Augmentation & Aug.\ ratio & minADE$_5$ \\
      \midrule
      Pretrained & -- & -- & 0.80$\pm$0.05 \\
      \midrule
      1k & \multirow{2}{*}{None}           & --        & 0.69$\pm$0.01 \\
      2k &                                 & --        & 0.64$\pm$0.00 \\
      \midrule
      1k & \multirow{2}{*}{Style transfer} & 1$\times$ & 0.68$\pm$0.00 \\
      1k &                                 & 2$\times$ & 0.70$\pm$0.01 \\
      \midrule
      1k & \multirow{4}{*}{Mirror data}    & 1$\times$ & 0.68$\pm$0.00 \\
      1k &                                 & 2$\times$ & 0.65$\pm$0.00 \\
      2k &                                 & 1$\times$ & 0.63$\pm$0.01 \\
      2k &                                 & 2$\times$ & \textbf{0.62$\pm$0.00} \\
      \bottomrule
    \end{tabular}
    \caption{}
    \label{tab:mirror-vs-other-augmentation}
  \end{subfigure}
  \caption{
    \textit{(a)} BC from mirror data alone suffices for policy learning.
    GT data is always best, but larger quantities of mirror-only data lead to equivalent performance.
    \textit{(b)} Geographic expansion into a new town. Here we fine-tune a model (first row Table~\ref{tab:mirror-as-fraction-of-total}) 
    with 1k or 2k new ground-truth samples and augment with
    style-transfer or mirror data (ratio relative to GT\@).  Mirror data outperforms style transfer and improves with higher ratios.
    Lower minADE$_5$ is better; $\pm$ is std.\ over three seeds.
    }
  \label{fig:mirror-learning-results}
\end{figure}
\subsection{Mirror Video Model}

In this section we demonstrate that a fine-tuned pre-trained video diffusion model can solve three different mirror video model tasks.  %from here
% In order to show this capacity we needed datasets in which pairs of demonstrators are visible in learners' views and the ground truth ego views are available.
To do this we need paired learner-demonstrator data in which demonstrators are visible in learners’ views and ground truth ego views are available for both agents.  Mirror learning evaluation, \cref{seC:mirror-learning}, also requires action labels. %Self-driving car companies like Waymo and Tesla are able to construct datasets like this relatively straightforwardly owing to the density of their vehicles. 
Large datasets with these features are not readily available so we constructed one in simulation using CARLA \citep{Dosovitskiy17}. %Doing so also gives us the ability to control and partition the environment in ways that are important for experimental control.  
This also lets us partition environments for controlled evaluation. Automotive data can be variable in terms of environment, but is generally highly constrained with respect to pose (most cars are relatively coplanar and their front-facing cameras are generally in a fixed pose relative to the vehicle). 
For this reason we also created a multiplayer Minecraft dataset.  Minecraft agents are humanoid in the sense that they perceive the world through time-varying, high-degree of freedom body and head poses and Minecraft environment is perceptually rich~\citep{baker2022video,he2025plaicraft}.  The May dataset contains too few suitable samples for training (see \cref{sec:may-data-details}) so we use it to evaluate the zero-shot real world performance of our automotive mirror video model and more generally to verify that this model is not fine-tuning to short-cuts \citep{geirhos2020shortcut} using, e.g., simulation artifacts.

% As an academic research lab we do not have access to such a dataset so we must use simulation; we use CARLA \citep{Dosovitskiy17} to construct such a paired dataset.
%comparable fleet-scale 

% Frank wrote As an academic research lab we do not have access to such a dataset so we must use simulation; we use CARLA \citep{Dosovitskiy17}

% As an academic research lab we do not have access to such a dataset so we must use simulation; we use CARLA \citep{Dosovitskiy17} to construct such a paired dataset.  

% and low dimensional in terms of action space (i.e. typically steering angle and instantaneous acceleration)

%The choice of automotive application here is one of convenience -- simulators exist, as do off-the-shelf fine-tunable pixels to control policies, etc.  The self-driving domain is arguably the domain where teleoperation data collection is most straightforward.  So readers should not interpret our work as being targeted to or for the autonomous vehicle domain necessarily.  Rather, readers should interpret the choice of automotive as being one of convenience with respect to the availability of simulator and policy tooling.  

%Again, en masse teleoperation data collection is also possible in this domain so, again, consider Minecraft to be a demonstration domain with good tooling for extracting ground truth and other signals necessary for establishing the efficacy of our approach.
% to here should probably be moved to the discussion or dropped entirely.  the point should be made somewhere but it is a lot of reviewer bait

\paragraph{Experimental Tasks} We collected our CARLA dataset by using a data-driven object level driving policy \citep{lioutas2025control} to generate around twenty-two thousand paired demonstrator/learner videos (8.5 seconds each at four frames per second) by ``driving'' realistic car-only traffic in multiple disjoint towns.  We selected and rendered demonstrator and learner views and corresponding instance segmentation masks (a functionality provided by CARLA) from pairs of vehicles from these scenarios that satisfied demonstrator from learner visibility thresholds.  These data were divided into training, validation, and test collections that allowed experimental control; specifically separating intra- and inter-town\footnote{CARLA towns are unreachable from each other and generally do not look the same.} physical locations.  Full details are provided in~\cref{sec:appendix:CARLA-data-collection}.  

% \cref{fig:mvm-carla-test-town-first-page-banner} shows a typical test town result, i.e.~a task from a physical environment {\em not present in the training dataset}.  Several findings are immediately made clear.   Affirmatively answering the first question, Cosmos, even after fine-tuning, clearly possesses rich enough features to solve the video mirroring task as defined.  Second, the instance segmentation mask is a sufficient signal for the model to identify and implicitly infer the pose of a demonstrator.  Finally, the fine-tuned Cosmos model has not memorized the training set, but instead has learned the video mirroring task. This is because the entire test town was not present in the training data. We provide additional qualitative and quantitative results in the Appendix section.

\cref{fig:mvm-carla-test-town-first-page-banner} shows a typical test town result, i.e.~a task from a physical environment {\em not present in the training dataset}.  These experimental results show that Cosmos, even after fine-tuning,  possesses rich enough features to solve the video mirroring task as defined.  Second, the instance segmentation mask is a sufficient signal for the model to identify and implicitly infer the pose of a demonstrator.  Finally, the fine-tuned Cosmos model has not memorized the training set, but instead has learned the video mirroring task. This is because none of the test town was present in the training data. We provide additional qualitative and quantitative results in~\cref{sec:appendix:carla_eval}.

\cref{fig:may-dataset-oneshot} shows that training on synthetic data does not destroy Cosmos' ability to mirror realistic scenes.  The May dataset is entirely withheld and Cosmos has not been trained with any paired, real-world video recordings. Moreover, the synthetic CARLA dataset contained no pedestrians.  The MVM nearly perfectly renders the environment from the demonstrator perspective, partially infers the presence and motion of pedestrians, and makes sensible inferences about occluded traffic presence and behavior (the presence and distance of any vehicle potentially in front of the demonstrator is occluded).  This zero-shot real-world performance suggests that training on synthetic data is not only sufficient to elicit mirror video model capabilities but that real world objects, behaviors, and more ``known'' to the pretrained video diffusion model may remain present in the MVM after fine-tuning. Access to paired real-world driving data would likely enable further gains, as fine-tuning the video mirroring model on even a small real dataset could reduce sim-to-real gaps arising from missing pedestrians, vehicle classes, and behaviors.

We also extracted six thousand three-second paired video clips from a multiplayer Minecraft dataset of real human gameplay~\citep{he2025plaicraft} constructed by detecting pairs of players that were visible from learner to demonstrator at various times through tracked location and viewing angle (details in~\cref{sec:appendix:plaicraft-dataset}).
Minecraft presents a substantially less constrained setting than CARLA. Participants can rapidly change their viewing direction in both pitch and yaw, and their relative poses can vary much more richly than the relative configurations of cars on roads. The visual and structural environment is also considerably more diverse. The multiplayer Minecraft dataset  contains years of continuous environment development, yielding many user-built structures and scene configurations that are assuredly not in Cosmos training data. \cref{fig:plaicraft_val_main_paper} shows that MVMs can nevertheless perform this more challenging mirroring task. Compared to CARLA, the samples exhibit substantially higher variability.  Note that samples exhibit much higher levels of variability reflecting the fact that the learner does not see or otherwise know its own appearance which varies across videos, and it does not have access to latent demonstrator state such as inventory contents, chat window status, or health.

\subsection{Mirror Learning}
\label{seC:mirror-learning}

Encouraged by the MVM results, we next evaluate whether or not behavior cloning of mirror data is effective for policy learning.  To test this we needed a task in which policy networks are available and the action space is low-enough dimensional to make inverse dynamics model learning practical on a limited computational budget.  This is another reason we chose the autonomous vehicle (AV) application; pixel to control networks with training code are now widely available~\citep{wu2022trajectoryguided,bartoccioni2025vavim,Renz2025cvpr} and the AV action space is two-dimensional (steering angle and acceleration).  Before presenting our mirror learning results we first explain our AV IDM.

\paragraph{Inverse Dynamics Model}
We based our AV IDM on the VaVim foundation model and its corresponding VaVam video action model~\citep{bartoccioni2025vavim}.  We chose this model as the basis of our IDM as it had been successfully utilized for AV motion planning and its training code was available. VaVaM is a stochastic flow-matching model of future actions conditioned on prior video observations.  An IDM, however, must infer the vehicle actions conditioned on the observations made while those actions were taken. To this end, we replaced the flow-matching future-action sequence head of the VaVaM architecture with our deterministic IDM head, using a squared error loss
\vspace{-0.5em}
\begin{equation}
    -\log p_\phi(a^{(i)}_{1:T-1} \mid o^{(i)}_{1:T} ) = \sum_{t=1}^{T-1}\left\lVert\frac{a_t^{(i)} - \mu_\phi(o_{1:T}^{(i)})_t}{2\sigma^2_\phi(o_{1:T}^{(i)})_t}\right\lVert_2^2 + \log\sigma_\phi(o_{1:T}^{(i)})_t. \label{eq:idm_nll}
\end{equation}
\vspace{-1.0em}
% \jy{Consider defining terms right before or after the equation}
% Our IDM shares a similar structure to the VaVaM denoiser model.  However, unlike VaVaM, we estimate $a_{1:T-1}^{(j)}$ using one function evaluation, and modify the output to predict both the mean action $\mu_\phi(o_{1:T}^{(j)})$ and the heteroscedastic aleatoric uncertainty $\sigma_\phi(o_{1:T}^{(j)})$ of each action. We found modeling both the mean and variance of the vehicle actions was essential to producing high quality action estimates. We train our VaVam IDM by minimizing \cref{eq:idm_nll} on a dataset of learner video action pairs. Once trained, we utilize the point estimate $\hat{a}^{(i)}_{1:T-1} = \mu_\phi(\hat{o}_{1:T}^{(j)})$ as action-labels for behavior cloning. 

Our IDM retains the overall VaVaM denoiser structure, but unlike VaVaM it predicts
$a_{1:T-1}^{(j)}$ in a single forward pass. The output head predicts both the mean
action $\mu_\phi(o_{1:T}^{(j)})$ and the heteroscedastic aleatoric uncertainty
$\sigma_\phi(o_{1:T}^{(j)})$ for each action. We found that modeling both mean and
variance was important for obtaining high-quality action estimates. We train the IDM
by minimizing \cref{eq:idm_nll} a dataset of learner video action pairs. Once trained, we use the point estimate $\hat{a}_{1:T-1}^{(i)} = \mu_\phi(\hat{o}_{1:T}^{(j)})$ as the action label for behavior cloning.

% \begin{table}[t]
%     \centering
%     \caption{Evidence that behavior cloning mirror data alone is sufficient for policy learning.  Size is number of BC samples, GT means ground truth, Mirror means mirror data.  Lower minADE$_5$, an open loop policy evaluation metric, is better.
%     %Evaluation results on \texttt{test\_1}, the test set constructed from the training towns, comparing policy heads trained from scratch using ground truth data and mirror data. 
%     }
%     \label{tab:mirror-data-only-vs-gt}
%     \begin{tabular}{ccc}
%         \toprule
%         Data & Size  & minADE$_5$  \\
%         \midrule
%         GT & 2500 & 1.03  \\
%         Mirror & 9600 & 0.95 \\
%         GT & 5000 & 0.91 \\
%         \bottomrule
%     \end{tabular}
% \end{table}

% \begin{figure}[t]
%     \centering
%     \includegraphics[width=0.5\linewidth]{Formatting_Instructions_For_NeurIPS_2026/figures/scaling_comparison.pdf}
% \caption{Evidence that behavior cloning mirror data alone is sufficient for policy learning.  While behavior cloning from ground truth (GT) data is unsurprisingly always best, here we see that it is possible to achieve equivalent performance using a greater quantity of mirror data only.  Lower minADE$_5$, an open loop policy evaluation metric, is better. Results averaged over three seeds.}
%     \label{fig:mirror-vs-gt}
% \end{figure}

% if we have an existing policy trained on a limited amount of ground truth data, does augmenting mirror data for finetuning  lead to policy improvements?

\begin{table}[tbp]
  \centering
  \caption{Open-loop policy evaluation (minADE$_5$) results for various  ground-truth plus mirror-generated dataset sizes. By design the GT+mirror rows have approximately same number of samples, 33.6k, and were formed by replacing part of the ground-truth data with mirror data. The policy was evaluated both in the same towns in which the training data was collected and zero-shot in a new, held-out town.  In all cases adding mirror data helps, both improving training environment and zero-shot held-out environment performance.}
  \label{tab:mirror-as-fraction-of-total}
  \setlength{\tabcolsep}{5pt}
  \begin{tabular}{lccccc}
    \toprule
    Setup & GT data (k) & Mirror data (k) & Mirror \% & Same towns & Held-out town \\
    \midrule
    GT only & 23.5 & 0.0  & 0\%  & 0.77$\pm$0.01 & 0.85$\pm$0.00 \\
        GT + mirror & 23.5 & 10.0 & 30\% & 0.75$\pm$0.01 & \textbf{0.76}$\pm$0.00 \\
    \midrule
    GT only & 27.9 & 0.0  & 0\%  & 0.76$\pm$0.01 & 0.80$\pm$0.05 \\
    GT + mirror & 27.9 & 5.6  & 15\% & \textbf{0.73}$\pm$0.01 & \textbf{0.76}$\pm$0.01 \\
   % \midrule
   % GT only & 33.6 & 0.0  & 0\%  & 0.74$\pm$0.01 & \textbf{0.73}$\pm$0.00 \\
    \bottomrule
  \end{tabular}
\end{table}

% \begin{table}[tbp]
%   \centering
%   \caption{Closed-loop Evaluation Town 10 20 routes balanced straight and turning scenarios, evaluated over three seed and seeded traffic manager for reproducibility.}
%   \label{tab:mirror-as-fraction-of-total}
%   \setlength{\tabcolsep}{5pt}
%   \begin{tabular}{lccccc}
%     \toprule
%     Setup & GT data (k) & Mirror data (k) & Mirror \% & Open-loop minADE$_5$ & Closed-loop Composed score \\
%     \midrule
%     GT only & 23.5 & 0.0  & 0\%  & 0.77$\pm$0.01 & 80.89$\pm$3.66 \\
%         GT + mirror & 23.5 & 10.0 & 30\% & 0.75$\pm$0.01 & \textbf{0.76}$\pm$0.00 \\
%    % \midrule
%    GT only & 33.6 & 0.0  & 0\%  & 0.74$\pm$0.01 & \textbf{0.73}$\pm$0.00 \\
%     \bottomrule
%   \end{tabular}
% \end{table}

% \begin{table}[t]
% \centering
% \caption{Closed-loop Evaluation Town 10 20 routes balanced straight and turning scenarios, evaluated over three seeds and seeded traffic manager for reproducibility.}
% \label{tab:closed_loop_results}
% \begin{tabular}{lcccc}
% \toprule
% Setup
% & Composed Score $\uparrow$
% & Collisions $\downarrow$
% & Route Completion (\%) $\uparrow$
% & Lane Violations (\%) $\downarrow$ \\
% \midrule
% \texttt{GT only (23.5k)}
% & $80.9 \pm 3.7$
% & $3.7 \pm 1.9$
% & $97.2 \pm 2.0$
% & $13.3 \pm 2.4$ \\

% \texttt{GT + mirror (23.5k+10.0k)}
% & $84.6 \pm 1.9$
% & $3.0 \pm 1.6$
% & $97.4 \pm 3.6$
% & $8.3 \pm 2.4$ \\

% \texttt{GT only (33.6k)}
% & $\mathbf{84.8 \pm 1.2}$
% & $\mathbf{2.3 \pm 1.7}$
% & $94.3\pm 5.3$
% & $8.3 \pm 2.4$ \\
% \bottomrule
% \end{tabular}
% \end{table}

\paragraph{Experimental Tasks} We answer two separate but important questions.  First, from a science perspective, is it possible to learn simply by behavior cloning mirror data?  We answer this question in the affirmative by training a policy in a new environment solely on mirror data.  Second, from a practical perspective, if we have an existing policy trained on a limited amount of ground truth data, does augmenting the fine-tuning data with mirror data lead to policy improvements?  If so, does it lead to improvements that are better than at least one other more common way of doing data augmentation?  We answer this in the affirmative by fine-tuning an existing AV policy on mirror data and show downstream performance improvement. Across these experiments, we try to get a sense of how ``valuable'' mirror data is, at least in the relative sense. Although ground truth first-person data remains the most valuable form of supervision, our results suggest that mirror learning can provide a practical source of training data when collecting first-person demonstrations is expensive or difficult.

\cref{fig:mirror-vs-gt}  demonstrates that BC of mirror data alone is sufficient to learn an effective policy. We trained stock VaVaM  conditioned on six history frames at 2 Hz to predict three seconds of future actions given a driving command. We trained six variants in total, using three different training dataset sizes and two different data sources.  To avoid leakage between training and evaluation the data used to train the MVM and IDM, third-person demonstrations used to generate mirror learning BC samples, and policy evaluation initial conditions are drawn from separate collections, \cref{sec:appendix:mirror_learning_experiments} details the exact splits. In particular, policy evaluation initial conditions are drawn from the same physical environments but with different weather, time of day, placement, and traffic configurations. Mirror-learning BC samples are generated by applying the trained MVM and IDM to third-person demonstrations, while ground-truth samples are standard first-person experience replay traces. We evaluate policies open-loop using minimum average predicted displacement error over five samples, minADE$_5$~\citep{bartoccioni2025vavim,wang2025alpamayo}. Unsurprisingly, ground truth BC data is optimal at all training dataset sizes. Critically, however, this experiment demonstrates that it is possible to learn an effective policy using zero ground truth first-person data. If you have a vision-only policy, an IDM and an MVM, you can learn from third-person demonstration alone. In this particular application the relative ``worth'' of a mirror sample is about half that of a ground truth sample.  The inter-line ``gap'' in the plot above suggests that one can achieve similar policy performance by either instrumenting and recording a certain amount of first person behavior cloning data or by watching and mirroring two times the same.  In settings where instrumenting agents and/or teleoperation is costly or dangerous this could be meaningfully exploitable.

While the possibility of learning from mirror data alone is scientifically interesting, current best engineering practice dictates that ground truth BC data will be captured and used as fully as possible whenever possible. The questions are whether or not using additional mirror data in behavior cloning pipelines is beneficial and whether behavior cloning has any advantage relative to other, more typical, methods of data augmentation. And in the AV setting specifically, whether using mirror learning to augment first person capture in new geography expansion tasks is beneficial.  We find that mirror learning amplifies geographic expansion by augmenting limited first-person data with mirror data.

% \cref{tab:mirror-vs-other-augmentation} examines a realistic mirror learning deployment case through the lens of a specific example of geographic expansion of a vision-based AV operator where a small number of vehicles is used to record ground-truth data in a new location.  Such data would be used in practice to directly fine-tune the policy.  Augmentation techniques such as style transfer \citep{ali2025world} might be utilized as well to compound the value of the new location ground truth data.  A new practice possible given our approach to mirror learning is to fine-tune the IDM too on new town data, then augment with mirror data from demonstrators in the new town data.  The results in the table suggest that adding mirror data is more effective than style transfer augmentation and remains effective at ratios up to two times the ground truth data size.  No additional augmentation ratios or ground truth volumes were considered simply due to computational constraints.  The style transfer approach was to text prompt Cosmos to change stylist aspects of the scene, examples of which are shown in \cref{sec:appendix:style-transfer-results}.  Note that we do not claim that mirror learning is superior to style transfer in general; testing mirror learning against other augmentation approaches is beyond the scope of this paper.

\cref{tab:mirror-vs-other-augmentation} examines a mirror learning deployment case: geographic expansion of a vision-based AV operator, where a small fleet records limited ground-truth data in a new location. Such data would typically be used to fine-tune the policy directly. Augmentation techniques such as style transfer \citep{ali2025world} might be utilized as well to compound the value of the new location ground truth data.  A new practice possible given our approach to mirror learning is to fine-tune the IDM too on new town data, then augment with mirror data from demonstrators in the new town. The results in the table suggest that adding mirror data is more effective than style transfer augmentation and remains effective at ratios up to two times the ground truth data size. No additional augmentation ratios or ground truth volumes were considered simply due to computational constraints. The style transfer approach was to text prompt Cosmos-Transfer~\citep{ali2025world} to change stylist aspects of the scene (see \cref{sec:appendix:style-transfer-results} for examples).

\cref{tab:mirror-as-fraction-of-total} completes our mirror learning evaluation by examining, simultaneously, whether mirror learning is generally helpful as a ratio of total data and whether mirror learning interferes or generally helps with zero-shot performance in held-out locations.  Results indicate that mirror learning is generally helpful, at various mixtures of ground truth  and mirror data scale.

\section{Related Work}

\paragraph{Learning from demonstrations by observation} must address two missing pieces, demonstrator actions and demonstrator egocentric observations. Prior work recovers missing actions with inverse dynamics models trained on the learner's own observations~\citep{torabi2018behavioral,agrawal2016learning}, but this requires demonstrations to already lie in the learner's observation space. To handle viewpoint mismatch, other methods learn view-invariant representations or explicit cross-view translation~\citep{sermanet2018time,stadie2017third,liu2018imitation,shang2021self} mostly in constrained environments. Most relevant is \citet{liu2018imitation}, who learn a context translation model conditioned on a single learner-context frame to translate demonstrations into the learner's observation context. However, they use the learned intermediate translation features as perceptual rewards for reinforcement learning rather than as synthetic supervision for behavior cloning.

% Most relevant is \citet{liu2018imitation}, who assume access to a single demonstrator egocentric context frame and use it to translate demonstrations into the learner's view, but use the result as a perceptual reward for reinforcement learning rather than as synthetic supervision for behavior cloning.

% removed for saving space 
% Recent work has explored imitation from human videos for robot learning. MimicPlay~\citep{wang2023mimicplay} uses multi-view human play data together with a small amount of robot teleoperation data to infer latent human plans, which are then used to guide a low-level robot policy. While this line of work similarly learns from passive visual demonstrations, it focuses on single-agent manipulation in static environments. By contrast, our setting is explicitly multi agent and dynamic; the learner must infer a demonstrator's first person trajectory while both agents interact with the same evolving scene.

Prior autonomous driving work performs learner-frame transfer in structured intermediate representations rather than generating mirrored RGB video~\citep{zhang2021learning,chen2022learning,feng2025rap}. Learning by Watching~\citep{zhang2021learning} uses occluded birdview rasterization, Learning from All Vehicles~\citep{chen2022learning} fuses RGB, segmentation, and LiDAR into a shared 2D representation, and RAP~\citep{feng2025rap} uses lightweight 3D rasterization with raster-to-real alignment. These methods expand training data through structured observation transfer, but do not produce realistic first-person RGB supervision for modern end-to-end driving policies.

\textbf{Third-to-first video transfer} synthesizes first-person videos from temporally synchronized third-person observations and is closely related to controllable video generation. Methods typically rely on strong geometric or multi-view supervision, either by decoupling structure transfer from appearance synthesis or by injecting explicit 3D priors~\citep{liu2024exocentric,mahdi2025exo2egosyn,kang2025egox}. EgoX~\citep{kang2025egox}, for example, builds on pretrained video diffusion models but is not geometry-free. It requires target egocentric camera poses, point-cloud derived egocentric priors, and fixed exocentric camera poses at inference. WorldWander~\citep{song2025worldwander} is also particularly relevant because it frames the problem as conditional video-to-video generation with a video diffusion model, but its EgoExo-8k dataset focuses on a single continuously tracked target that is typically centered in the frame. By contrast, our approach does not require explicit 3D scene reconstruction, allows both learner and demonstrator to move throughout the scene in both CARLA and Minecraft, and operates in a more challenging multi-agent setting 
in which multiple candidate demonstrators may be present. Unlike WorldWander, we also show that synthesized demonstrator ego-view videos can be paired with IDM-inferred actions to form mirror data for behavior cloning.

% and operates in a more challenging multi-agent setting. Unlike WorldWander, we also show that mirror data, when paired with an IDM, can serve as synthetic supervision for behavior cloning.

% and operates in a more challenging multi-agent setting where demonstrator positions vary throughout the learner's observations. Unlike WorldWander, we also show that synthesized demonstrator ego-view videos can be paired with IDM-inferred actions to form mirror data for behavior cloning.

\textbf{End-to-End Driving Policy.} Our choice,  VaVAM~\cite{bartoccioni2025vavim}, first pre-trains an autoregressive driving video model, VaViM, on large-scale unannotated driving videos then uses those representations in a lightweight imitation-learning stage for trajectory prediction.  Alternatives such as SimLingo~\cite{Renz2025cvpr} and NVIDIA's Alpamayo-R1~\cite{wang2025alpamayo} exist and are good alternatives to consider in future work.

\section{Conclusions}
Our central empirical finding is that pretrained video diffusion models can possess sufficiently rich internal representations of physical scene structure to support mirror video synthesis after lightweight fine-tuning — without requiring explicit camera parameters, pose estimation, or scene reconstruction. We demonstrated this capacity across two visually and structurally diverse domains (CARLA autonomous driving and open-world multiplayer Minecraft), and showed zero-shot generalization from synthetic training data to real-world robotaxi footage.

For the autonomous vehicle setting, where full mirror learning evaluation was tractable, we showed that (i) an effective driving policy can be learned using mirror data alone, with no first-person ground truth; (ii) augmenting existing ground-truth data with mirror data consistently improves open-loop policy evaluation (minADE$_5$) both in training environments and in held-out geographic locations; and (iii) mirror data augmentation outperforms style transfer at matched augmentation ratios in the geographic expansion setting.

These results suggest a path toward scalable, safe policy acquisition in settings where teleoperation is costly, dangerous, or infeasible. Mirror learning as presented need not entirely replace first-person data collection, but it meaningfully reduces its required volume and opens the possibility of leveraging the rich observational signals that are already present in any multi-agent environment. We view the current work as an early instantiation of a broader class of observation-driven learning systems, and anticipate that improvements in video world model capacity will further increase the fidelity and downstream value of synthesized mirror data. 

Code, datasets, and model checkpoints will be released upon acceptance to facilitate reproducibility and to provide baseline metrics for future work on mirror video modeling.

\clearpage

\clearpage
\bibliographystyle{plainnat}
\bibliography{neurips}
\newpage
\appendix

\section{Appendix}

We organize the appendix as follows. We first describe the datasets used throughout the paper, which we view as an important contribution that supports future development of mirror video modeling, then summarize training details shared across experiments, and finally present experiment-specific settings together with the corresponding quantitative and qualitative results.

\subsection{Self-Driving Datasets}

\begin{figure}[t]
    \centering
    \includegraphics[width=1.0\linewidth]{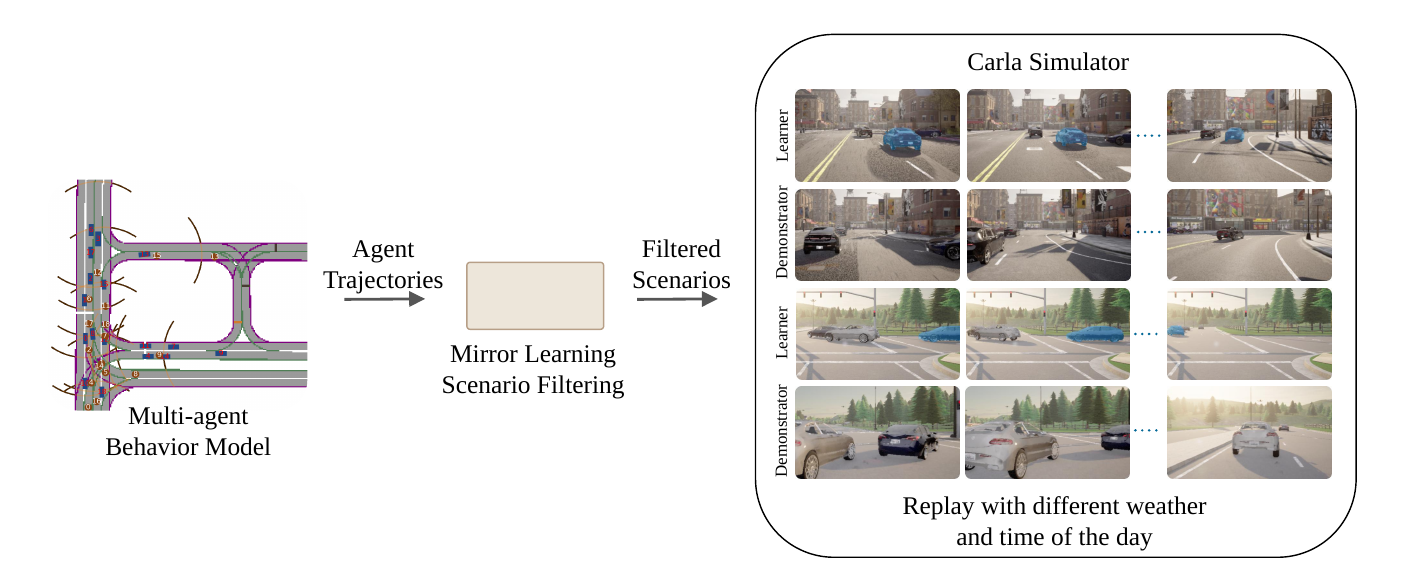}
\caption{\textbf{Data generation pipeline.} Diverse multi-agent traffic scenarios are generated in Bird's-eye-view CARLA maps using a realistic scenario generator trained on real-world trajectories~\citep{lioutas2025control}. After task-specific filtering for our mirror learning setting, the selected scenarios are replayed in the simulator to record candidate paired observations and the agents' executed trajectories, while demonstrator instance masks are obtained from CARLA's instance segmentation sensor.}
    \label{fig:data_pipeline}
\end{figure}

% \caption{\textbf{Data generation pipeline.} Diverse multi-agent traffic scenarios are generated in bird's-eye-view CARLA maps using a realistic scenario generator trained on real-world trajectories~\citep{lioutas2025control}. After task-specific filtering for our third-to-first transfer setting, the selected scenarios are replayed in simulation to record candidate paired observations and the agents' executed trajectories, while demonstrator instance masks are obtained from CARLA's instance segmentation sensor.}

% \subsubsection{Setup}
\subsubsection{CARLA Dataset}
\label[appendix]{sec:appendix:CARLA-data-collection}

% \begin{table}[th]
% \centering
% \caption{CARLA dataset sample counts by town.}
% \label{tab:carla3towns-stats}
% \begin{tabular}{lr}
% \toprule
% Town & \# Samples  \\
% \midrule
% Town04 & 5172  \\
% Town06 & 13391  \\
% Town10 & 13976 \\
% Town03 (Held-out town) & 4587 \\
% \midrule
% \textbf{Total} & \textbf{37126} \\
% \bottomrule
% \end{tabular}
% \end{table}

\begin{table}[t]
\centering
\caption{Counts of paired demonstrator--learner clips in the CARLA dataset by Town.
Each clip spans 34 frames at 4\,fps (8.5\,s), for a total of approximately 52 hours of paired data.}
\label{tab:carla3towns-stats}
\begin{tabular}{lr}
\toprule
Town & \# of paired clips \\
\midrule
Town04          & 3{,}096  \\
Town06          & 7{,}286  \\
Town10          & 6{,}571  \\
Town03 (Held-out town)         & 5{,}013  \\
\midrule
\textbf{Total}  & \textbf{21{,}966} \\
\bottomrule
\end{tabular}
\end{table}

Existing open-source end-to-end driving datasets are predominantly egocentric (first-person) and generally do not provide paired third-person and first-person observations of the same scene~\cite{fong2021panoptic,xu2025wod,nvidia_physicalai_autonomous_vehicles_2025}. In contrast, this kind of paired data exists naturally in proprietary robotaxi fleets, where multiple instrumented vehicles can observe the same scene from different viewpoints, but such data is not publicly available. Moreover, the scale of open driving data remains orders of magnitude smaller than what industrial systems are believed to leverage. For example, NVIDIA's open driving release contains on the order of 1,700 hours of real-world data, whereas Tesla has publicly stated that it trains on over 10 billion miles of fleet data. Motivated by the lack of publicly available paired data with action labels at sufficient scale, we study our approach in the CARLA simulator.

A potential concern is that simulated data, especially agent behavior produced by CARLA's default Traffic Manager, may be insufficiently realistic. To mitigate this sim-to-real gap in paired-data mining, we use a realistic multi-agent scenario generation model~\citep{lioutas2025control} trained on a large corpus of real-world trajectories to generate diverse traffic scenarios in bird's-eye-view CARLA maps.  We further apply task-specific filtering to retain only scenarios that satisfy the requirements of our third-to-first video transfer setting (e.g., adequate third-person visibility of the demonstrator). These scenarios are then replayed in the simulator to identify candidate paired observations for recording. Instance masks are obtained through CARLA's instance segmentation sensor. The full data creation pipeline is shown in the Figure~\ref{fig:data_pipeline}. We will release the dataset and the collection pipeline upon acceptance.  

% \paragraph{CARLA Dataset Collection.}
% We collected our CARLA dataset using a data-driven object-level driving policy~\citep{lioutas2025control} to generate thirty-two thousand paired demonstrator/learner videos. Each video is six seconds long and rendered at four frames per second. The data are generated by ``driving'' realistic vehicle-only traffic in four disjoint CARLA towns. For each scenario, we select pairs of vehicles that satisfy visibility constraints between the learner and demonstrator, and render the corresponding demonstrator view, learner view, and instance segmentation masks using CARLA's built-in segmentation renderer. 

% We collected our CARLA dataset using a data-driven object-level driving policy~\citep{lioutas2025control} to generate thirty-two thousand paired demonstrator/learner videos. 

\paragraph{CARLA Dataset Collection.}
% To generate diverse realistic traffic scenarios on the Carla map, we first use a traffic scenario initialization model~\citep{} to initialize the traffic agents starting position and speed on the Carla map the use  a data-driven object-level driving policy~\citep{lioutas2025control} to generate thirty-two thousand paired demonstrator/learner videos. Both the traffic initialization and the policy model are trained with large amount real multi-agent dirving behavior data for generating realistic scene and interactions. Each video is six seconds long and rendered at four frames per second. The data are generated by ``driving'' realistic vehicle-only traffic in four disjoint CARLA towns; the number of collected samples per town is reported in Table~\ref{tab:carla3towns-stats}.  For each replayed scenario, we select pairs of learner and demonstrator vehicles that satisfy visibility constraints. The demonstrator must be within the learner camera field of view, with horizontal field of view $70^\circ$ and range $24$ m, and have an unoccluded line of sight. We retain pairs with at least one continuous visible window of $30$ or more timesteps at 10 Hz. We discard low-motion windows in which the target is stalled, defined as speed below $0.1$ m/s for at least $10$ consecutive timesteps, since these examples provide limited interaction signal. 

\begin{figure*}[t]
    \centering
    \includegraphics[width=\textwidth]{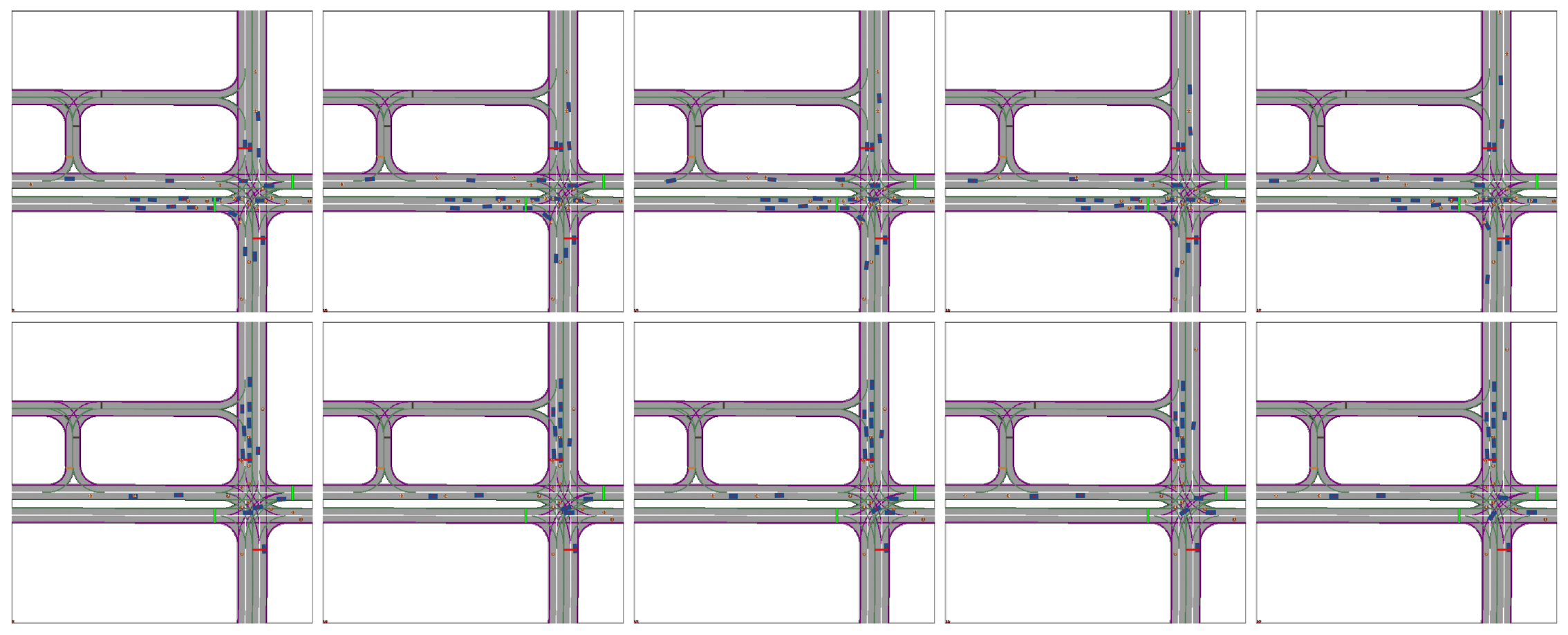}
    \caption{
    Bird's-eye-view traffic scenarios generated by a multi-agent driving behavior model~\citep{lioutas2025control} in CARLA Town10 and used in our experiments.
    The road geometry is shown in gray, blue rectangular boxes indicate vehicles, and traffic lights are annotated at their corresponding map locations.
    This visualization provides spatial context for the scene layout, vehicle distribution, and signalized intersections in the Town10 environment. Brown circles labeled with different numbers indicate the goal positions for each agent. 
    }
    \label{fig:town10-Bird's-eye-view-map}
\end{figure*}

We generate diverse and realistic traffic scenarios in CARLA by using a traffic scenario initialization model~\citep{zwartsenberg2022conditional} to sample agent starting positions and speeds, and then roll out a data-driven object-level driving policy~\citep{lioutas2025control}. Both the initialization model and the policy are trained on a large scale real world multi-agent driving behavior data, enabling realistic traffic scenes and interactions. This process yields 37,000 paired demonstrator and learner videos. Each video is rendered at four frames per second for six seconds, while the underlying simulation trajectories are generated at 10 Hz. The scenarios are generated by rolling out realistic vehicle-only traffic in four distinct CARLA towns; the number of collected samples per town is reported in Table~\ref{tab:carla3towns-stats}. Samples of the traffic scenario are shown in~\cref{fig:town10-Bird's-eye-view-map}

For each replayed scenario, we select learner and demonstrator vehicle pairs that satisfy visibility constraints using the underlying 10 Hz trajectories. The demonstrator must lie within the learner's camera field of view, using a horizontal field of view of $70^\circ$ and a range of $24$ m, and must have an unoccluded line of sight. We retain pairs with at least one continuous visible window of $30$ or more simulation timesteps. We discard low-motion windows in which the target is stalled, defined as speed below $0.1$ m/s for at least $10$ consecutive simulation timesteps, since these examples provide limited interaction signal.

In all recordings, RGB cameras use the shared first-person-view preset, which places the camera approximately $2.2$ m forward and $1.5$ m above the vehicle origin with default rotation. This transform is used for both learner and demonstrator RGB streams. The instance segmentation camera used to obtain demonstrator masks is mounted on the learner vehicle with the same transform. During replay, the learner vehicle is fixed to Tesla Model 3, while other replayed agents, including target vehicles, are sampled from a small passenger vehicle pool.

Because all replayed demonstrator and learner vehicles are passenger cars, their front-view camera placements are similar across vehicle types. We therefore use a shared egocentric camera convention. The mirror video model predicts the demonstrator's front-view video in this common observation space. Since the inverse dynamics model is trained only on our ego vehicle type, decoded actions are interpreted as ego-executable actions that would reproduce the predicted mirror video, rather than as the target demonstrator's original controls.

% \paragraph{Scenario-Level Splits.}

\paragraph{Split Rationale.}

\begin{figure*}[t]
    \centering
    \begin{tikzpicture}
        \node[anchor=south west, inner sep=0] (img) at (0,0)
            {\includegraphics[width=\textwidth]{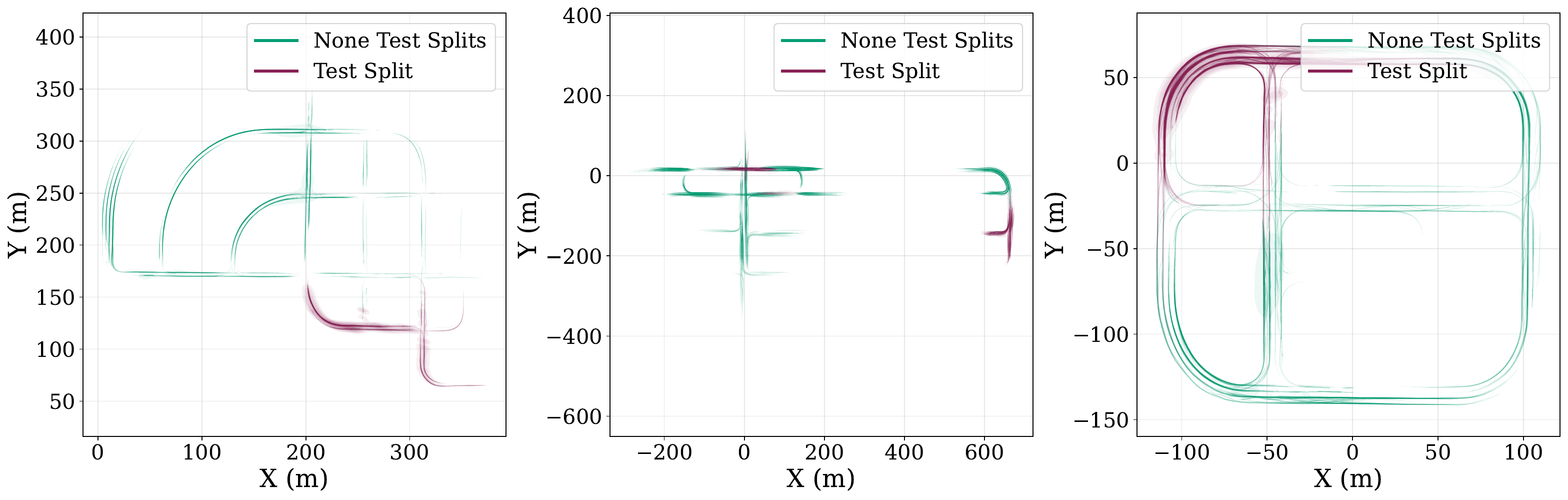}};

        \begin{scope}[x={(img.south east)}, y={(img.north west)}]
            \node[anchor=north, font=\small\bfseries] at (0.19, -0.035) {Town 04};
            \node[anchor=north, font=\small\bfseries] at (0.530, -0.035) {Town 06};
            \node[anchor=north, font=\small\bfseries] at (0.865, -0.035) {Town 10};
        \end{scope}
    \end{tikzpicture}

    \vspace{1.1em}
    \caption{Spatial trajectory density comparison between the test split and the remaining data from Town 04, Town 06, and Town 10. We partition the data from Town 04, Town 06, and Town 10 into five disjoint splits: a training split for MVM training, a validation split for MVM model selection, a mirror-data generation split, a policy-validation split, and a test split. Importantly, the learner observations used to generate mirror data are excluded from MVM training. The test split contains different scenarios and spatial regions that are underrepresented in the other splits, as shown in this figure.
}
    \label{fig:town10_train_test_traj}
\end{figure*}
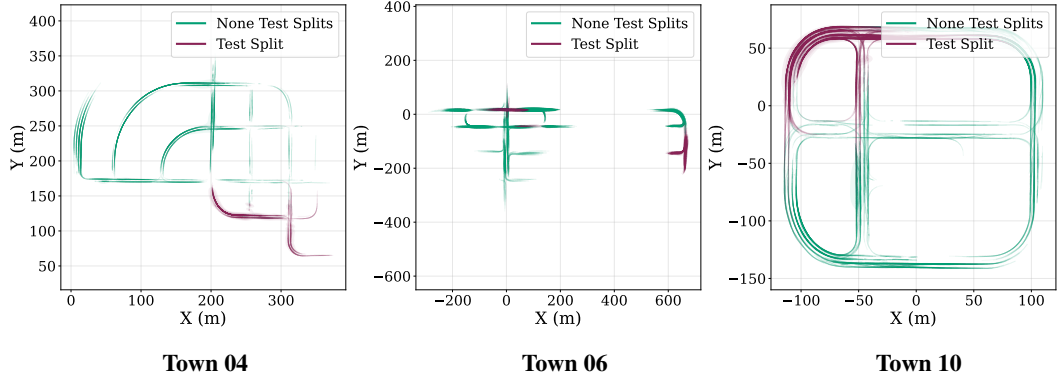

We split the CARLA data at the traffic-scenario level, where a traffic scenario denotes a single recorded multi-agent interaction rollout generated by the multi-agent trajectory prediction model and a recording denotes one paired demonstrator--learner clip extracted from that rollout. All recordings from the same scenario are assigned to the same split. Among the four towns, we reserve Town 03 as a held-out town for evaluation. We partition the data from Town 04, Town 06, and Town 10 into five disjoint splits: a training split for MVM training, a validation split for MVM model selection, a mirror-data-generation split, a policy-validation split, and a test split. Importantly the learner observations used to generate mirror data are excluded from MVM training. The test split uses different scenarios and spatial regions that are underrepresented to the rest of the splits. Figure~\ref{fig:town10_train_test_traj} compares trajectory densities from the test split and the remaining data from Town 04, Town 06, and Town 10. 

We do not enforce geographic separation among the MWM training, MWM validation, mirror-data-generation, and policy-validation splits within the non-held-out towns (Town 04, Town 06, and Town 10). This choice reflects two considerations. First, in practical autonomous driving deployment, additional fleet data is often collected from locations that have already been traversed. Second, even within a fixed town, variation in traffic scenarios, weather, and times of day provides a meaningful setting for evaluating whether the mirror video model generalizes beyond the exact conditions seen during training. We reserve stronger geographic generalization for evaluation on Town 03.

% Spatial trajectory density comparison between Train Large and Test One. In each town, test locations are separated from training locations or placed in less dense regions for policy evaluation.

\subsubsection{May Mobility Multi-Agent Dataset}
\label{sec:may-data-details}
We also evaluate zero-shot transfer on the multi-agent subset of MARS~\cite{li2024multiagent}, a real-world multimodal autonomous-driving dataset collected from May Mobility vehicles. It contains 53 synchronized 30-second scenes with over 15,000 frames, capturing nearby vehicles with overlapping viewpoints for collaborative perception. Each vehicle records six surround-view RGB cameras, a 128-channel LiDAR, IMU, and GPS at 10 Hz in a NuScenes-compatible format. Only a small subset of scenes is suitable for our zero-shot setting, where one fleet vehicle observes another and provides ground-truth ego-view for the demonstrator. After manually filtering for paired learner-demonstrator data in which the demonstrator is visible in the learner's view, fewer than 20 scenes meet these requirements.

\paragraph{Mask Extraction.}
In general, we obtain the demonstrator's mask sequence using SAM~3~\citep{carion2025sam3segmentconcepts} in video mode with a text prompt such as ``car,'' which yields per-frame instance masks together with temporally consistent object identifiers. In many settings, this is sufficient to track the target agent directly through its persistent identifier. For the May Mobility data, we use an additional dataset-specific selection step only for convenience, since ground-truth ego-view video is only available for a single demonstrator robotaxi and we therefore require a reliable way to recover its corresponding track in the learner view for evaluation. Specifically, we use ego-pose annotations, camera intrinsics, and a fixed 3D vehicle cuboid to project an approximate 2D target bounding box into each frame, and then select the SAM~3 instance mask that best matches this box using a score based on box overlap and mask coverage, with a small temporal consistency bonus to reduce flicker. This yields an occlusion-robust target mask sequence without manual per-frame annotation.

\subsection{Minecraft Dataset}
\label[appendix]{sec:appendix:plaicraft-dataset}

\paragraph{Training and Validation Split for the Minecraft Dataset}
The Minecraft data used in this study is derived from the PLAICraft dataset~\citep{he2025plaicraft}, which comprises multiple gameplay sessions from a diverse set of participants. To make this data suitable for model training, we implement a systematic preprocessing pipeline. Our pipeline identifies visibility events between pairs of players through a three-stage refinement process. First, we generated a large set of candidates via geometric constraints on distance, field of view, and vertical alignment. Second, we employ OpenCV-based heuristics to remove cases where the learner view is obstructed by GUI elements using edge, color, and shape cues. Finally, SAM3~\citep{carion2025sam3segmentconcepts} based instance segmentation verifies true visibility by generating and propagating player masks across short video clips. Candidate clips are retained only if sufficiently large masks persist across most frames (more than 80\%). 

The extracted clips are fixed-length temporal segments of 3 seconds, sampled at 10 frames per second (FPS), resulting in 30 frames per clip at a spatial resolution of 640x360 pixels. The dataset consists of a total of 712 gameplay sessions, which are partitioned at the session level into 676 training sessions and 36 validation sessions. We assign entire gameplay sessions exclusively to either the training or validation split to eliminate temporal and contextual overlap, preventing data leakage and ensuring that validation samples come from unseen gameplay trajectories.

To further increase the number of training samples and capture temporal dynamics, we construct windowed segments using a sliding window strategy within each clip. Specifically, each 30-frame clip is subdivided into sequences of length 25 frames with a stride of 2, producing 3 overlapping segments per clip. This results in a total of 6,003 clips (18,009 segments) for training and 253 clips (759 segments) for validation, yielding 6,256 clips and 18,768 windowed segments overall.

\subsection{Shared Training Details}
\label[appendix]{sec:appendix:shared_training_details}
Unless otherwise noted, the following settings are shared across experiments in the paper.

\paragraph{Mirror Video Model.}
We fine-tune NVIDIA Cosmos with a learning rate of $3 \times 10^{-4}$. For the CARLA dataset, the model takes as input a 25-frame video clip at 512$\times$288 resolution sampled at 4~Hz together with the corresponding instance mask at each frame, and predicts a 25-frame video clip at the same frame rate. For the Minecraft dataset, we use the same resolution at 10 fps. At inference time, we use Cosmos's built-in multistep flow-matching solver~\cite{zhao2023unipc} with 35 denoising steps. We train MVM on the CARLA dataset using four H100 GPUs for 15{,}000 steps with a batch size of 10 per GPU, which takes approximately five days. The Minecraft model was trained on eight L40S GPUs for 9800 steps with a batch size of 5 per GPU, which took approximately 42 hours. Cosmos also takes a text prompt as input. In this work, we use a fixed generic prompt for each dataset. For CARLA, we use: \emph{``Photoreal CARLA simulation, single continuous take. Imagine the first-person perspective of the target car from a third-person observation, with consistent vehicle and lane geometry. Follow kinematics: no jumps, cuts, or teleports.''} We leave the exploration of more informative text prompts to future work.

\paragraph{Policy Learning.}
For downstream policy learning, we adopt VaViM as the visual backbone, initialize from the pretrained model, and add an action prediction head as in VaVAM, without any architecture modifications. The backbone hidden size is 768. Unless otherwise noted, all policy variants share the same architecture and training protocol. Following VaVAM, we evaluate forecasting accuracy using minimal average displacement error, minADE$_5$, which has been used for end-to-end policy evaluation in works such as Alpamayo-R1~\citep{wang2025alpamayo}. The policy is conditioned on a fixed history window of 6 observations and predicts 6 future actions at 2~Hz. We tune the learning rate with a hyperparameter sweep and use $10^{-3}$ for all policy experiments. 

\paragraph{Inverse Dynamics Model.}
For IDM training, we use a learning rate of $0.0194$ with a batch size of 16. The model receives a sequence of 7 observations as input and predicts the 6 actions between consecutive observations. Driving commands are heuristically classified from the output actions using the heuristic provided in the VaVaM code. We train for 10 epochs and observe that the validation average displacement error (ADE) has converged by the end of training. Training is performed using 4 NVIDIA RTX A5000 GPUs. We train the IDM on learner observations and actions from the MVM training and validation splits in the non-held-out towns, while keeping the mirror-data generation split disjoint from the IDM training data.

\subsection{Mirror Video Model Experiments}

This section provides supplementary results for the mirror video model. We begin with CARLA evaluation, including both quantitative results and qualitative generalization examples, then present zero-shot transfer to the real-world May Mobility dataset, and finally show qualitative results on Minecraft. We also include additional mirror video qualitative results in \cref{sec:appendix:addl-mvm-qual}.

\subsubsection{CARLA Evaluation}
\label[appendix]{sec:appendix:carla_eval}
Unless otherwise noted, for the CARLA dataset we train the mirror video model on the MVM training split mentioned in ~\cref{sec:appendix:CARLA-data-collection}. After careful filtering, which yields 14,500 training segments constructed with a sliding window stride of 4. Quantitative evaluation is performed on 1,024 validation samples from the MVM validation split across Town 04, Town 06, and Town 10 (same towns used in training). Town 03 is excluded from mirror video model training and reserved for qualitative generalization analysis.

\paragraph{Quantitative Results.}
To facilitate reproducibility and provide baseline metrics for future work on mirror video modeling, we report quantitative results for the mirror video model in~\cref{tab:carla-metrics}. Metrics are computed over 1,024 validation samples for Town 04, Town 06, and Town 10. Town 03 is a held-out town not included in mirror video model training. We report FVD, PSNR, and SSIM, where lower FVD indicates better video distribution fidelity and higher PSNR/SSIM indicate better frame-level reconstruction quality.

\begin{table}[t]
    \centering
    \caption{Quantitative evaluation on CARLA towns for the mirror video model. Town 04, Town 06, and Town 10 are training towns and Town 03 is held out. Lower FVD is better; higher PSNR and SSIM are better.}
    \label{tab:carla-metrics}
    \small
    \begin{tabular}{lccc}
        \toprule
        \textbf{Town} & \textbf{FVD} $\downarrow$ & \textbf{PSNR} $\uparrow$ & \textbf{SSIM} $\uparrow$ \\
        \midrule
        Town 04 & 56.27 & 19.12 & 0.59 \\
        Town 06 & 39.88 & 23.51 & 0.76 \\
        Town 10 & 39.63 & 19.61 & 0.54 \\
        \midrule
        Town 03 & 58.82 & 19.98 & 0.66 \\
        \bottomrule
    \end{tabular}
\end{table}

% \begin{table}[t]
%     \centering
%     \caption{Quantitative evaluation on the held out Minecraft dataset for the mirror video model. Lower FVD is better; higher PSNR and SSIM are better.}
%     \label{tab:minecraft-metrics}
%     \small
%     \begin{tabular}{lccc}
%         \toprule
%         \textbf{FVD} $\downarrow$ & \textbf{PSNR} $\uparrow$ & \textbf{SSIM} $\uparrow$ \\
%         \midrule
%         227.60 & 16.42 & 0.45 \\
%         \bottomrule
%     \end{tabular}
% \end{table}

\paragraph{Qualitative Results.}
We show qualitative results beginning with the held-out Town 03 split, followed by results from the training towns in the additional results section.

\FloatBarrier
\subsubsection{CARLA Held-Out Town}
\label[appendix]{sec:appendix:harder-test-town}
% We additionally evaluate on CARLA Town03, a held-out urban environment with large junctions and a roundabout. This map provides a diverse city-driving setting for assessing qualitative generalization to unseen scenes. We show qualitative results in Figures~\ref{fig:carla_town03_1}--\ref{fig:carla_town03_7}. In addition, in Figures~\ref{fig:carla_town03_passing_1} to Figures~\ref{fig:carla_town03_passing_4}Figures~\ref{fig:carla_town03_passing_4} we show challenging scenarios in the held-out town where the learner has limited overlap with the demonstrator's front-facing view. 

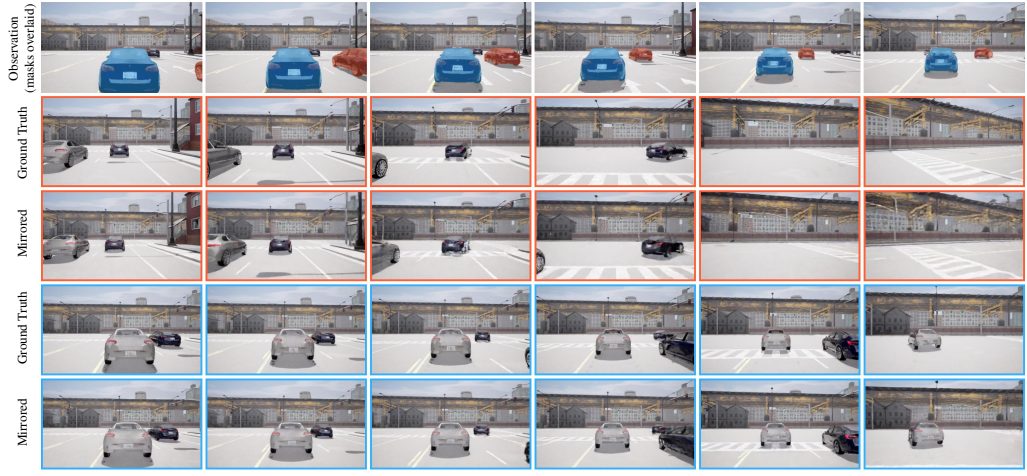
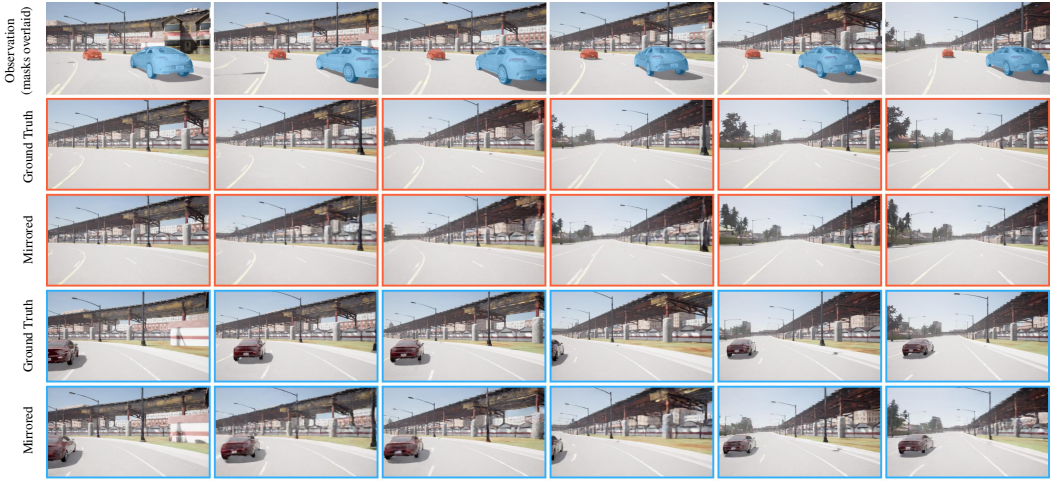
\begin{figure*}[th]
    \centering

    \begin{subfigure}{\textwidth}
        \centering
        \begin{tikzpicture}
            \node[anchor=south west, inner sep=0] (img) at (0,0)
                {\includegraphics[width=0.93\linewidth]{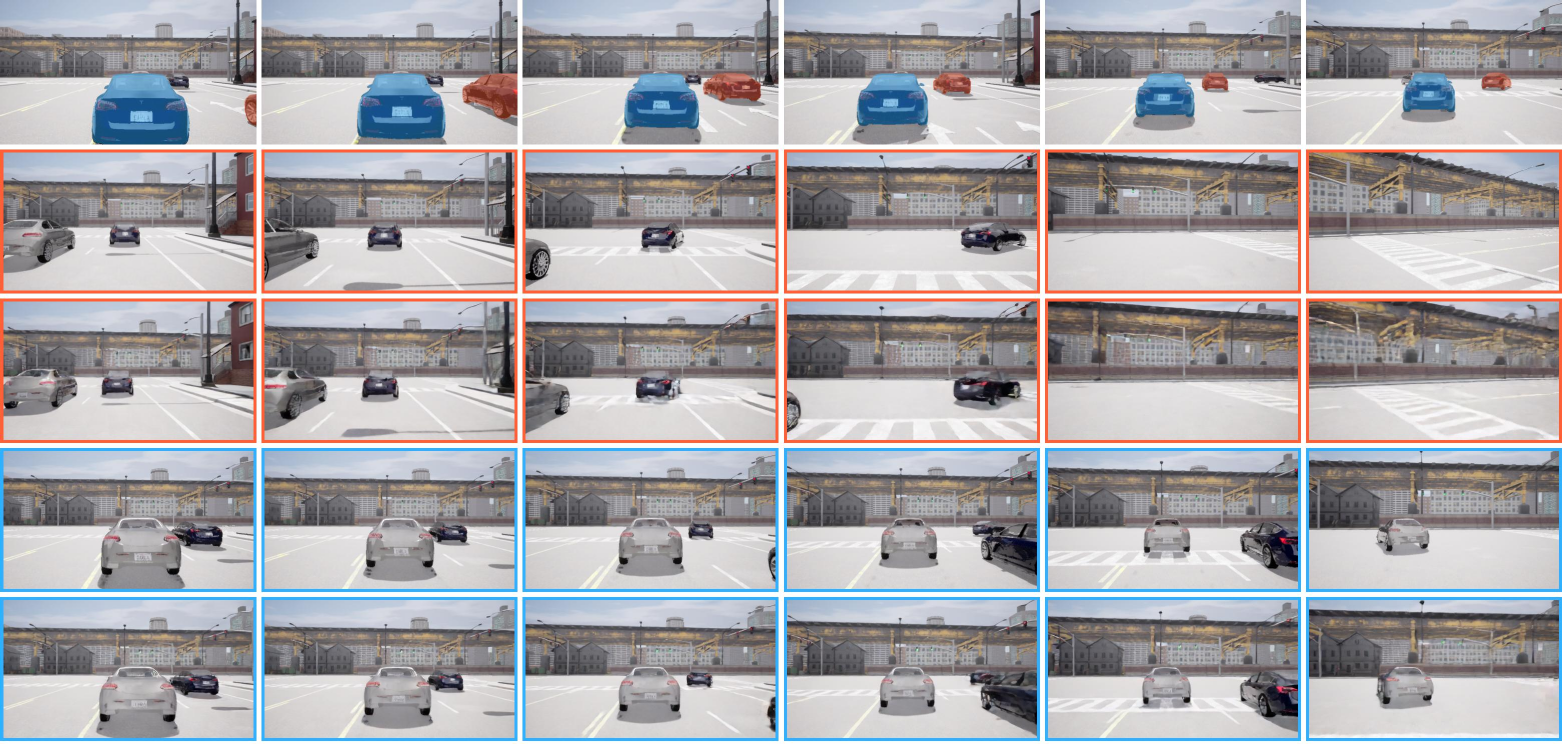}};
            \node[anchor=center, font=\fontsize{5}{6}\selectfont, align=center, rotate=90] at (-0.25, 5.6) {Observation\\(masks overlaid)};
            \node[anchor=center, font=\fontsize{5}{6}\selectfont, align=center, rotate=90] at (-0.25, 4.35) {Ground Truth};
            \node[anchor=center, font=\fontsize{5}{6}\selectfont, align=center, rotate=90] at (-0.25, 3.15) {Mirrored};
            \node[anchor=center, font=\fontsize{5}{6}\selectfont, align=center, rotate=90] at (-0.25, 1.90) {Ground Truth};
            \node[anchor=center, font=\fontsize{5}{6}\selectfont, align=center, rotate=90] at (-0.25, 0.65) {Mirrored};
        \end{tikzpicture}
        \caption{}
        % \label{fig:carla_town03_1}
    \end{subfigure}

    \vspace{0.5em}

    \begin{subfigure}{\textwidth}
        \centering
        \begin{tikzpicture}
            \node[anchor=south west, inner sep=0] (img) at (0,0)
                {\includegraphics[width=0.95\linewidth]{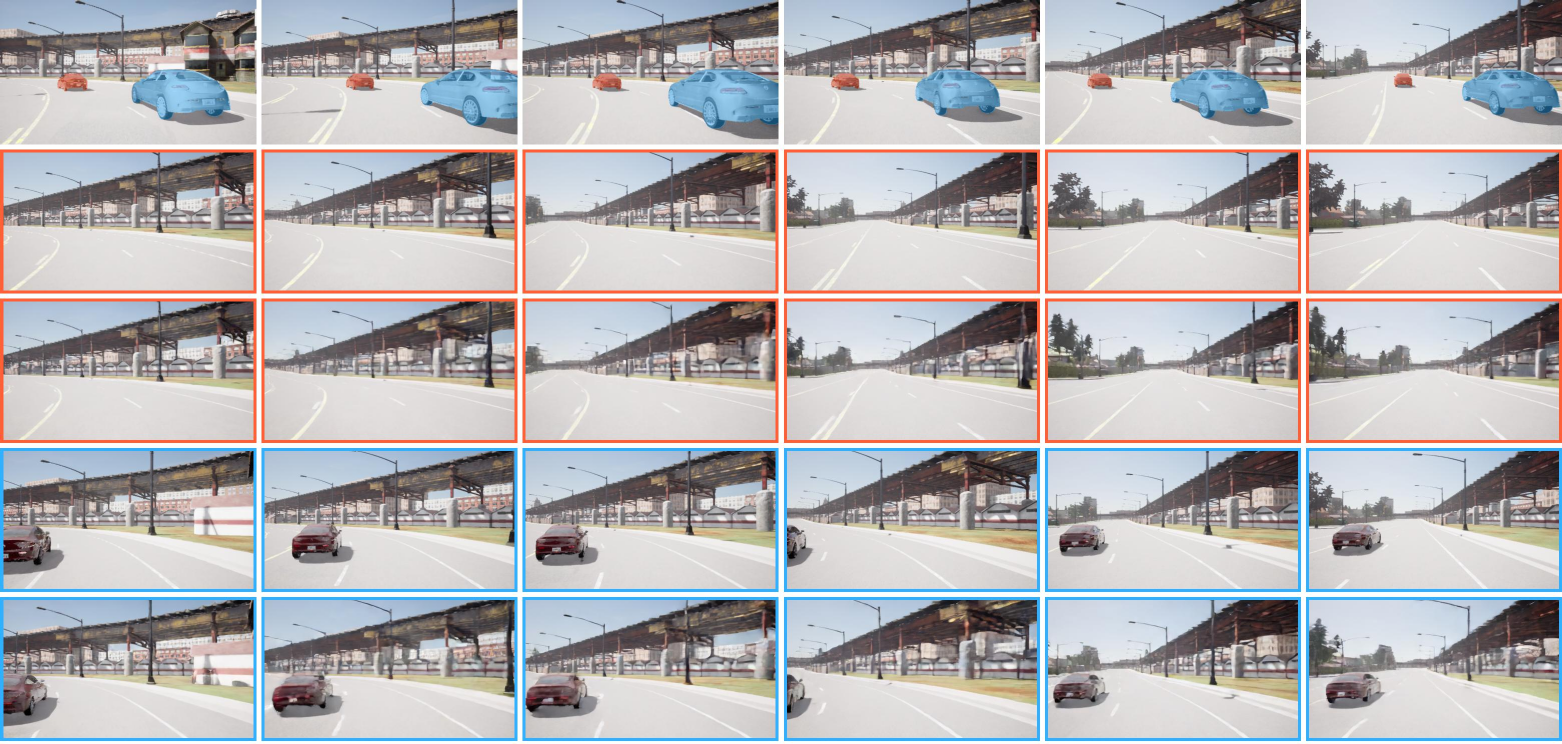}};
            \node[anchor=south, font=\fontsize{5}{6}\selectfont, rotate=90, align=center] at (-0.05, 5.70) {Observation\\(masks overlaid)};
            \node[anchor=south, font=\fontsize{5}{6}\selectfont, rotate=90, align=center] at (-0.05, 4.430) {Ground Truth};
            \node[anchor=south, font=\fontsize{5}{6}\selectfont, rotate=90, align=center] at (-0.05, 3.20) {Mirrored};
            \node[anchor=south, font=\fontsize{5}{6}\selectfont, rotate=90, align=center] at (-0.05, 1.925) {Ground Truth};
            \node[anchor=south, font=\fontsize{5}{6}\selectfont, rotate=90, align=center] at (-0.05, 0.675) {Mirrored};
        \end{tikzpicture}
        \caption{}
        % \label{fig:carla_town03_4}
    \end{subfigure}

    \caption{Qualitative mirroring results in the held-out town for MVM. In each example, the top row shows the learner egocentric view with two demonstrator attention masks overlaid in orange and blue. For each demonstrator, we show the ground-truth egocentric video and one sample from our mirror video model. (a) The orange agent takes a right turn while the blue agent drives straight, demonstrating target-specific egocentric generation. (b) MVM produces temporally consistent scene predictions by leveraging future context from the observation sequence: although the distant building is not visible until the sixth observation frame, the model anticipates its presence in the last generated frame for both agents.}
    % \label{fig:carla_town03_qualitative}
    \label{fig:carla_town03_1}
\end{figure*}

\begin{figure*}[th]
    \centering

    \begin{subfigure}{\textwidth}
        \centering
        \begin{tikzpicture}
            \node[anchor=south west, inner sep=0] (img) at (0,0)
                {\includegraphics[width=0.95\linewidth]{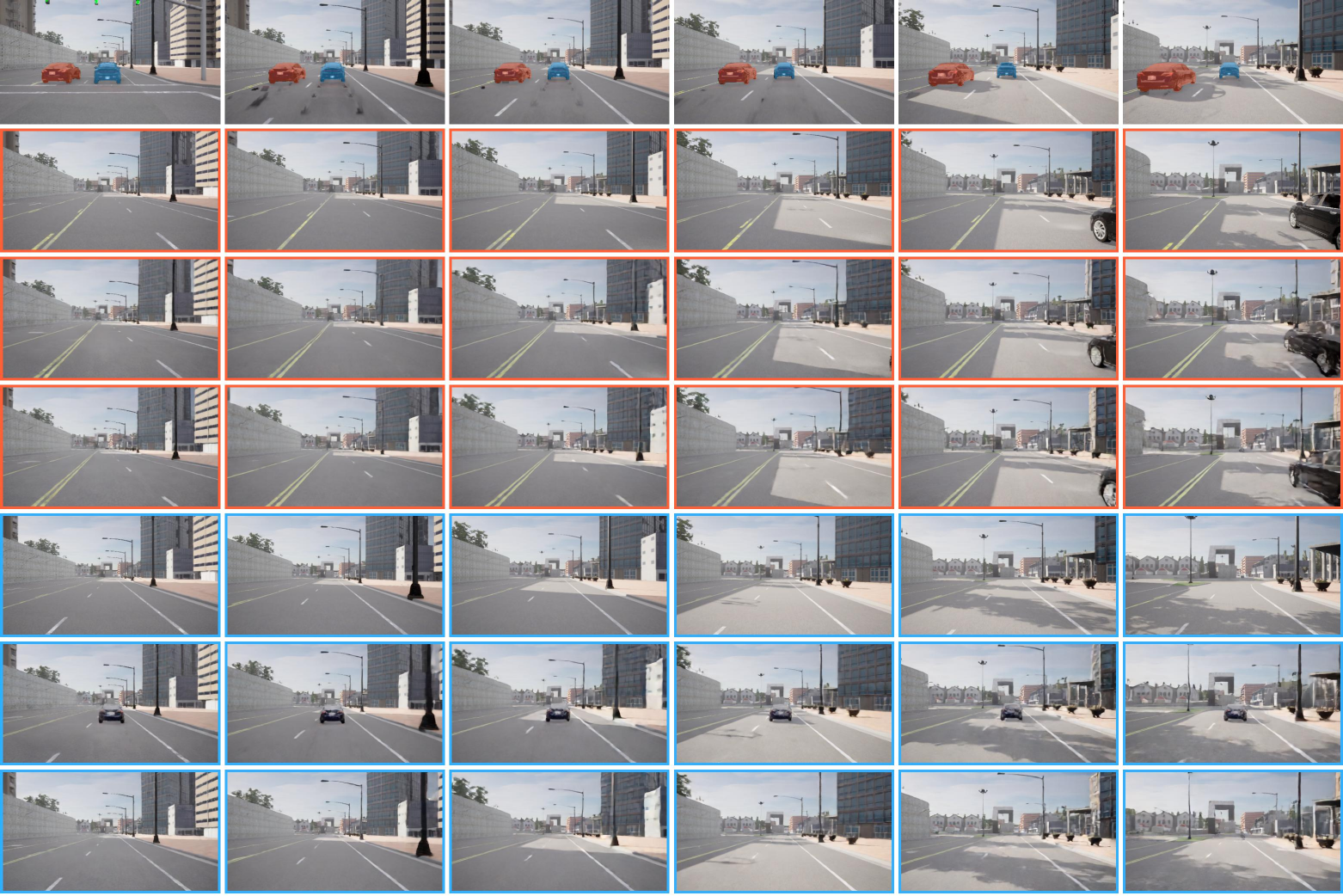}};

            \node[anchor=south, font=\fontsize{5}{6}\selectfont, rotate=90, align=center] at (-0.05, 8.25) {Observation\\(masks overlaid)};
            \node[anchor=south, font=\fontsize{5}{6}\selectfont, rotate=90, align=center] at (-0.05, 7.0) {Ground Truth};
            \node[anchor=south, font=\fontsize{5}{6}\selectfont, rotate=90, align=center] at (-0.05, 5.675) {Mirrored 1};
            \node[anchor=south, font=\fontsize{5}{6}\selectfont, rotate=90, align=center] at (-0.05, 4.425) {Mirrored 2};
            \node[anchor=south, font=\fontsize{5}{6}\selectfont, rotate=90, align=center] at (-0.05, 3.175) {Ground Truth};
            \node[anchor=south, font=\fontsize{5}{6}\selectfont, rotate=90, align=center] at (-0.05, 1.925) {Mirrored 1};
            \node[anchor=south, font=\fontsize{5}{6}\selectfont, rotate=90, align=center] at (-0.05, 0.675) {Mirrored 2};
        \end{tikzpicture}
        \caption{}
        \label{fig:carla_town03_diversity_car}
    \end{subfigure}

    \vspace{0.5em}

    \begin{subfigure}{\textwidth}
        \centering
        \begin{tikzpicture}
            \node[anchor=south west, inner sep=0] (img) at (0,0)
                {\includegraphics[width=0.95\linewidth]{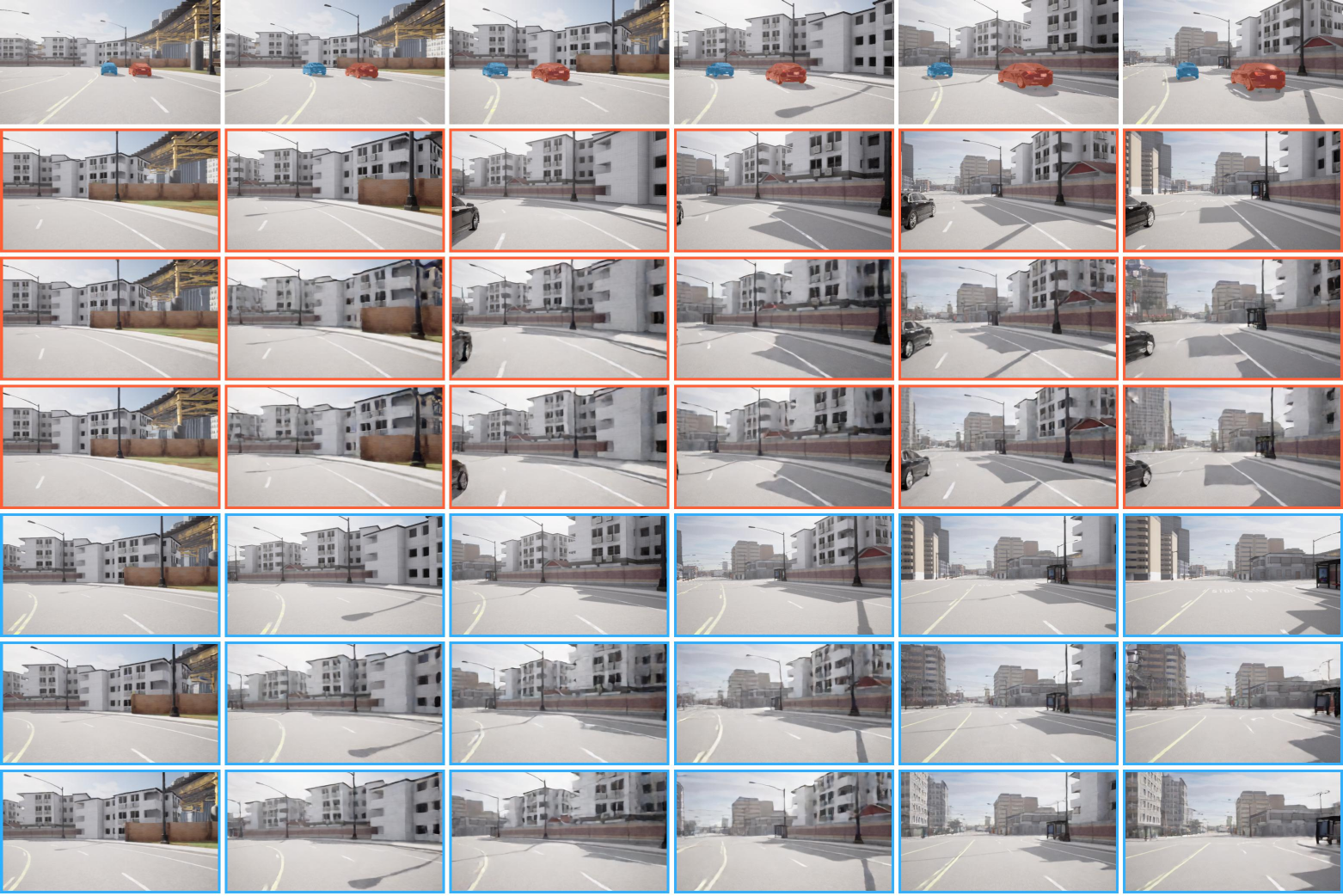}};

            \node[anchor=south, font=\fontsize{5}{6}\selectfont, rotate=90, align=center] at (-0.05, 8.25) {Observation\\(masks overlaid)};
            \node[anchor=south, font=\fontsize{5}{6}\selectfont, rotate=90, align=center] at (-0.05, 7.0) {Ground Truth};
            \node[anchor=south, font=\fontsize{5}{6}\selectfont, rotate=90, align=center] at (-0.05, 5.675) {Mirrored 1};
            \node[anchor=south, font=\fontsize{5}{6}\selectfont, rotate=90, align=center] at (-0.05, 4.425) {Mirrored 2};
            \node[anchor=south, font=\fontsize{5}{6}\selectfont, rotate=90, align=center] at (-0.05, 3.175) {Ground Truth};
            \node[anchor=south, font=\fontsize{5}{6}\selectfont, rotate=90, align=center] at (-0.05, 1.925) {Mirrored 1};
            \node[anchor=south, font=\fontsize{5}{6}\selectfont, rotate=90, align=center] at (-0.05, 0.675) {Mirrored 2};
        \end{tikzpicture}
        \caption{}
        \label{fig:carla_town03_diversity_building}
    \end{subfigure}

    \caption{Qualitative diversity results in the held-out town for MVM. In each example, the top row shows the learner egocentric view with two demonstrator attention masks overlaid in orange and blue. For each demonstrator, we show the ground-truth egocentric video and two samples from our MVM, demonstrating target-specific egocentric generation from a shared observation. (a) MVM captures uncertainty in occluded scenes by generating diverse plausible outcomes for the blue agent: one sample predicts a car ahead, whereas another does not. (b) MVM also captures uncertainty in the surrounding environment by generating diverse but plausible structures, such as buildings appearing on the left side of the frame when this information is unobserved by the learner.}
    \label{fig:carla_town03_diversity}
\end{figure*}
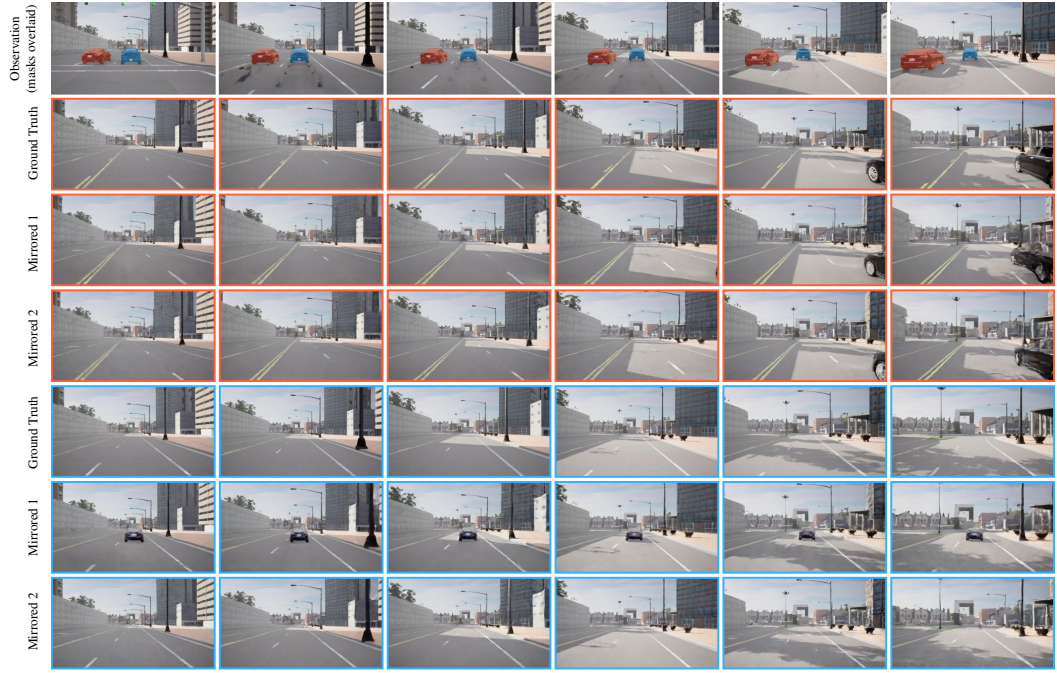
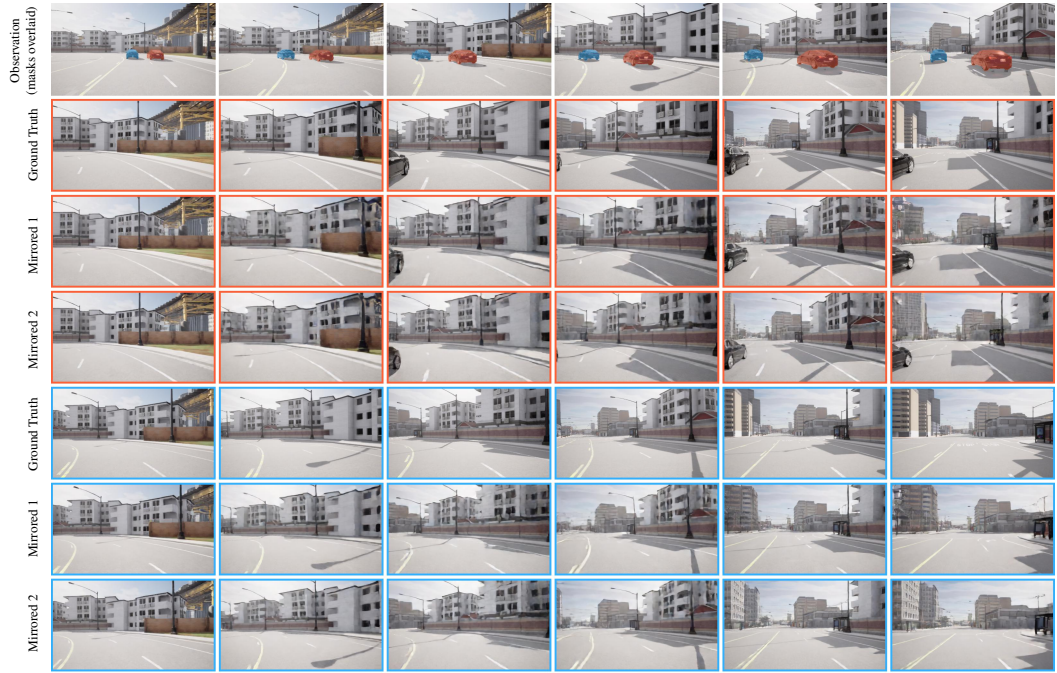

We next present qualitative results on CARLA Town 03, a held-out urban environment with large junctions and a roundabout. This map provides a challenging test of whether the mirror video model generalizes beyond the towns used for training. \Cref{fig:carla_town03_1} shows that MVM generates target-specific and temporally consistent egocentric videos in the held-out town, correctly capturing distinct agent behaviors and anticipating scene content from future observation context. \Cref{fig:carla_town03_diversity} further shows that MVM captures uncertainty under partial observability by generating diverse but plausible egocentric samples, including variations in occluded vehicles and surrounding scene structure.

% \begin{figure*}[t]
%     \centering
%     \begin{tikzpicture}
%         \node[anchor=south west, inner sep=0] (img) at (0,0)
%             {\includegraphics[width=0.93\textwidth]{figures/multi_town03_test_select/007_000.pdf}};
%         \node[anchor=center, font=\fontsize{5}{6}\selectfont, align=center, rotate=90] at (-0.25, 5.6) {Observation\\(masks overlaid)};
%         \node[anchor=center, font=\fontsize{5}{6}\selectfont,align=center, rotate=90] at (-0.25, 4.35) {Ground Truth};
%         \node[anchor=center, font=\fontsize{5}{6}\selectfont, align=center, rotate=90] at (-0.25, 3.15) {Mirrored};
%         \node[anchor=center, font=\fontsize{5}{6}\selectfont, align=center, rotate=90] at (-0.25, 1.90) {Ground Truth};
%         \node[anchor=center, font=\fontsize{5}{6}\selectfont, align=center, rotate=90] at (-0.25, 0.65) {Mirrored};
%     \end{tikzpicture}
%     \caption{Qualitative mirroring results in the Held-out town. Starting from the same observing-agent video, we select two mirror-target agents using orange and blue instance masks. For each target, we show the ground-truth egocentric video and  sampled mirror data from our mirror video model, demonstrating target-specific egocentric generation from a shared observation. In this scene, the orange agent takes a right turn while the blue agent drive straight.}

%     \label{fig:annotated_grid}
% \end{figure*}

% In preamble:
% \usepackage{subcaption}

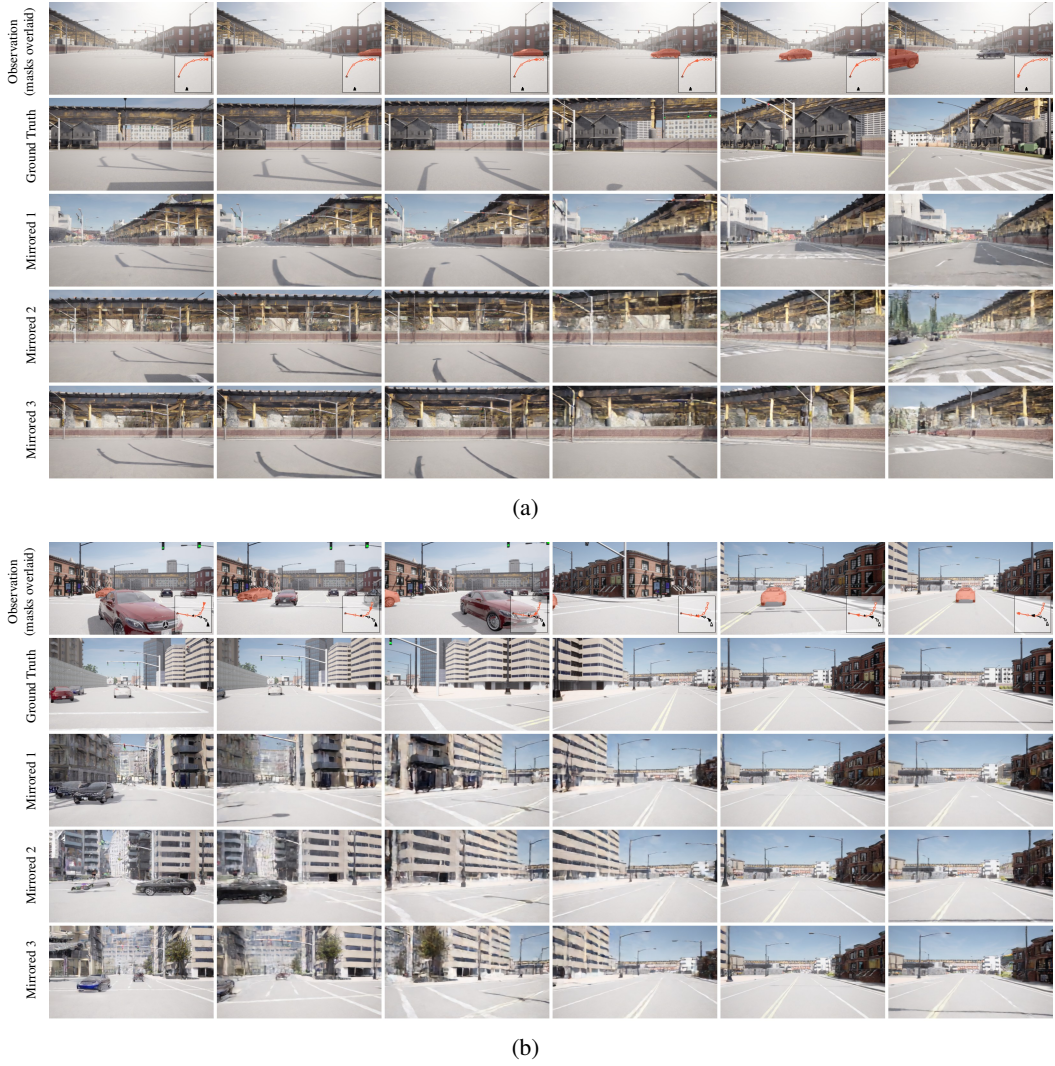
\begin{figure*}[th]
    \centering

    \begin{subfigure}{\textwidth}
        \centering
        \begin{tikzpicture}
            \node[anchor=south west, inner sep=0] (img) at (0,0)
                {\includegraphics[width=0.95\linewidth]{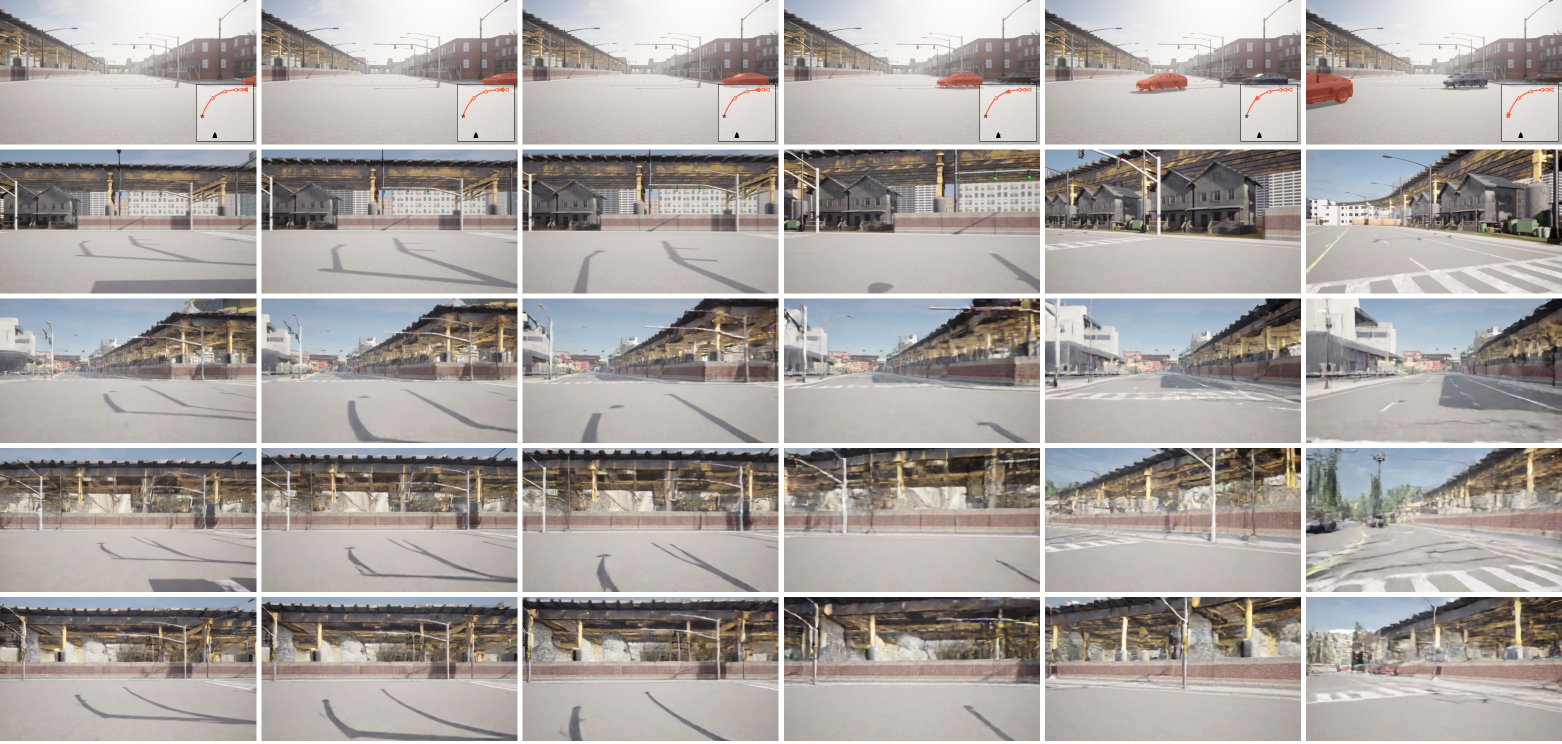}};

            \node[anchor=south, font=\fontsize{5}{6}\selectfont, rotate=90, align=center] at (-0.05, 5.675) {Observation\\(masks overlaid)};
            \node[anchor=south, font=\fontsize{5}{6}\selectfont, rotate=90, align=center] at (-0.05, 4.425) {Ground Truth};
            \node[anchor=south, font=\fontsize{5}{6}\selectfont, rotate=90, align=center] at (-0.05, 3.175) {Mirrored 1};
            \node[anchor=south, font=\fontsize{5}{6}\selectfont, rotate=90, align=center] at (-0.05, 1.925) {Mirrored 2};
            \node[anchor=south, font=\fontsize{5}{6}\selectfont, rotate=90, align=center] at (-0.05, 0.675) {Mirrored 3};
        \end{tikzpicture}
        \caption{}
        \label{fig:carla_town03_passing_1}
    \end{subfigure}

    \vspace{0.5em}

    \begin{subfigure}{\textwidth}
        \centering
        \begin{tikzpicture}
            \node[anchor=south west, inner sep=0] (img) at (0,0)
                {\includegraphics[width=0.95\linewidth]{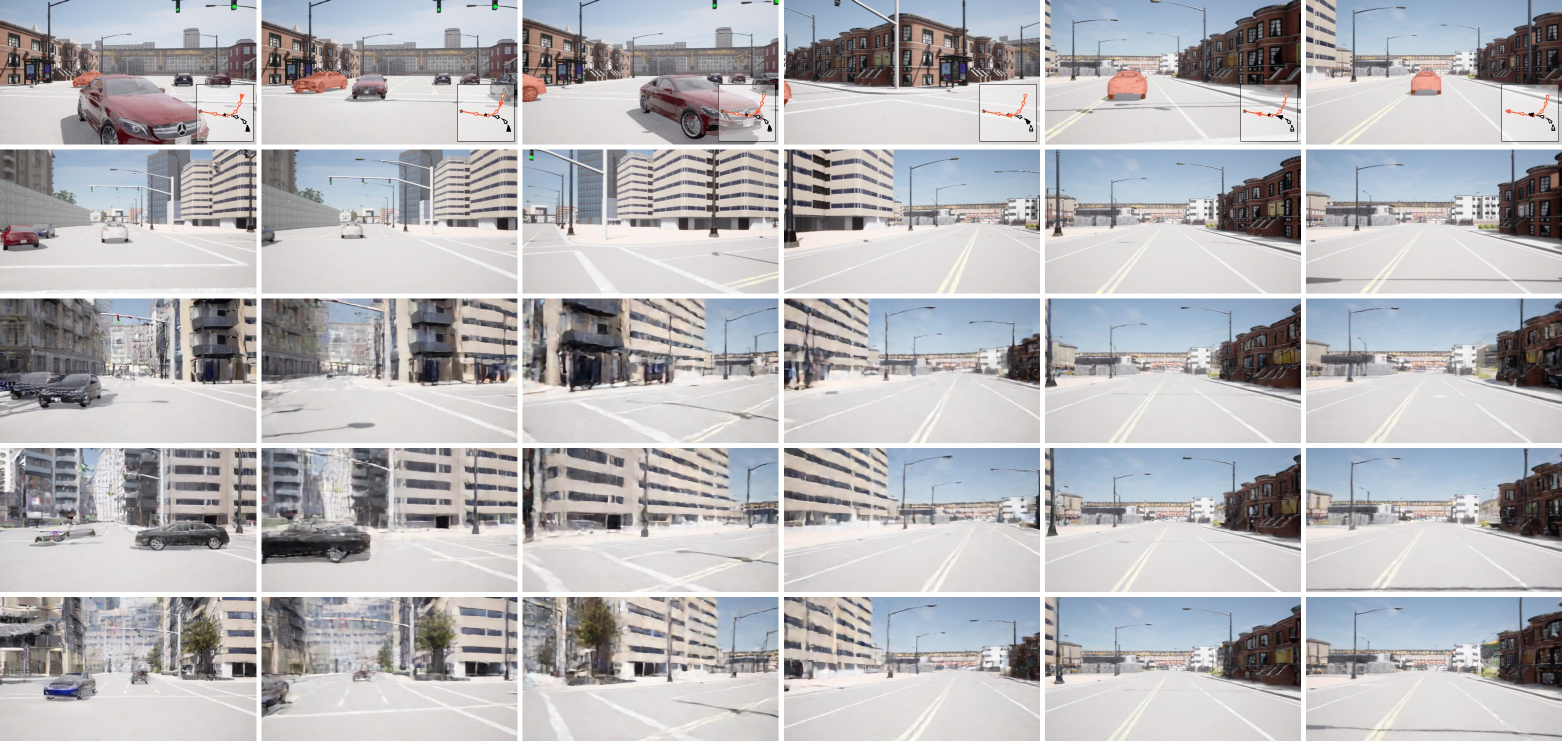}};

            \node[anchor=south, font=\fontsize{5}{6}\selectfont, rotate=90, align=center] at (-0.05, 5.675) {Observation\\(masks overlaid)};
            \node[anchor=south, font=\fontsize{5}{6}\selectfont, rotate=90, align=center] at (-0.05, 4.425) {Ground Truth};
            \node[anchor=south, font=\fontsize{5}{6}\selectfont, rotate=90, align=center] at (-0.05, 3.175) {Mirrored 1};
            \node[anchor=south, font=\fontsize{5}{6}\selectfont, rotate=90, align=center] at (-0.05, 1.925) {Mirrored 2};
            \node[anchor=south, font=\fontsize{5}{6}\selectfont, rotate=90, align=center] at (-0.05, 0.675) {Mirrored 3};
        \end{tikzpicture}
        \caption{}
        \label{fig:carla_town03_passing_2}
    \end{subfigure}

    \caption{Qualitative passing-diversity results in the held-out town. In these challenging pass-by scenarios, the learner observes the demonstrator from a different viewpoint, with limited overlap with the demonstrator's front-facing view. To clarify the interaction geometry, we additionally show a bird's-eye-view trajectory subplot alongside the learner's observations, oriented to the learner's heading in the first frame. The dark triangle denotes the learner, and the orange triangle denotes the demonstrator. (a) Given the learner's observation and the demonstrator's instance mask, the mirror video model predicts the correct left-turn behavior across multiple sampled egocentric videos. (b) In this scenario, the learner makes a left turn while the demonstrator turns right in the opposite direction. The two agents have limited observation overlap at the beginning of the scene, leading to diverse predicted backgrounds; once the learner completes the turn and the views become more aligned, the predicted scene becomes consistent.}
    \label{fig:carla_town03_passing}
\end{figure*}

\FloatBarrier
\subsubsection{May Mobility Zero-Shot Transfer}
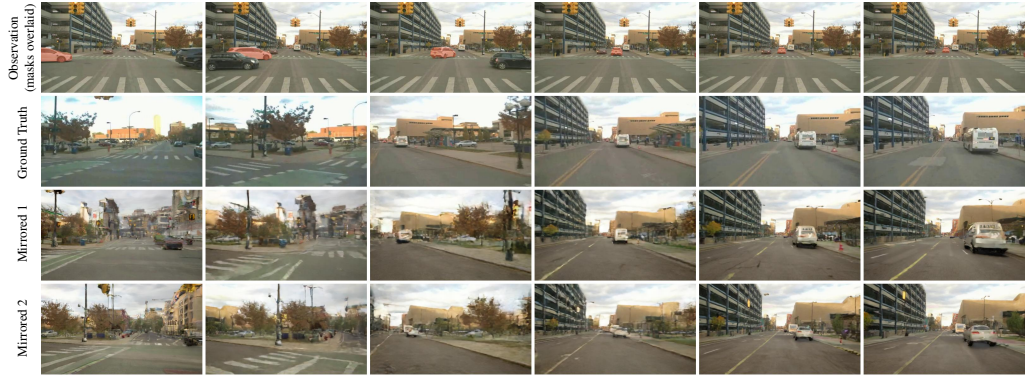
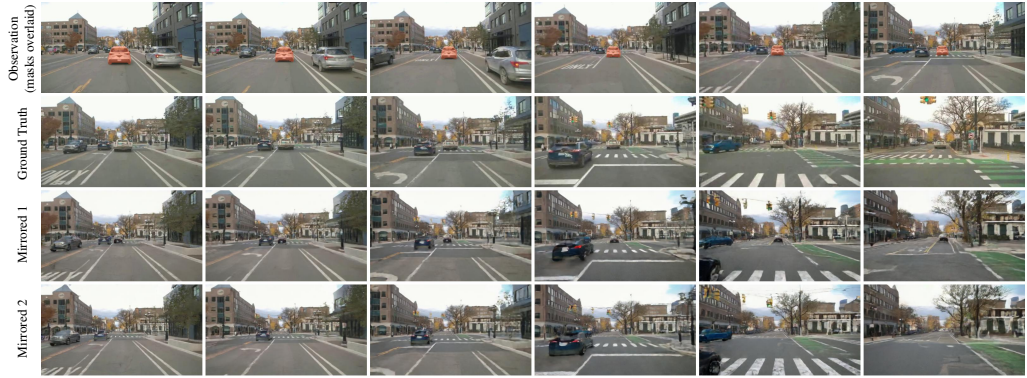
\begin{figure*}[th]
    \centering
    \captionsetup[subfigure]{font=small, justification=centering}

    \begin{subfigure}{\textwidth}
        \centering
        \begin{tikzpicture}
            \node[anchor=south west, inner sep=0] (img) at (0,0)
                {\includegraphics[width=0.93\textwidth]{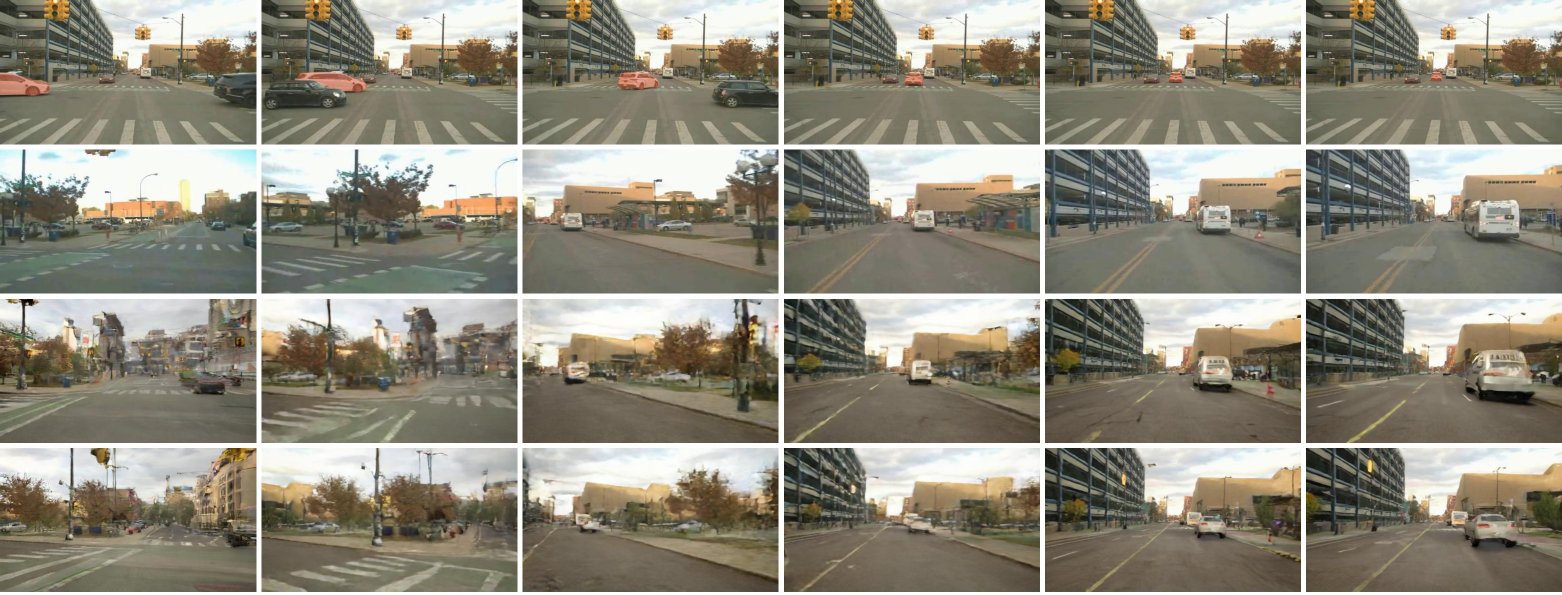}};
            \node[anchor=center, font=\fontsize{5}{6}\selectfont, align=center, rotate=90] at (-0.25, 4.35) {Observation\\(masks overlaid)};
            \node[anchor=center, font=\fontsize{5}{6}\selectfont, align=center, rotate=90] at (-0.25, 3.10) {Ground Truth};
            \node[anchor=center, font=\fontsize{5}{6}\selectfont, align=center, rotate=90] at (-0.25, 1.90) {Mirrored 1};
            \node[anchor=center, font=\fontsize{5}{6}\selectfont, align=center, rotate=90] at (-0.25, 0.65) {Mirrored 2};
        \end{tikzpicture}
        \caption{The demonstrator makes a left turn while partially occluded by a black Mini, yet MVM predicts consistent demonstrator-centric egocentric views.}
        \label{fig:mars_1}
    \end{subfigure}

    \vspace{0.4em}

    \begin{subfigure}{\textwidth}
        \centering
        \begin{tikzpicture}
            \node[anchor=south west, inner sep=0] (img) at (0,0)
                {\includegraphics[width=0.93\textwidth]{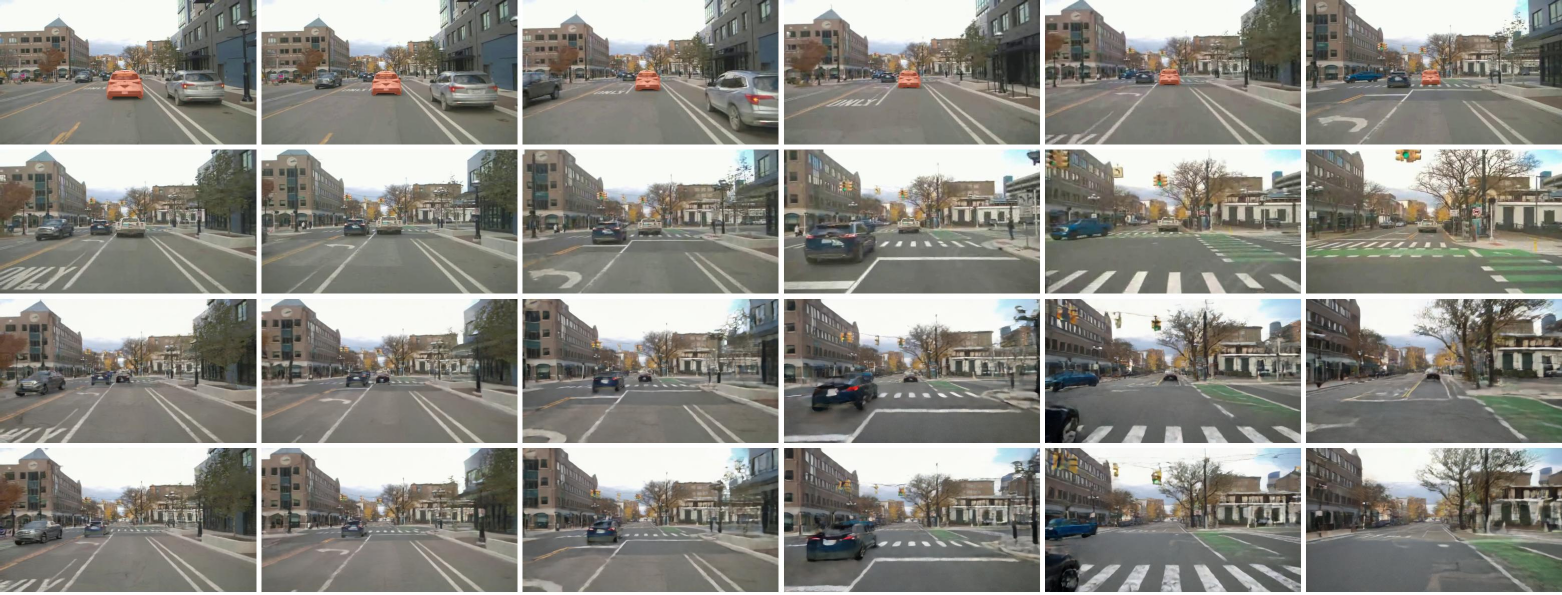}};
            \node[anchor=center, font=\fontsize{5}{6}\selectfont, align=center, rotate=90] at (-0.25, 4.35) {Observation\\(masks overlaid)};
            \node[anchor=center, font=\fontsize{5}{6}\selectfont, align=center, rotate=90] at (-0.25, 3.10) {Ground Truth};
            \node[anchor=center, font=\fontsize{5}{6}\selectfont, align=center, rotate=90] at (-0.25, 1.90) {Mirrored 1};
            \node[anchor=center, font=\fontsize{5}{6}\selectfont, align=center, rotate=90] at (-0.25, 0.65) {Mirrored 2};
        \end{tikzpicture}
        \caption{MVM generates plausible real-world demonstrator-centric views from learner observations and SAM3 instance masks.}
        \label{fig:mars_3}
    \end{subfigure}

    \caption{Qualitative zero-shot mirroring results on real-world data using our mirror video model trained only in the CARLA simulator. Given the learner's observations and the demonstrator instance mask generated by SAM3, we show the demonstrator's egocentric ground-truth video and two mirrored samples generated by our model.}
    \label{fig:mars_qualitative_occlusion}
\end{figure*}

\begin{figure*}[t]
    \centering
    \captionsetup[subfigure]{font=small, justification=centering}

    \begin{subfigure}{\textwidth}
        \centering
        \begin{tikzpicture}
            \node[anchor=south west, inner sep=0] (img) at (0,0)
                {\includegraphics[width=0.93\textwidth]{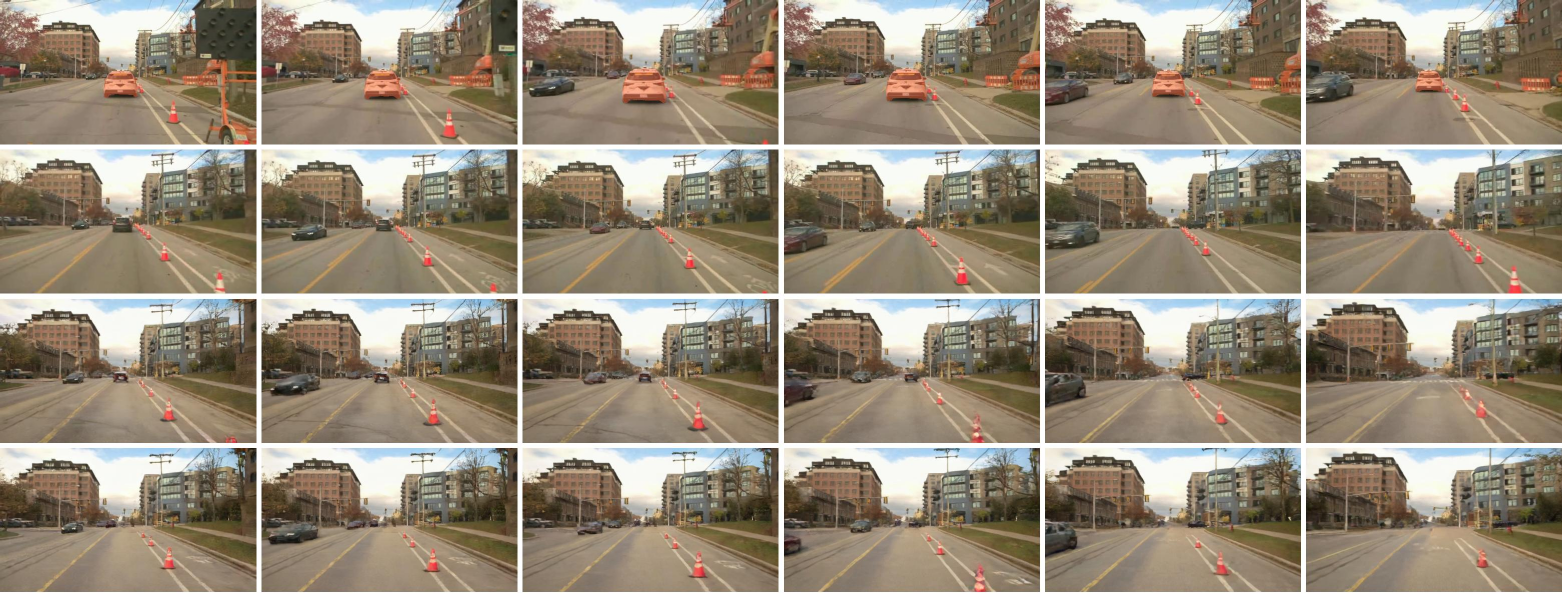}};
            \node[anchor=center, font=\fontsize{5}{6}\selectfont, align=center, rotate=90] at (-0.25, 4.35) {Observation\\(masks overlaid)};
            \node[anchor=center, font=\fontsize{5}{6}\selectfont, align=center, rotate=90] at (-0.25, 3.10) {Ground Truth};
            \node[anchor=center, font=\fontsize{5}{6}\selectfont, align=center, rotate=90] at (-0.25, 1.90) {Mirrored 1};
            \node[anchor=center, font=\fontsize{5}{6}\selectfont, align=center, rotate=90] at (-0.25, 0.65) {Mirrored 2};
        \end{tikzpicture}
        \caption{MVM transfers to a real-world scene containing traffic cones, which are absent from the CARLA training data.}
        \label{fig:mars_4}
    \end{subfigure}

    \vspace{0.4em}

    \begin{subfigure}{\textwidth}
        \centering
        \begin{tikzpicture}
            \node[anchor=south west, inner sep=0] (img) at (0,0)
                {\includegraphics[width=0.93\textwidth]{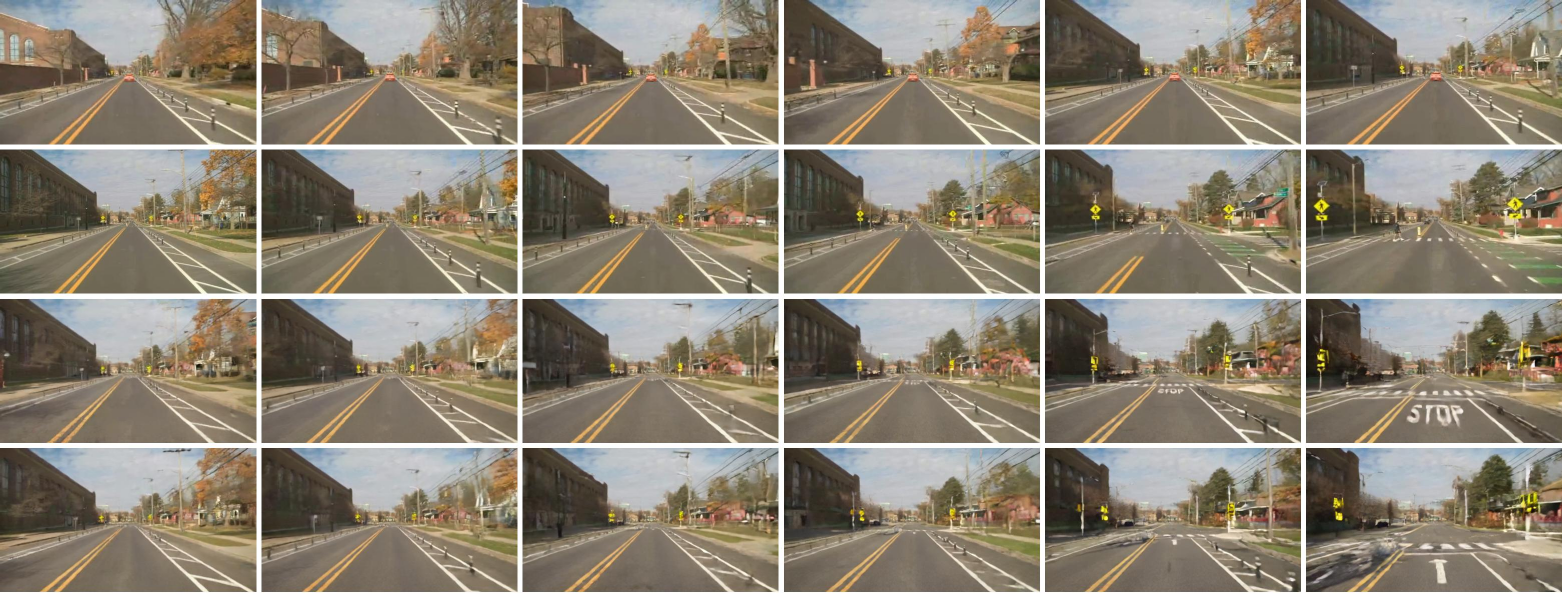}};
            \node[anchor=center, font=\fontsize{5}{6}\selectfont, align=center, rotate=90] at (-0.25, 4.35) {Observation\\(masks overlaid)};
            \node[anchor=center, font=\fontsize{5}{6}\selectfont, align=center, rotate=90] at (-0.25, 3.10) {Ground Truth};
            \node[anchor=center, font=\fontsize{5}{6}\selectfont, align=center, rotate=90] at (-0.25, 1.90) {Mirrored 1};
            \node[anchor=center, font=\fontsize{5}{6}\selectfont, align=center, rotate=90] at (-0.25, 0.65) {Mirrored 2};
        \end{tikzpicture}
        \caption{MVM generates diverse plausible road markings, including stop text and a straight-ahead arrow.}
        \label{fig:mars_5}
    \end{subfigure}

    \vspace{0.4em}

    \begin{subfigure}{\textwidth}
        \centering
        \begin{tikzpicture}
            \node[anchor=south west, inner sep=0] (img) at (0,0)
                {\includegraphics[width=0.93\textwidth]{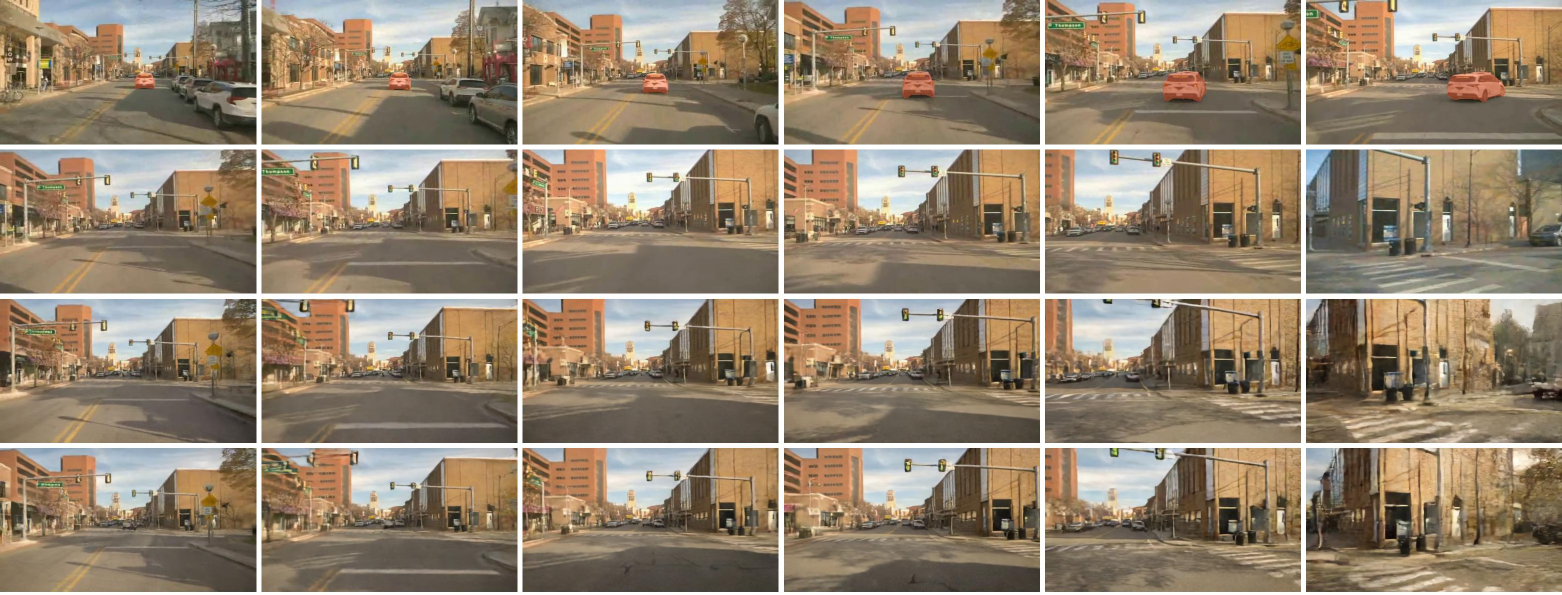}};
            \node[anchor=center, font=\fontsize{5}{6}\selectfont, align=center, rotate=90] at (-0.25, 4.35) {Observation\\(masks overlaid)};
            \node[anchor=center, font=\fontsize{5}{6}\selectfont, align=center, rotate=90] at (-0.25, 3.10) {Ground Truth};
            \node[anchor=center, font=\fontsize{5}{6}\selectfont, align=center, rotate=90] at (-0.25, 1.90) {Mirrored 1};
            \node[anchor=center, font=\fontsize{5}{6}\selectfont, align=center, rotate=90] at (-0.25, 0.65) {Mirrored 2};
        \end{tikzpicture}
        \caption{MVM infers the demonstrator's right-turn behavior from real-world learner observations.}
        \label{fig:mars_2}
    \end{subfigure}

    \caption{Qualitative zero-shot mirroring results on real-world data using our mirror video model trained only in the CARLA simulator. Given the learner's observations and the demonstrator instance mask generated by SAM3, we show the demonstrator's egocentric ground-truth video and two mirrored samples generated by our model.}
    \label{fig:mars_qualitative}
\end{figure*}
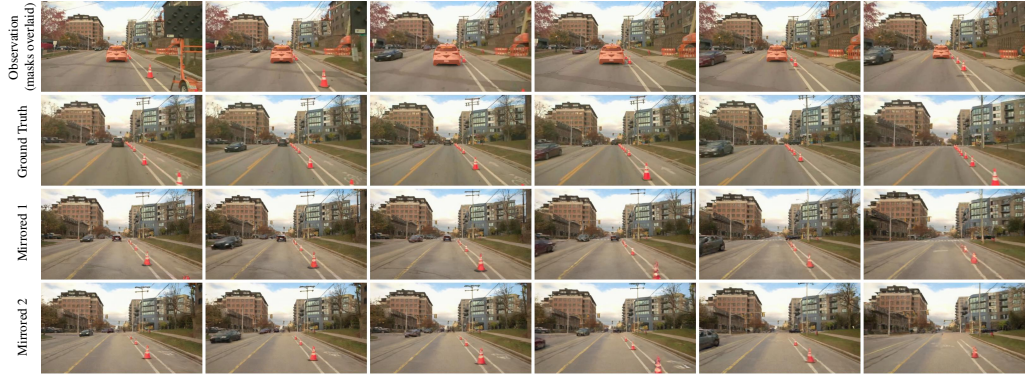
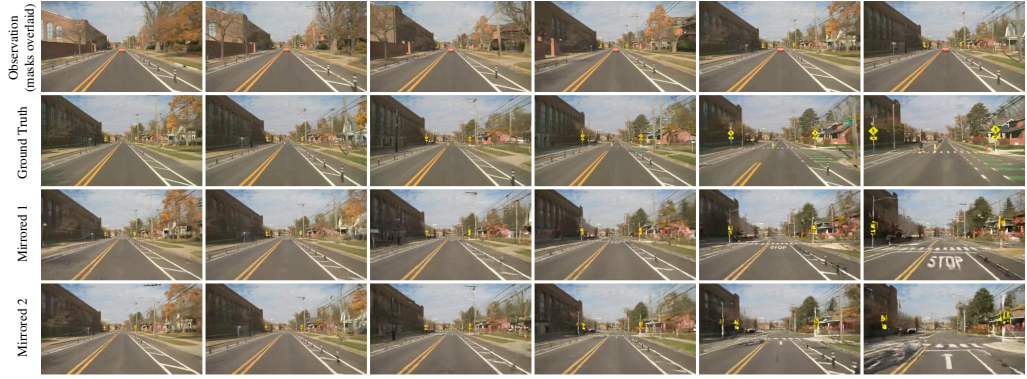
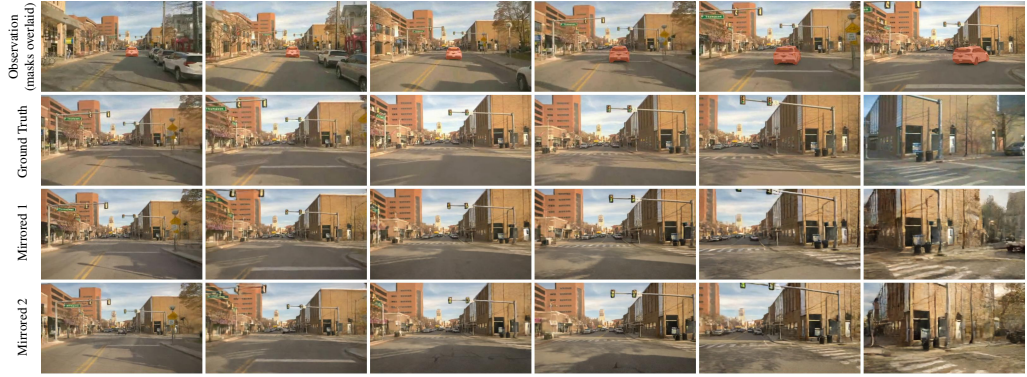

% 2024\_03\_06\_scene\_4\_clip0.mp4 \\
% 2023\_12\_08\_scene\_2\_clip0.mp4 mask need to be improved \\
% 2023\_12\_15\_scene\_3\_clip0.mp4 Night \\
% 2023\_12\_13\_scene\_8\_clip0.mp4 Night \\
% 2023\_12\_08\_scene\_1\_clip0.mp4 \\
% 2023\_12\_07\_scene\_1\_clip0.mp4 far \\
% 2023\_11\_09\_scene\_4\_clip1.mp4  \\
% 2023\_11\_09\_scene\_3\_clip0.mp4  \\
% 2023\_11\_08\_scene\_2\_clip0.mp4  \\
% 2023\_11\_02\_scene\_13\_clip0.mp4 \\
% 2023\_11\_02\_scene\_12\_clip0.mp4 \\
% 2023\_11\_02\_scene\_11\_clip0. \\
% 2023\_11\_02\_scene\_10\_clip0.mp4 \\
% 2023\_11\_02\_scene\_4\_clip0.mp4 \\
% 2023\_11\_02\_scene\_1\_clip0. \\
% 2023\_11\_01\_scene\_5\_clip0.mp4 \\
% 2023\_11\_01\_scene\_4\_clip1.mp4 \\
% 2023\_10\_30\_scene\_3\_clip0.mp4 \\
% 2023\_10\_30\_scene\_2\_clip0. \\
% 2023\_10\_30\_scene\_1\_clip0.mp4
We show qualitative zero-shot transfer results on the May Mobility dataset described above in Figures~\ref{fig:mars_qualitative_occlusion}--\ref{fig:mars_qualitative}. Target-agent masks are obtained with SAM~3~\citep{carion2025sam3segmentconcepts} together with the dataset-specific selection procedure described in the dataset section.

% In preamble:
% \usepackage{subcaption}

\clearpage
\FloatBarrier
\subsubsection{Minecraft Evaluation}
\paragraph{Experiment-Specific Setup.}
Minecraft experiments use the Plaicraft training and validation splits described above. We train and evaluate the mirror video model on 25-frame windows extracted from 3-second clips. Below, we show quantitative results and qualitative results on the held-out validation split. Additional qualitative results on the training split in the additional results section.

\begin{table}[th]
    \centering
    \caption{Quantitative evaluation on the held out Minecraft dataset for the mirror video model. Lower FVD is better; higher PSNR and SSIM are better.}
    \label{tab:minecraft-metrics}
    \small
    \begin{tabular}{lccc}
        \toprule
        \textbf{FVD} $\downarrow$ & \textbf{PSNR} $\uparrow$ & \textbf{SSIM} $\uparrow$ \\
        \midrule
        227.60 & 16.42 & 0.45 \\
        \bottomrule
    \end{tabular}
\end{table}

\paragraph{Quantitative Results}
To facilitate reproducibility and provide baseline metrics for future work on 
mirror video modeling, we report quantitative results for the mirror video model in~\cref{tab:minecraft-metrics}. Results are computed over 750 validation samples from the validation set. We report FVD, PSNR, and SSIM, where lower FVD indicates better video distribution fidelity and higher PSNR/SSIM indicate better frame-level reconstruction quality.

\paragraph{Qualitative results on Validation-Set}
\label[appendix]{sec:appendix:plaicraft-examples}

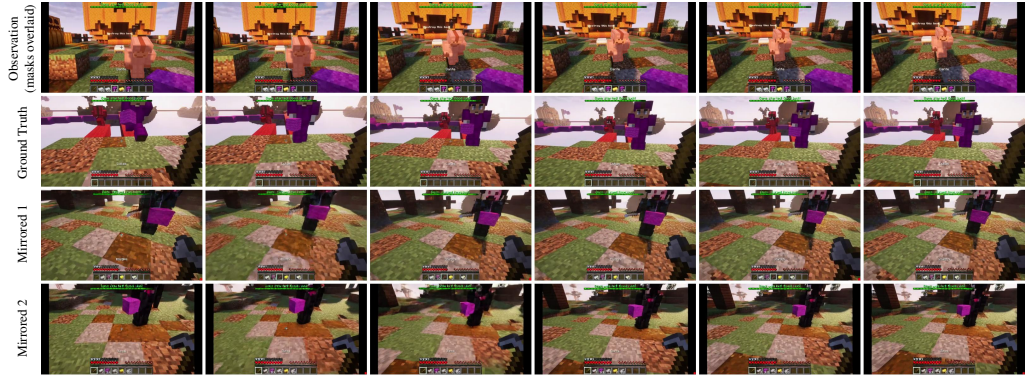
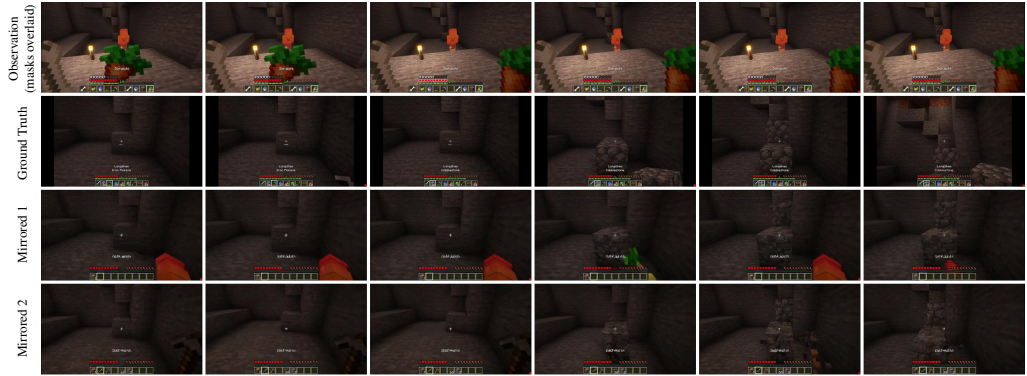
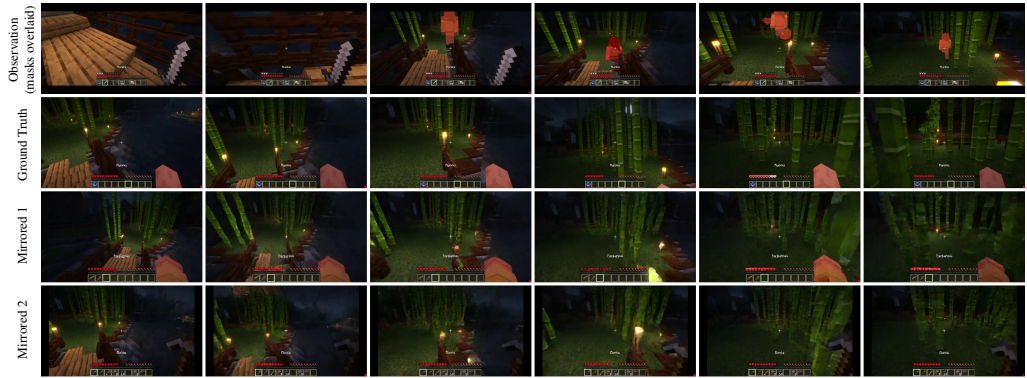
\begin{figure*}[t]
    \centering
    \captionsetup[subfigure]{font=small, justification=centering}

    \begin{subfigure}{\textwidth}
        \centering
        \begin{tikzpicture}
            \node[anchor=south west, inner sep=0] (img) at (0,0)
                {\includegraphics[width=0.93\textwidth]{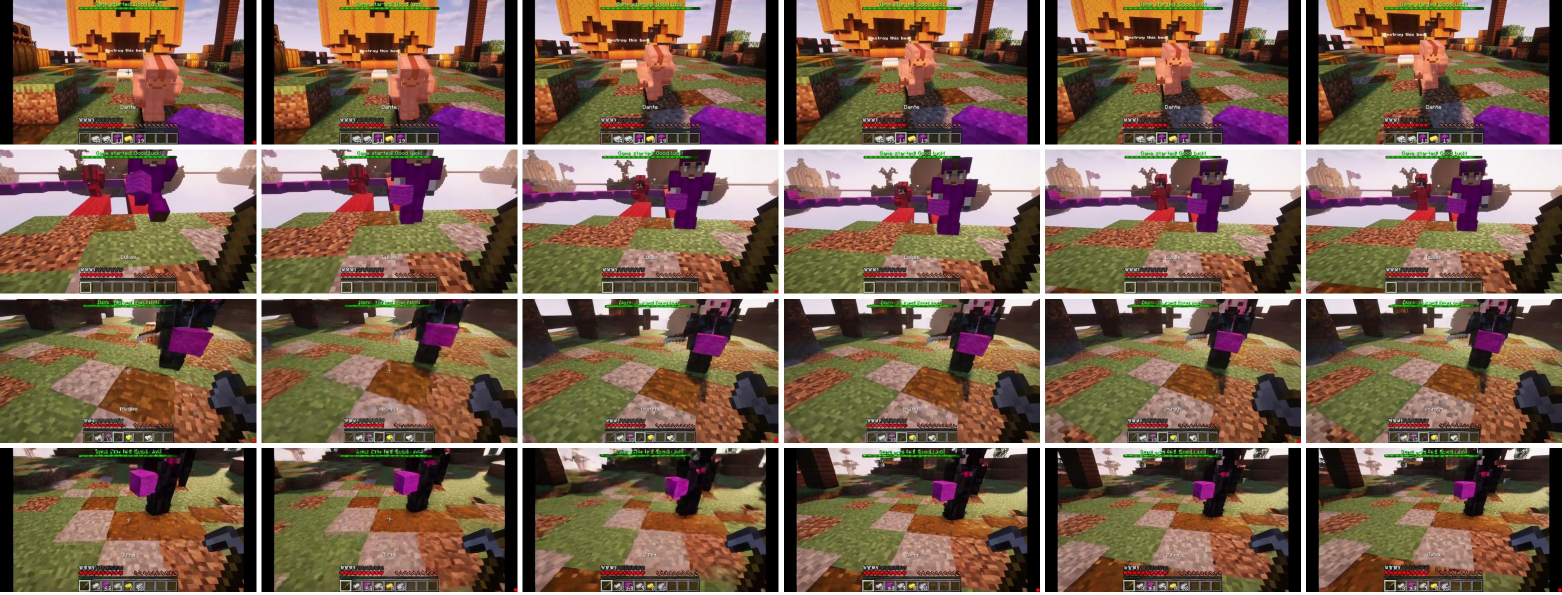}};
            \node[anchor=center, font=\fontsize{5}{6}\selectfont, align=center, rotate=90] at (-0.25, 4.35) {Observation\\(masks overlaid)};
            \node[anchor=center, font=\fontsize{5}{6}\selectfont, align=center, rotate=90] at (-0.25, 3.10) {Ground Truth};
            \node[anchor=center, font=\fontsize{5}{6}\selectfont, align=center, rotate=90] at (-0.25, 1.90) {Mirrored 1};
            \node[anchor=center, font=\fontsize{5}{6}\selectfont, align=center, rotate=90] at (-0.25, 0.65) {Mirrored 2};
        \end{tikzpicture}
        \caption{Mirror video model generation captures how the observing agent is holding a purple block.}
    \end{subfigure}

    \vspace{0.4em}

    \begin{subfigure}{\textwidth}
        \centering
        \begin{tikzpicture}
            \node[anchor=south west, inner sep=0] (img) at (0,0)
                {\includegraphics[width=0.93\textwidth]{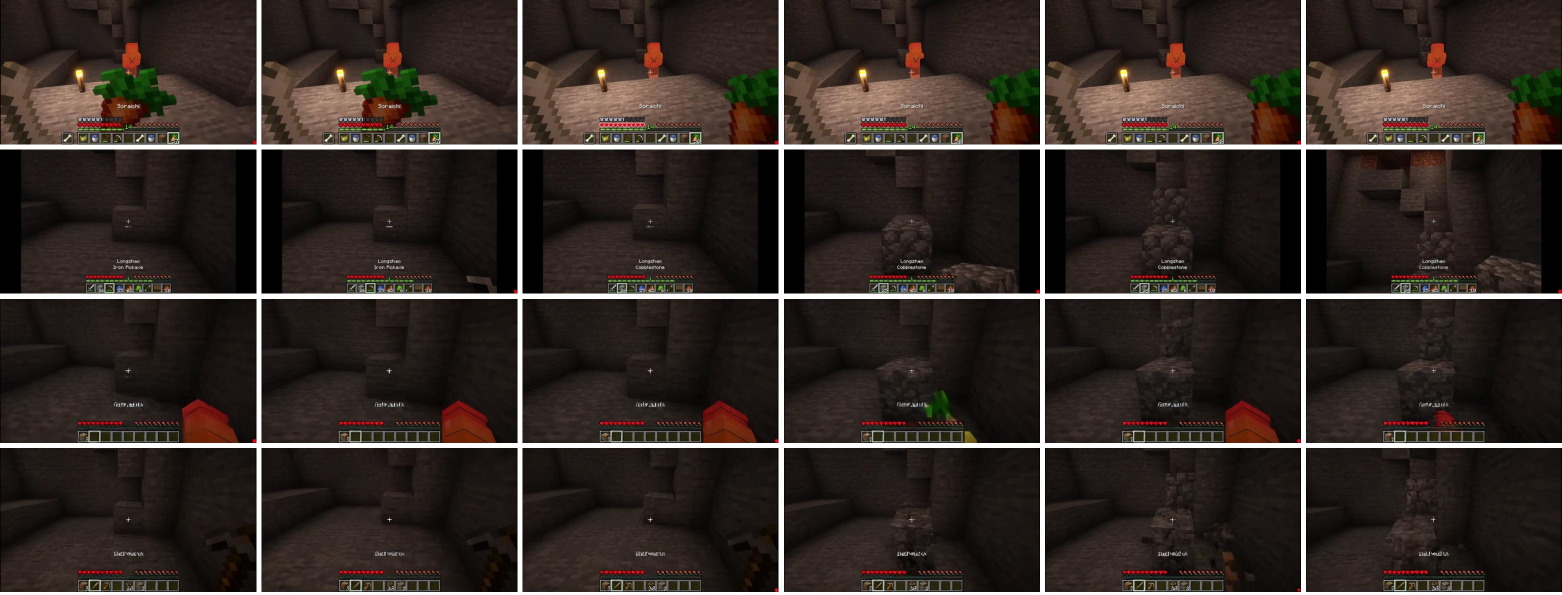}};
            \node[anchor=center, font=\fontsize{5}{6}\selectfont, align=center, rotate=90] at (-0.25, 4.35) {Observation\\(masks overlaid)};
            \node[anchor=center, font=\fontsize{5}{6}\selectfont, align=center, rotate=90] at (-0.25, 3.10) {Ground Truth};
            \node[anchor=center, font=\fontsize{5}{6}\selectfont, align=center, rotate=90] at (-0.25, 1.90) {Mirrored 1};
            \node[anchor=center, font=\fontsize{5}{6}\selectfont, align=center, rotate=90] at (-0.25, 0.65) {Mirrored 2};
        \end{tikzpicture}
        \caption{Mirror video model is able to generate a first-person view of a player engaged in block placing action.}
    \end{subfigure}

    \vspace{0.4em}

    \begin{subfigure}{\textwidth}
        \centering
        \begin{tikzpicture}
            \node[anchor=south west, inner sep=0] (img) at (0,0)
                {\includegraphics[width=0.93\textwidth]{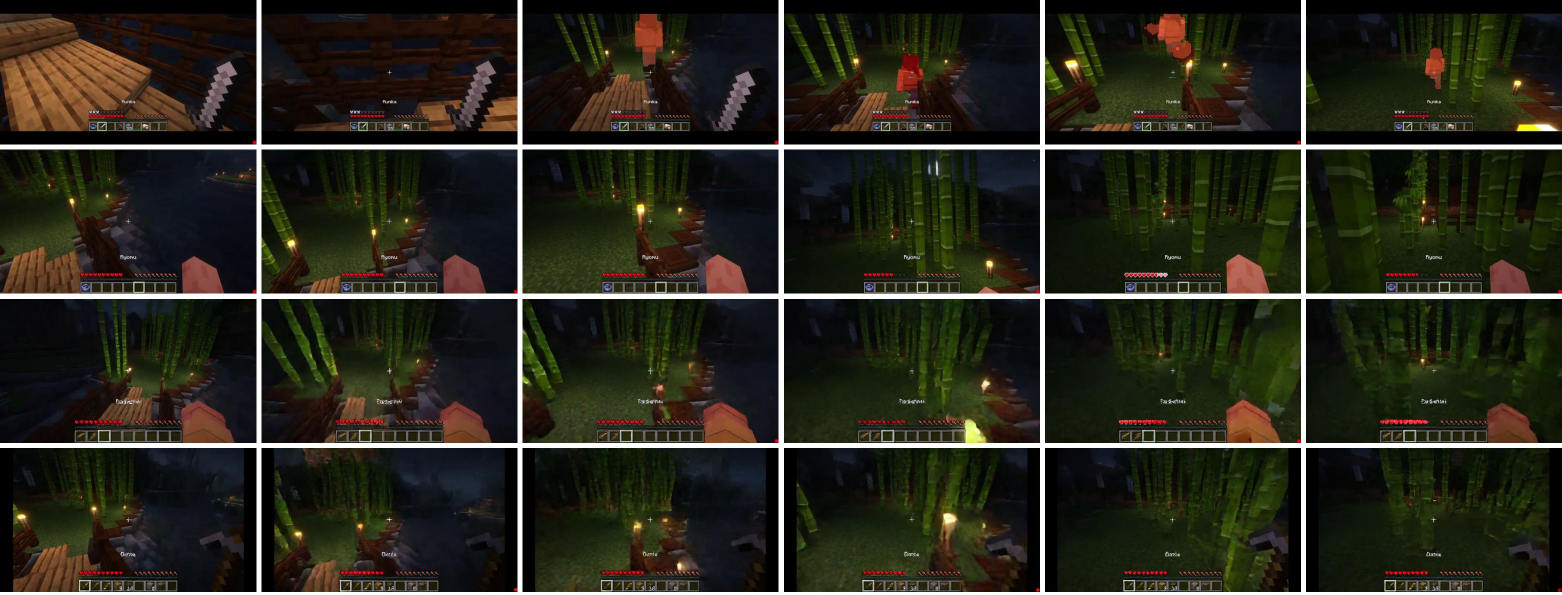}};
            \node[anchor=center, font=\fontsize{5}{6}\selectfont, align=center, rotate=90] at (-0.25, 4.35) {Observation\\(masks overlaid)};
            \node[anchor=center, font=\fontsize{5}{6}\selectfont, align=center, rotate=90] at (-0.25, 3.10) {Ground Truth};
            \node[anchor=center, font=\fontsize{5}{6}\selectfont, align=center, rotate=90] at (-0.25, 1.90) {Mirrored 1};
            \node[anchor=center, font=\fontsize{5}{6}\selectfont, align=center, rotate=90] at (-0.25, 0.65) {Mirrored 2};
        \end{tikzpicture}
        \caption{Mirror video model captures first-person view of running through passable bamboo blocks.}
    \end{subfigure}

     \caption{Validation set qualitative mirroring results for a scenario the mirror network has not been trained on. We select a target agent using orange instance masks. For the target, we show the ground-truth egocentric video and two sampled predictions from the perspective transfer network, demonstrating egocentric generation from a shared observation.}
    \label{fig:plaicraft_val1a}
\end{figure*}

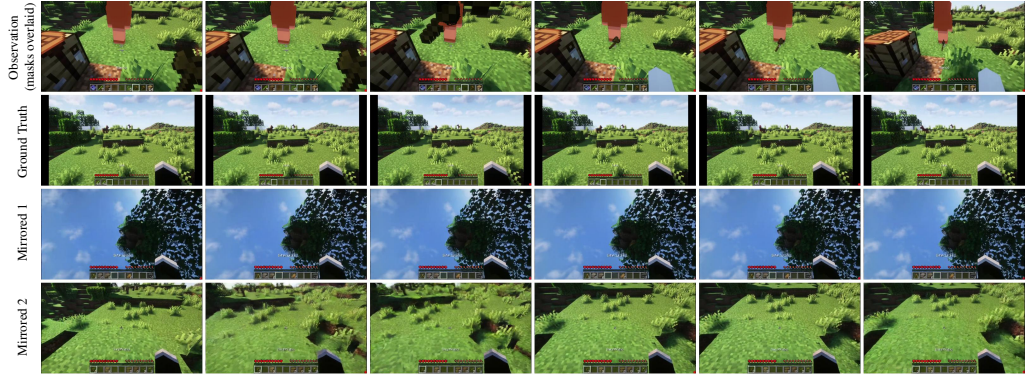
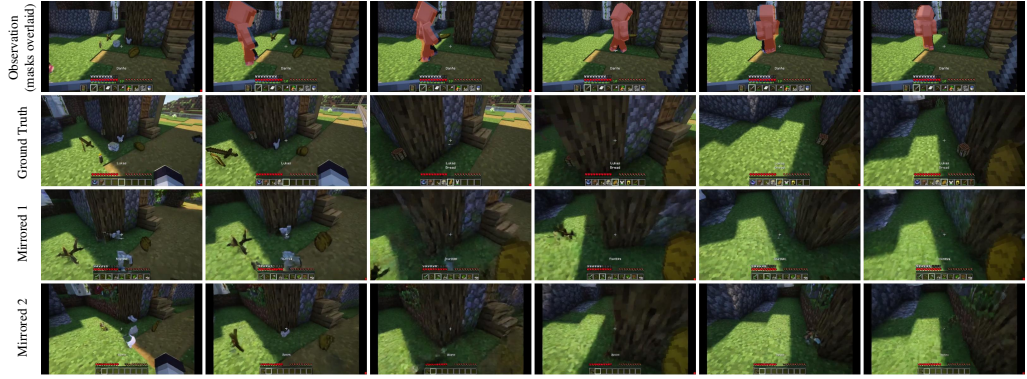
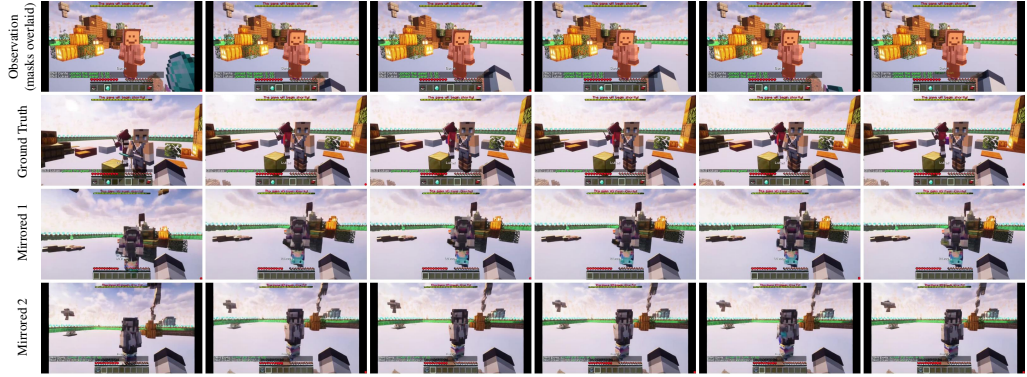
\begin{figure*}[t]
    \centering
    \captionsetup[subfigure]{font=small, justification=centering}

    \begin{subfigure}{\textwidth}
        \centering
        \begin{tikzpicture}
            \node[anchor=south west, inner sep=0] (img) at (0,0)
                {\includegraphics[width=0.93\textwidth]{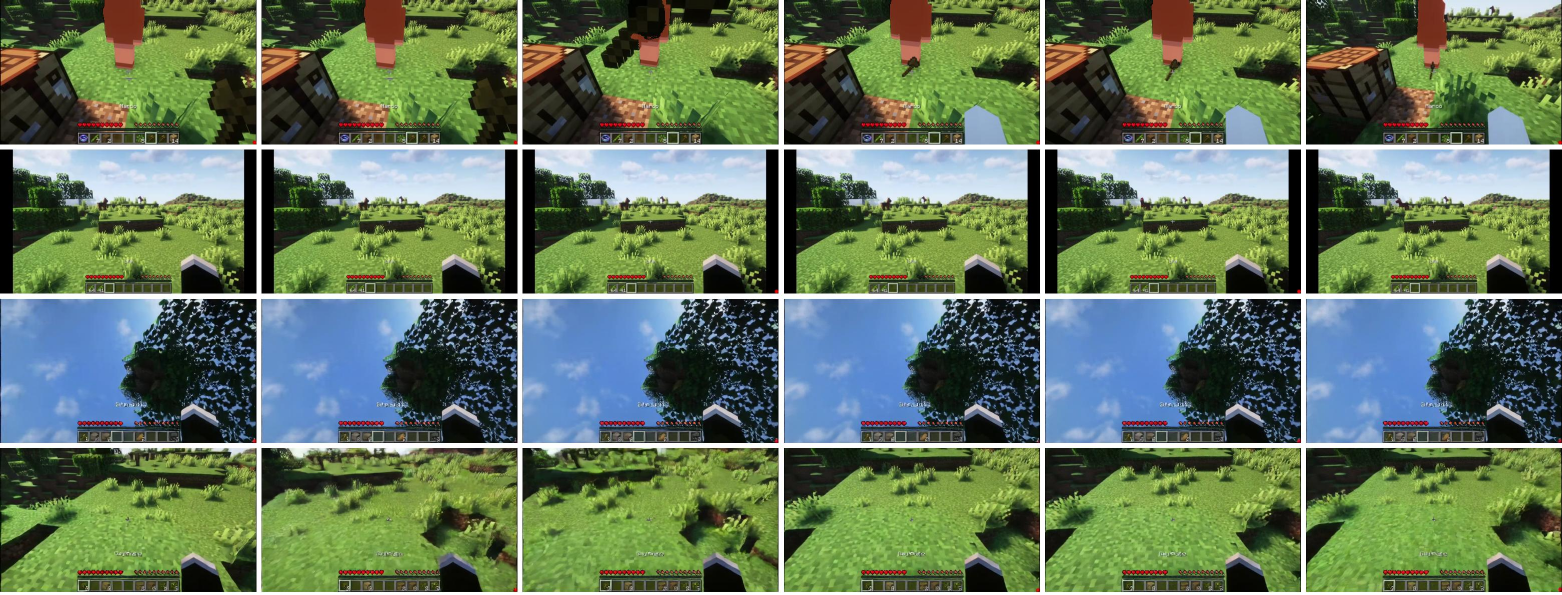}};
            \node[anchor=center, font=\fontsize{5}{6}\selectfont, align=center, rotate=90] at (-0.25, 4.35) {Observation\\(masks overlaid)};
            \node[anchor=center, font=\fontsize{5}{6}\selectfont, align=center, rotate=90] at (-0.25, 3.10) {Ground Truth};
            \node[anchor=center, font=\fontsize{5}{6}\selectfont, align=center, rotate=90] at (-0.25, 1.90) {Mirrored 1};
            \node[anchor=center, font=\fontsize{5}{6}\selectfont, align=center, rotate=90] at (-0.25, 0.65) {Mirrored 2};
        \end{tikzpicture}
        \caption{Mindful of the ambiguity over the head orientation of the observed player, the mirror video model generates first-person views that looks at different directions.}
    \end{subfigure}

    \vspace{0.4em}

    \begin{subfigure}{\textwidth}
        \centering
        \begin{tikzpicture}
            \node[anchor=south west, inner sep=0] (img) at (0,0)
                {\includegraphics[width=0.93\textwidth]{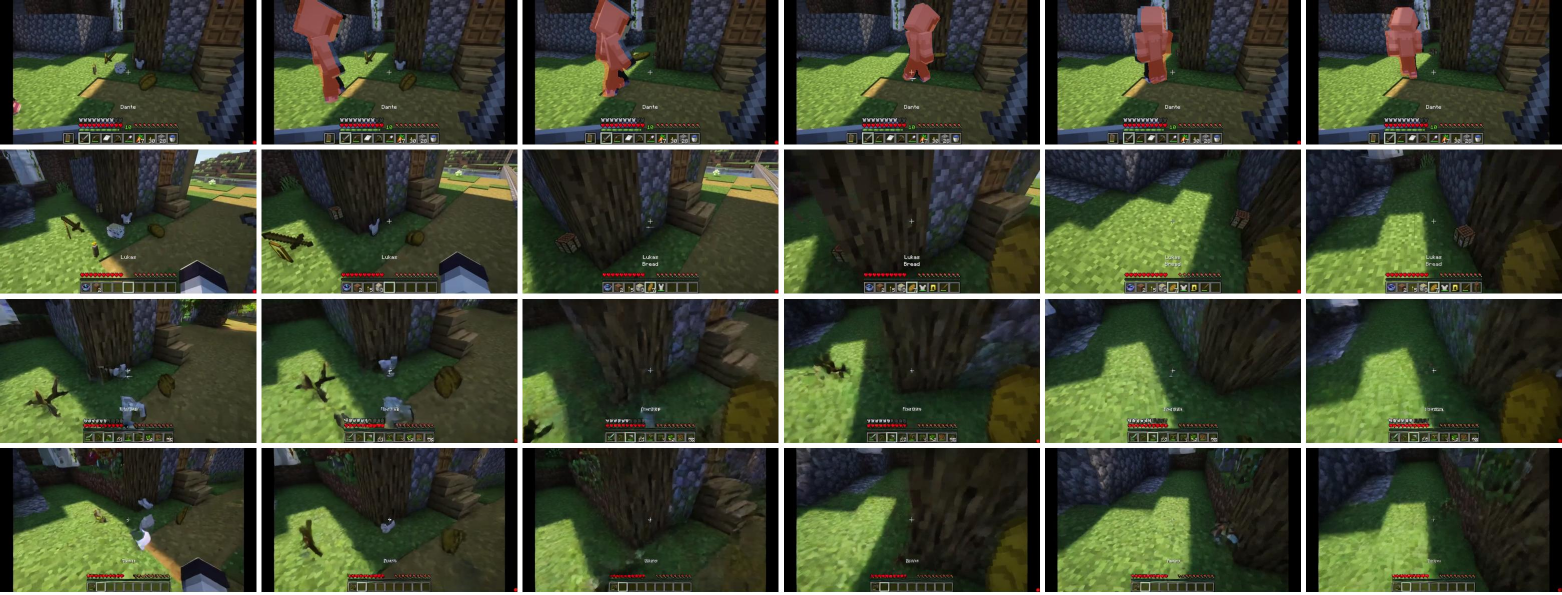}};
            \node[anchor=center, font=\fontsize{5}{6}\selectfont, align=center, rotate=90] at (-0.25, 4.35) {Observation\\(masks overlaid)};
            \node[anchor=center, font=\fontsize{5}{6}\selectfont, align=center, rotate=90] at (-0.25, 3.10) {Ground Truth};
            \node[anchor=center, font=\fontsize{5}{6}\selectfont, align=center, rotate=90] at (-0.25, 1.90) {Mirrored 1};
            \node[anchor=center, font=\fontsize{5}{6}\selectfont, align=center, rotate=90] at (-0.25, 0.65) {Mirrored 2};
        \end{tikzpicture}
        \caption{Mirror video model renders fine-grained details like floating items.}
    \end{subfigure}

    \vspace{0.4em}

    \begin{subfigure}{\textwidth}
        \centering
        \begin{tikzpicture}
            \node[anchor=south west, inner sep=0] (img) at (0,0)
                {\includegraphics[width=0.93\textwidth]{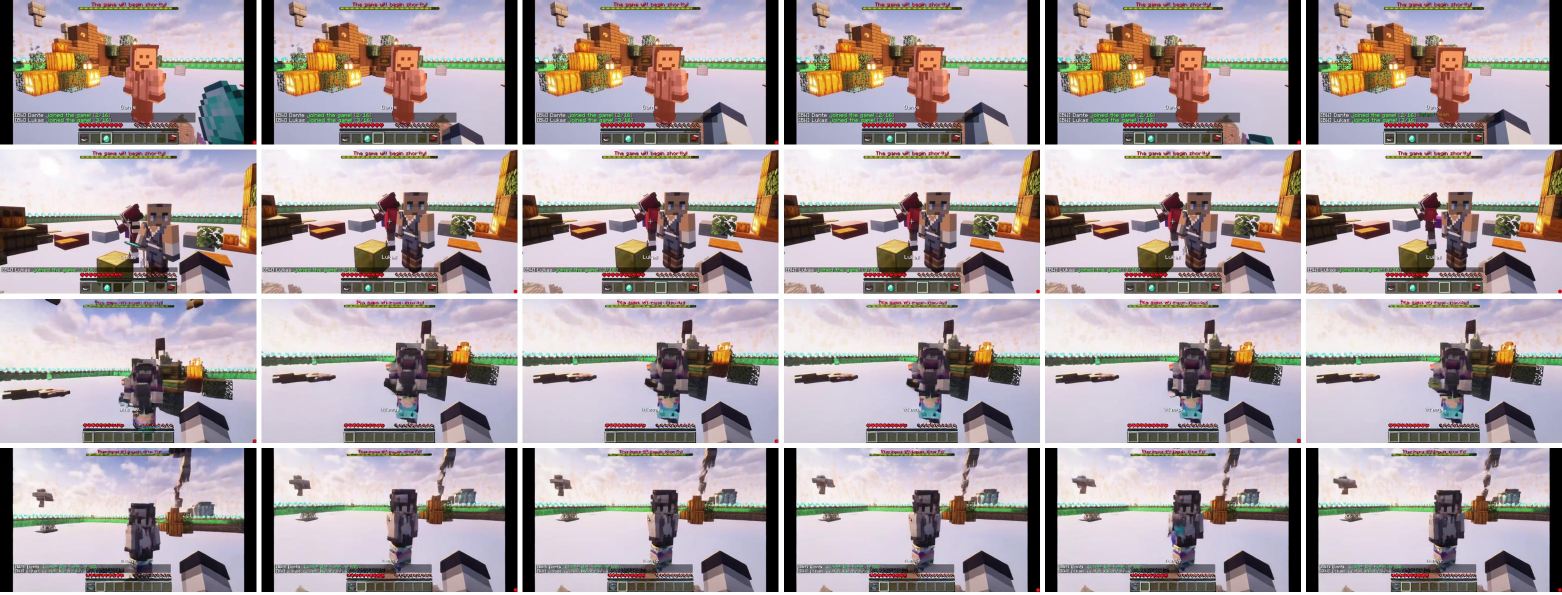}};
            \node[anchor=center, font=\fontsize{5}{6}\selectfont, align=center, rotate=90] at (-0.25, 4.35) {Observation\\(masks overlaid)};
            \node[anchor=center, font=\fontsize{5}{6}\selectfont, align=center, rotate=90] at (-0.25, 3.10) {Ground Truth};
            \node[anchor=center, font=\fontsize{5}{6}\selectfont, align=center, rotate=90] at (-0.25, 1.90) {Mirrored 1};
            \node[anchor=center, font=\fontsize{5}{6}\selectfont, align=center, rotate=90] at (-0.25, 0.65) {Mirrored 2};
        \end{tikzpicture}
        \caption{Mirror video model renders diverse observing player appearances in its generations.}
    \end{subfigure}

     \caption{Validation set qualitative mirroring results for a scenario the mirror network has not been trained on. We select a target agent using orange instance masks. For the target, we show the ground-truth egocentric video and two sampled predictions from the perspective transfer network, demonstrating egocentric generation from a shared observation.}
    \label{fig:plaicraft_val2a}
\end{figure*}

\Cref{fig:plaicraft_val1a,fig:plaicraft_val2a} show qualitative mirroring results of our MWM model on Minecraft validation set. Compared to the CARLA dataset, Minecraft presents a more challenging setting because the relative camera poses between the learner and demonstrator are much more flexible and unconstrained. As a result, the learner and demonstrator often observe each other from viewpoints with limited perspective overlap, for example when the two players are simply facing one another. This partial observability leads to higher diversity across generated samples. Nevertheless, MVM can exploit visual cues in the learner observation, such as the demonstrator's pose and held weapon, to generate plausible demonstrator-centric egocentric views.

\clearpage
\FloatBarrier

\subsection{Mirror Learning Experiments}
\label[appendix]{sec:appendix:mirror_learning_experiments}
This section provides supplementary details for the downstream mirror learning results in the main paper. Specifically, it documents the experimental setups used for Figure~\ref{fig:mirror-learning-results} and Table~\ref{tab:mirror-as-fraction-of-total}, and concludes with qualitative examples for the style-transfer baseline used in the geographic-expansion experiment. Details of our CARLA dataset split are provided in the ``Split Rationale'' paragraph of~\cref{sec:appendix:CARLA-data-collection}.

\paragraph{Figure~5(a): Mirror-Only Behavior Cloning.}

For the mirror-only behavior cloning experiment in Figure~\ref{fig:mirror-learning-results}(a), we train the stock VaVAM architecture, conditioned on six history frames at 2~Hz, to predict three seconds of future actions given a driving command. We train policy variants at three dataset scales using two data sources: ground-truth first-person demonstrations and mirror data.
Ground-truth first-person demonstrations are subsampled from the MWM training split. Mirror data is generated from the disjoint mirror-data-generation split in Town~04, Town~06, and Town~10. As described in the dataset split rationale in~\cref{sec:appendix:CARLA-data-collection}, the CARLA dataset is split at the traffic-scenario level, so each scenario appears in exactly one split. Within each town, weather and other scene conditions vary across scenarios. Learner observations used for mirror-data generation are excluded from both MWM and IDM training. We train both the mirror video model and the IDM on the MWM training split, and the IDM used for this experiment achieves an average displacement error of 0.42 on the MWM validation split. Across dataset scales, we subsample with a fixed random seed so that the town distribution is matched as closely as possible.

% We evaluate all policies using open-loop minADE$_5$ on the test split in the non-held-out towns with 10{,}294 segments constructed from scenarios and spatial regions that are separated from or underrepresented in the main training distribution. Unless otherwise noted, we use the same learning rate across all data scales and train each policy until convergence.

We evaluate all policies using open-loop minADE$_5$ on the test split in the non-held-out towns, which contains 10{,}294 segments constructed from scenarios and spatial regions that are separated from or underrepresented in the main training distribution. Unless otherwise noted, we use the same learning rate across all data scales and train each policy until convergence.

\paragraph{Figure~5(b): Geographic Expansion.}

For the geographic-expansion experiment in~\cref{tab:mirror-vs-other-augmentation}, we start from the pretrained policy trained on 27.9k first-person ground-truth samples, corresponding to the third row of \cref{tab:mirror-as-fraction-of-total}, and fine-tune it with a learning rate of $10^{-6}$ on either 1k or 2k additional ground-truth samples from the held-out town, Town~03. Within Town~03, the adaptation data and evaluation data are split at the traffic scenario level, so no scenario is shared between fine-tuning and evaluation. We compare three augmentation choices: no augmentation, style-transfer augmentation, and mirror-data augmentation. For augmentation with mirror data, we keep the mirror video model, pretrained on the MWM training split from the non-held-out towns, fixed without further training or fine-tuning on Town~03. We pair it with an IDM pretrained on the MWM training split and further fine-tuned on the additional first-person ground-truth data from Town~03 to generate extra behavior-cloning samples from learner videos in the held-out town. The fine-tuned IDM achieves an average displacement error of 0.67 when fine-tuned on 1k Town~03 first-person ground-truth samples and 0.66 when fine-tuned on 2k samples. For style transfer, we preserve the same scene layout and ego trajectory by conditioning on the original edge structure while varying the text prompt; examples are shown in~\cref{sec:appendix:style-transfer-results}.

\paragraph{Table~1: Fixed-Budget Data Mixtures.}
% For \cref{tab:mirror-as-fraction-of-total}, we keep the overall training budget approximately fixed at 33.6k segments and vary the amount of mirror data that replaces ground-truth data. The mirror data is generated from learner observations in the Valid Large split using a mirror video model and inverse dynamics model trained on Train Large. The IDM used for this experiment has an average displacement error of 0.42 measured on a held-out set within Valid Large. Policies are evaluated both on \texttt{test\_1} in the same towns and zero-shot on held-out \texttt{Town03}. This experiment isolates whether mirror data is useful as a fraction of the total training set, rather than only as an additive augmentation source. Mirror video model and inverse dynamics model trained on Train Large. The IDM used for this experiment has an average displacement error of 0.42 measured on a held-out set within Valid Large.

For \cref{tab:mirror-as-fraction-of-total}, we keep the overall training budget approximately fixed at 33.6k segments and vary the amount of mirror data that replaces ground-truth data. The mirror data is generated from learner observations in the disjoint mirror-data-generation split in Town~04, Town~06, and Town~10 using a mirror video model and IDM trained on the MWM training split. The IDM used for this experiment achieves an average displacement error of 0.42 on the MWM validation split. Policies are evaluated both on the test split in the non-held-out towns and zero-shot on the held-out town, Town~03. This experiment isolates whether mirror data is useful as a fraction of the total training set, rather than only as an additive augmentation source.

\subsubsection{Style-Transfer Baseline Examples}
\label[appendix]{sec:appendix:style-transfer-results}
\label{sec:style_transfer}

We apply simulation-to-real style transfer to CARLA clips using Cosmos-Transfer2~\citep{azzolini2025cosmos}, a 2B-parameter video diffusion model conditioned on structural control signals. Edge maps extracted from simulator frames are used as layout conditioning, preserving road geometry, building placement, and vehicle layout while allowing photorealistic appearance changes. We run inference at 720p for 93-frame windows with 35 diffusion steps and guidance scale 3, cycling text prompts for diverse lighting and weather conditions. Prompts are generated by a large language model.

\begin{figure*}[p]
    \centering
    \captionsetup[subfigure]{font=scriptsize, justification=centering}

    \begin{subfigure}{\textwidth}
        \centering
        \begin{tikzpicture}
            \node[anchor=south west, inner sep=0] (img) at (0,0)
                {\includegraphics[width=0.93\textwidth]{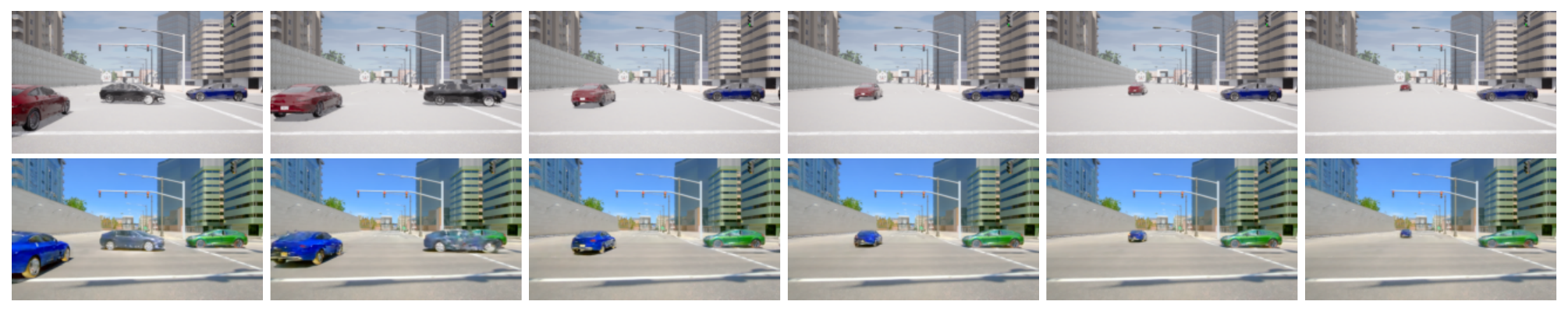}};
            \node[anchor=center, font=\fontsize{5}{6}\selectfont, align=center, rotate=90] at (-0.25, 1.90) {Ground Truth};
            \node[anchor=center, font=\fontsize{5}{6}\selectfont, align=center, rotate=90] at (-0.25, 0.675) {Style Transfer};
        \end{tikzpicture}
        \caption{\emph{A first-person driving simulator scene in the sparkling light of a clear spring day with a gentle breeze, with an assortment of brightly colored cars in every direction. Maintaining a steady forward motion, the camera glides through the cityscape, keeping roads, intersections, and buildings in a clear and continuous frame.}}
        \label{fig:style_transfer_1}
    \end{subfigure}

    \vspace{0.4em}

    \begin{subfigure}{\textwidth}
        \centering
        \begin{tikzpicture}
            \node[anchor=south west, inner sep=0] (img) at (0,0)
                {\includegraphics[width=0.93\textwidth]{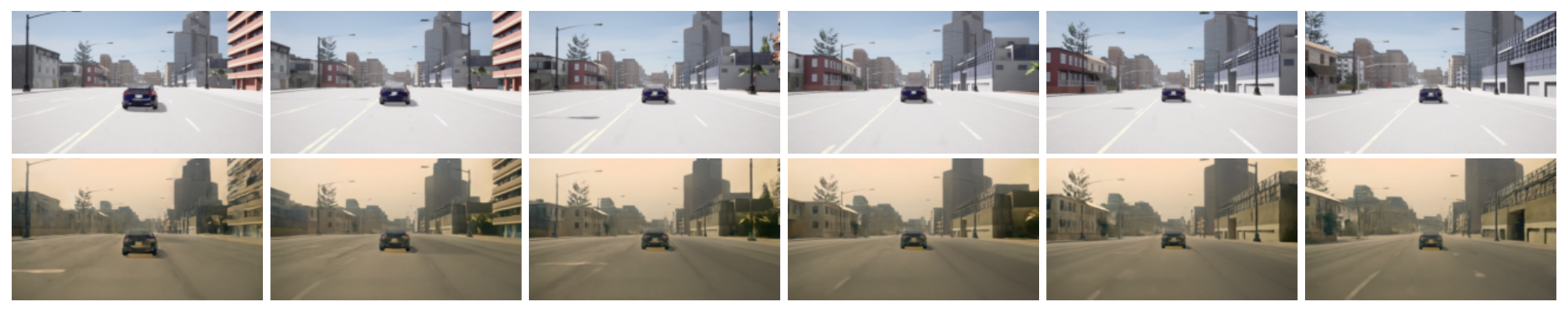}};
            \node[anchor=center, font=\fontsize{5}{6}\selectfont, align=center, rotate=90] at (-0.25, 1.90) {Ground Truth};
            \node[anchor=center, font=\fontsize{5}{6}\selectfont, align=center, rotate=90] at (-0.25, 0.675) {Style Transfer};
        \end{tikzpicture}
        \caption{\emph{A first-person driving simulator scene in the sparkling light of a clear spring day with a gentle breeze, with an assortment of brightly colored cars in every direction. Maintaining a steady forward motion, the camera glides through the cityscape, keeping roads, intersections, and buildings in a clear and continuous frame.}}
        \label{fig:style_transfer_2}
    \end{subfigure}

    \vspace{0.4em}

    \begin{subfigure}{\textwidth}
        \centering
        \begin{tikzpicture}
            \node[anchor=south west, inner sep=0] (img) at (0,0)
                {\includegraphics[width=0.93\textwidth]{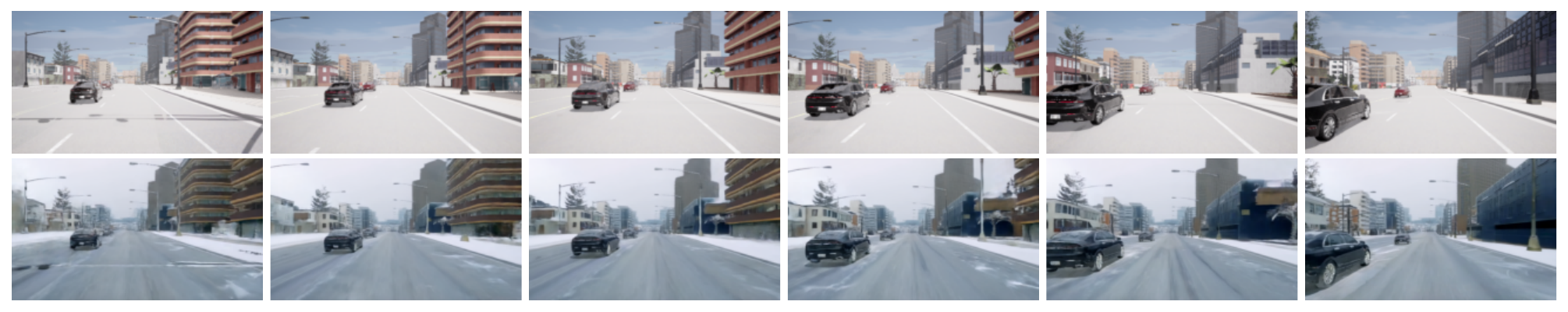}};
            \node[anchor=center, font=\fontsize{5}{6}\selectfont, align=center, rotate=90] at (-0.25, 1.90) {Ground Truth};
            \node[anchor=center, font=\fontsize{5}{6}\selectfont, align=center, rotate=90] at (-0.25, 0.675) {Style Transfer};
        \end{tikzpicture}
        \caption{\emph{A first-person driving simulator scene on a snowy winter day with white-dusted streets and rooftops, alongside dark charcoal and deep navy luxury vehicles. Maintaining a steady forward motion, the camera glides through the cityscape, keeping roads, intersections, and buildings in a clear and continuous frame.}}
        \label{fig:style_transfer_3}
    \end{subfigure}

    \vspace{0.4em}

    \begin{subfigure}{\textwidth}
        \centering
        \begin{tikzpicture}
            \node[anchor=south west, inner sep=0] (img) at (0,0)
                {\includegraphics[width=0.93\textwidth]{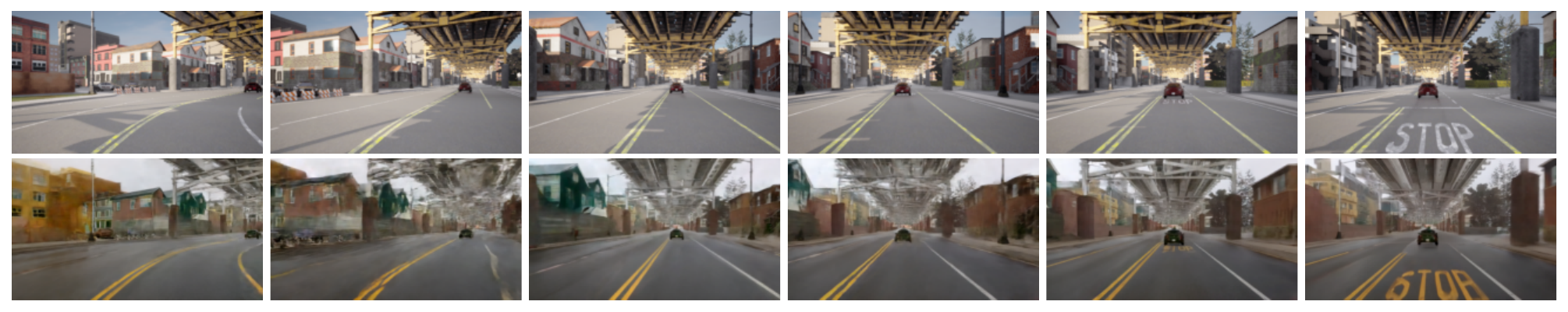}};
            \node[anchor=center, font=\fontsize{5}{6}\selectfont, align=center, rotate=90] at (-0.25, 1.90) {Ground Truth};
            \node[anchor=center, font=\fontsize{5}{6}\selectfont, align=center, rotate=90] at (-0.25, 0.675) {Style Transfer};
        \end{tikzpicture}
        \caption{\emph{A first-person driving simulator scene as a gentle rain mists the air beneath pale overcast skies, with an assortment of brightly colored cars in every direction. The perspective advances smoothly through urban intersections and streets, holding a stable forward-facing view that captures the road and the surrounding architecture.}}
        \label{fig:style_transfer_4}
    \end{subfigure}

    \caption{Style-transfer examples using text prompts and edge conditioning with Cosmos-Transfer 2.5-2B. In each subfigure, the top row shows the ground-truth driving sequence and the bottom row shows the corresponding style-transferred output.}
    \label{fig:style_transfer}
\end{figure*}
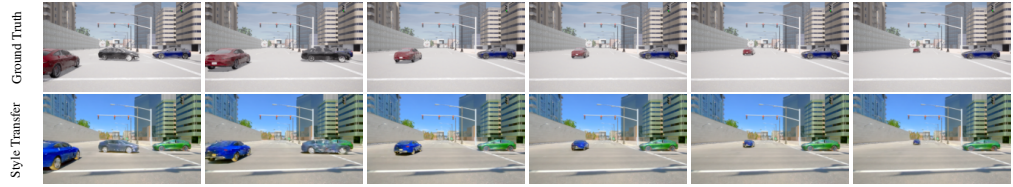
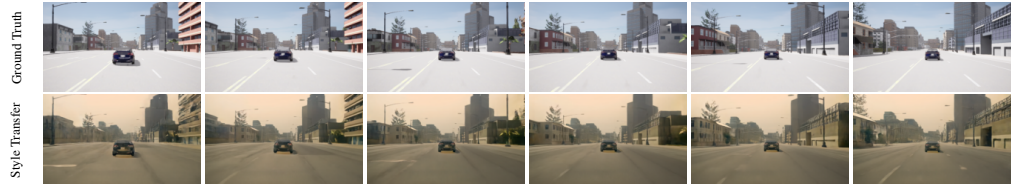
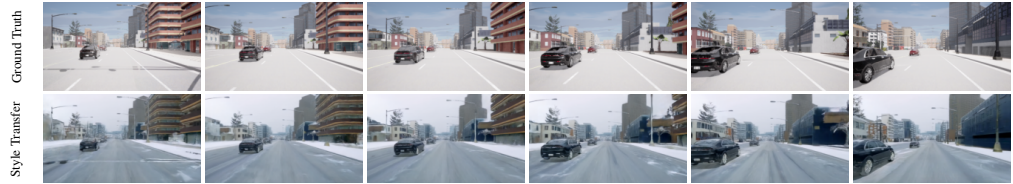
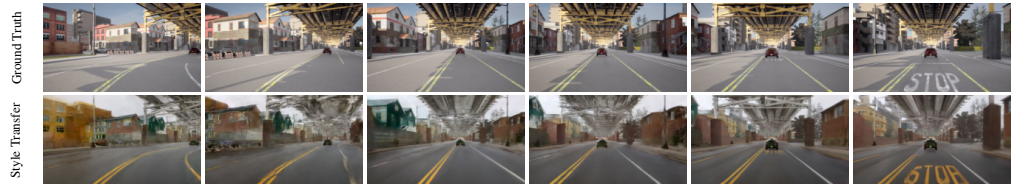

\FloatBarrier
\subsection{Additional Mirror Video Qualitative Results}
\label{sec:appendix:addl-mvm-qual}

In this final subsection, we provide additional qualitative mirror video results for completeness.

\FloatBarrier
\subsubsection{CARLA Seen-Town Test Set}

We then show qualitative results on Test One, a seen-town benchmark built from scenarios and spatial regions that are separated from or underrepresented in the main CARLA training distribution. These examples complement the held-out-town results by isolating generalization to new situations within otherwise familiar towns.

\begin{figure*}[th]
    \centering

    \begin{subfigure}{\textwidth}
        \centering
        \begin{tikzpicture}
            \node[anchor=south west, inner sep=0] (img) at (0,0)
                {\includegraphics[width=0.93\textwidth]{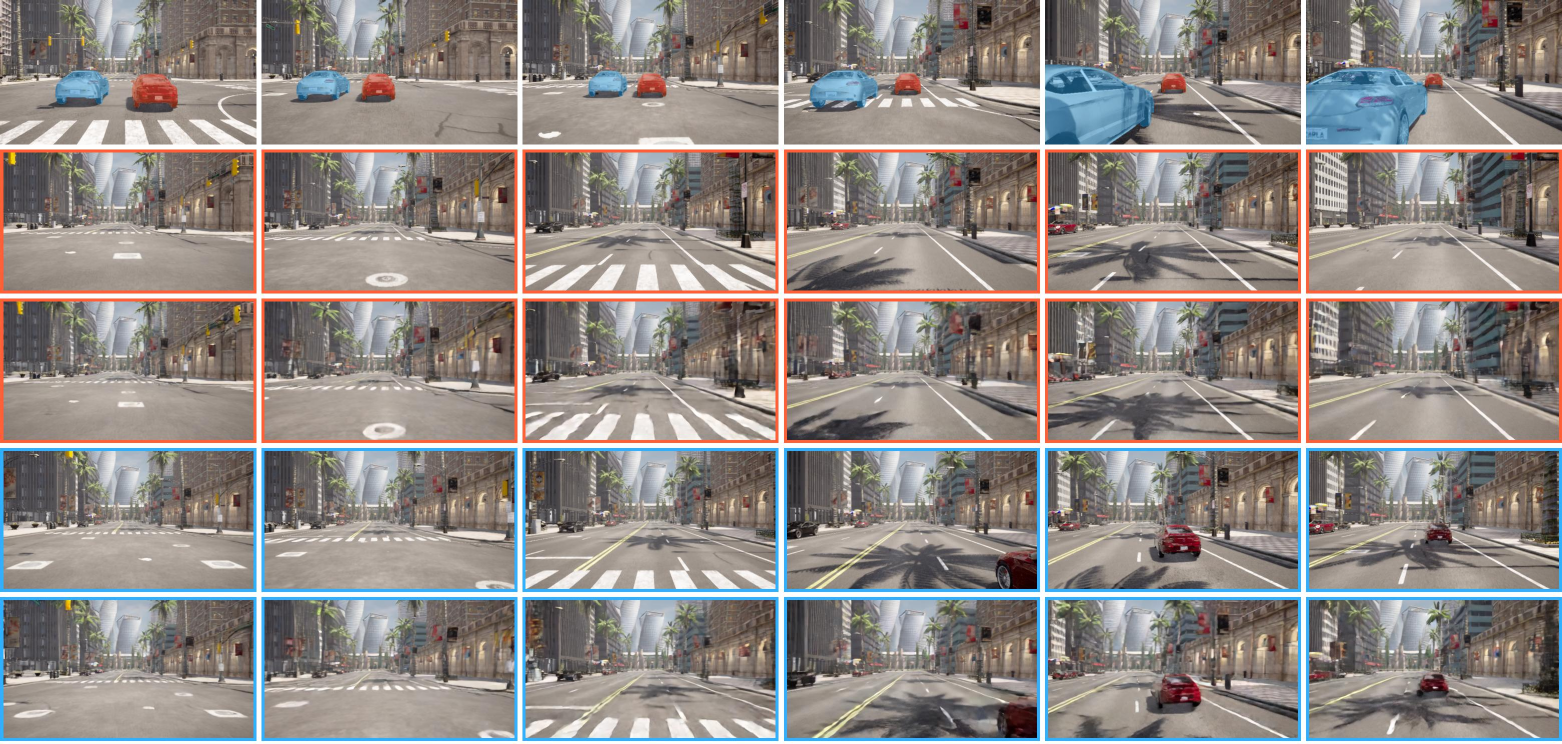}};
            \node[anchor=center, font=\fontsize{5}{6}\selectfont, align=center, rotate=90] at (-0.25, 5.6) {Observation\\(masks overlaid)};
            \node[anchor=center, font=\fontsize{5}{6}\selectfont, align=center, rotate=90] at (-0.25, 4.35) {Ground Truth};
            \node[anchor=center, font=\fontsize{5}{6}\selectfont, align=center, rotate=90] at (-0.25, 3.15) {Mirrored};
            \node[anchor=center, font=\fontsize{5}{6}\selectfont, align=center, rotate=90] at (-0.25, 1.90) {Ground Truth};
            \node[anchor=center, font=\fontsize{5}{6}\selectfont, align=center, rotate=90] at (-0.25, 0.65) {Mirrored};
        \end{tikzpicture}
        \caption{}
        \label{fig:annotated_grid_town10}
    \end{subfigure}

    \vspace{0.5em}

    \begin{subfigure}{\textwidth}
        \centering
        \begin{tikzpicture}
            \node[anchor=south west, inner sep=0] (img) at (0,0)
                {\includegraphics[width=0.93\textwidth]{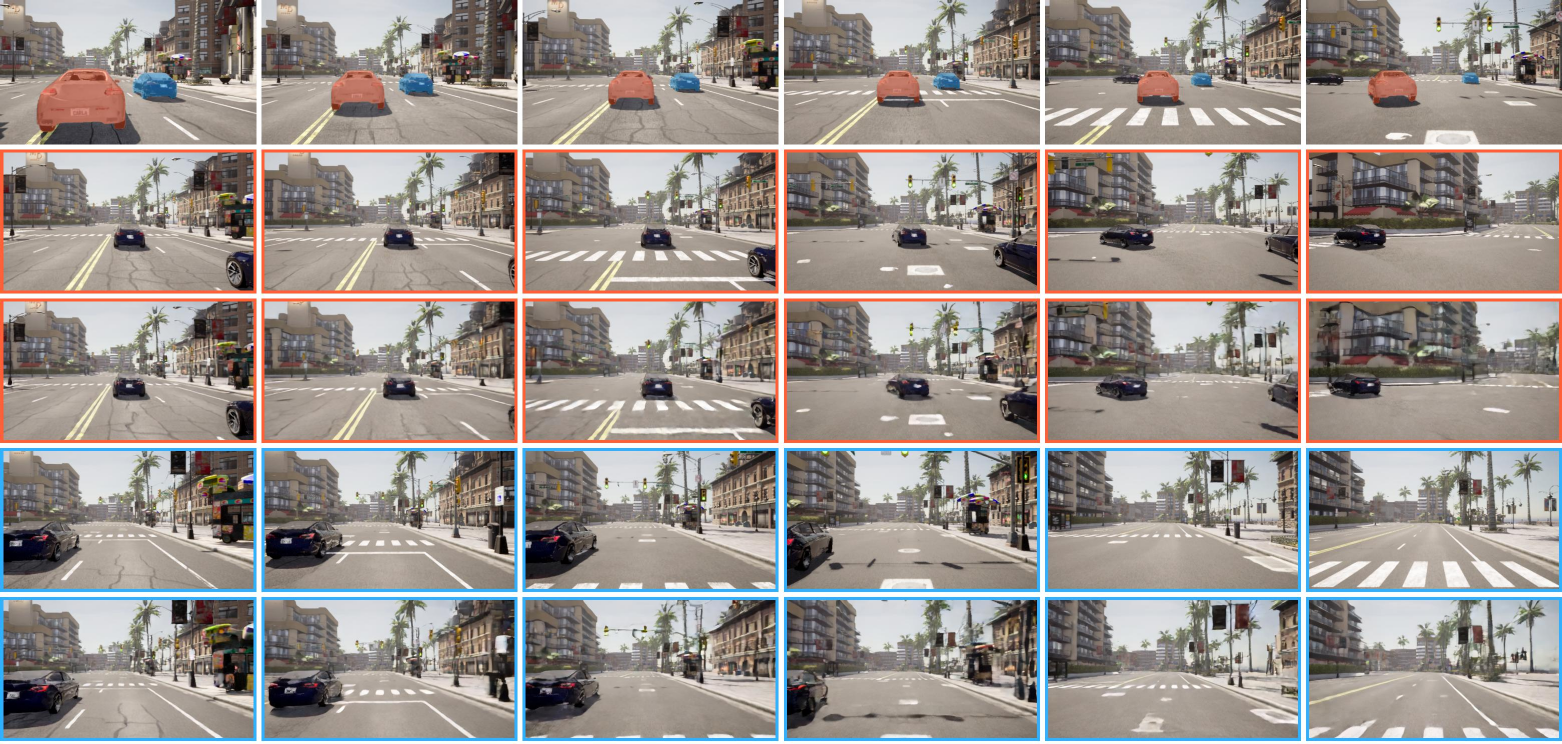}};
            \node[anchor=center, font=\fontsize{5}{6}\selectfont, align=center, rotate=90] at (-0.25, 5.6) {Observation\\(masks overlaid)};
            \node[anchor=center, font=\fontsize{5}{6}\selectfont, align=center, rotate=90] at (-0.25, 4.35) {Ground Truth};
            \node[anchor=center, font=\fontsize{5}{6}\selectfont, align=center, rotate=90] at (-0.25, 3.15) {Mirrored};
            \node[anchor=center, font=\fontsize{5}{6}\selectfont, align=center, rotate=90] at (-0.25, 1.90) {Ground Truth};
            \node[anchor=center, font=\fontsize{5}{6}\selectfont, align=center, rotate=90] at (-0.25, 0.65) {Mirrored};
        \end{tikzpicture}
        \caption{}
        \label{fig:annotated_grid_additional}
    \end{subfigure}

    \caption{Demonstrator-specific view transformation in CARLA. In each example, MVM is conditioned on the same learner observation but different demonstrator masks, shown in orange and blue. For each selected demonstrator, we show the ground-truth demonstrator egocentric video and the corresponding mirrored sample generated by our method. Because only the demonstrator mask changes between the orange and blue samples, these results demonstrate that MVM generates target-specific egocentric observations.}
    \label{fig:annotated_grid_combined}
\end{figure*}
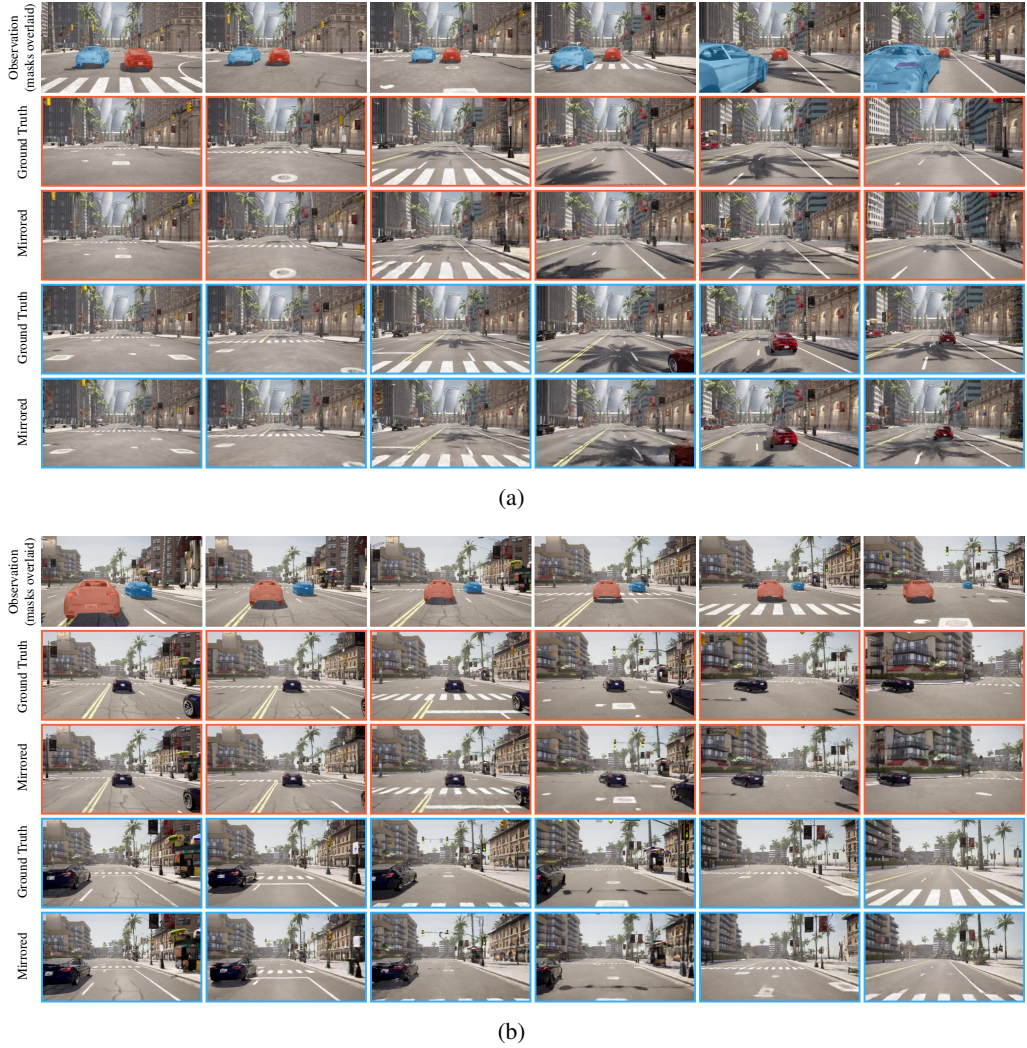

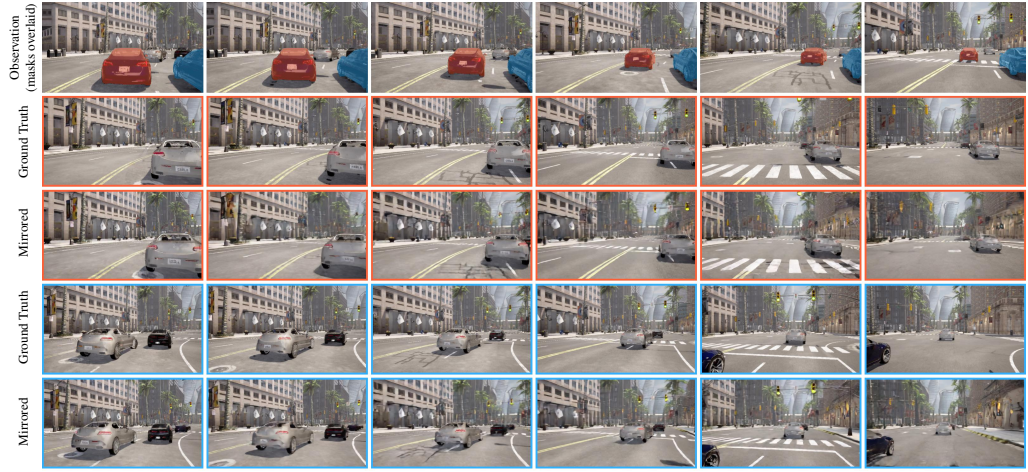
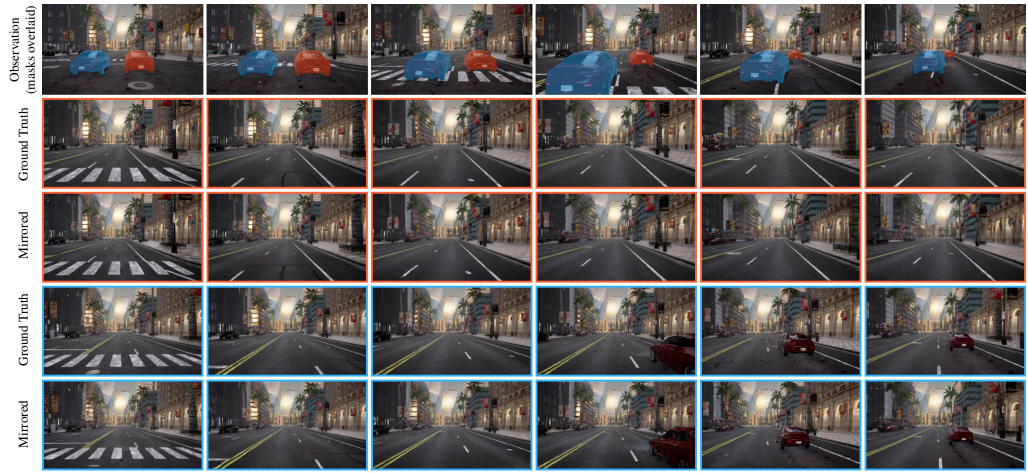
\begin{figure*}[th]
    \centering

    \begin{subfigure}{\textwidth}
        \centering
        \begin{tikzpicture}
            \node[anchor=south west, inner sep=0] (img) at (0,0)
                {\includegraphics[width=0.93\textwidth]{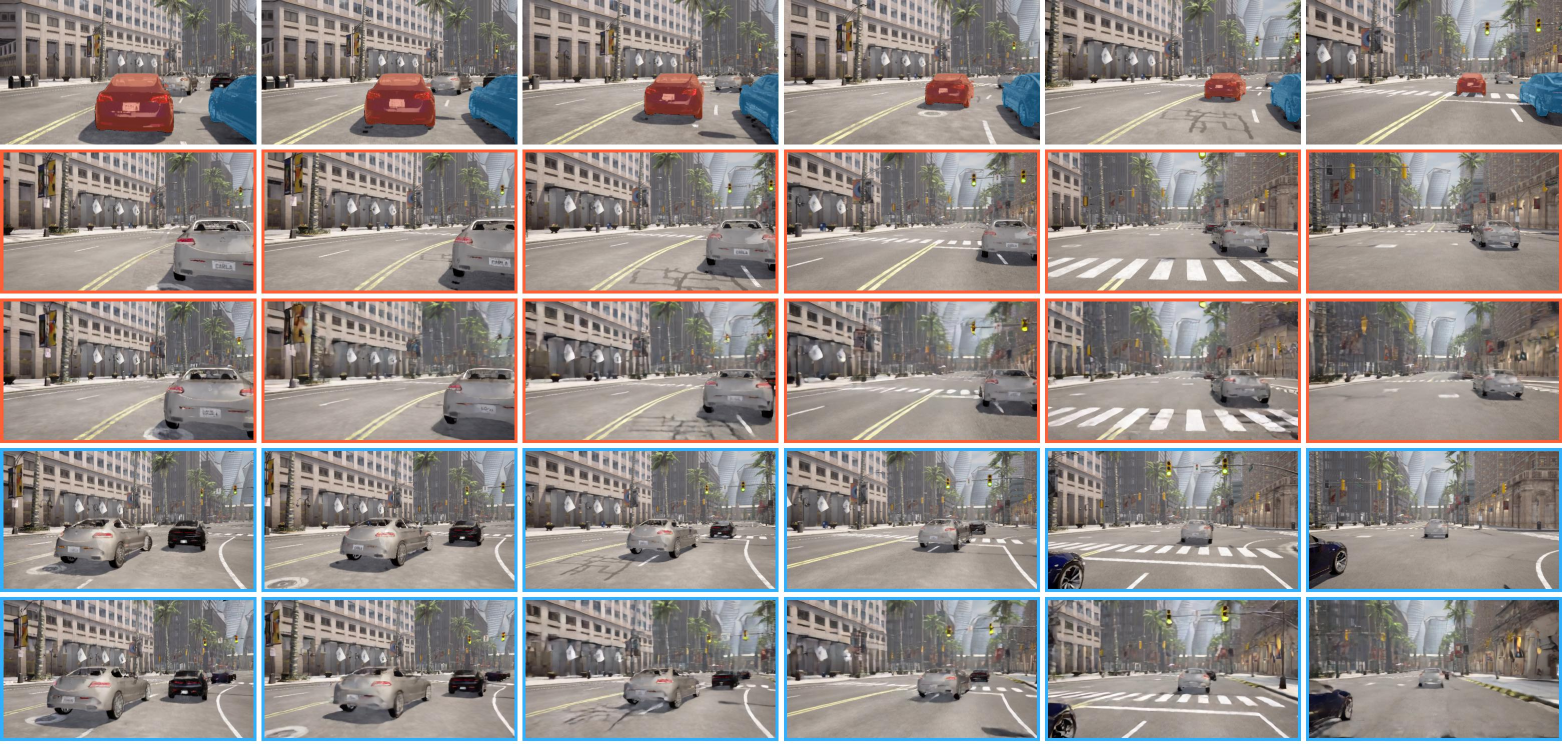}};
            \node[anchor=center, font=\fontsize{5}{6}\selectfont, align=center, rotate=90] at (-0.25, 5.6) {Observation\\(masks overlaid)};
            \node[anchor=center, font=\fontsize{5}{6}\selectfont, align=center, rotate=90] at (-0.25, 4.35) {Ground Truth};
            \node[anchor=center, font=\fontsize{5}{6}\selectfont, align=center, rotate=90] at (-0.25, 3.15) {Mirrored};
            \node[anchor=center, font=\fontsize{5}{6}\selectfont, align=center, rotate=90] at (-0.25, 1.90) {Ground Truth};
            \node[anchor=center, font=\fontsize{5}{6}\selectfont, align=center, rotate=90] at (-0.25, 0.65) {Mirrored};
        \end{tikzpicture}
        \caption{}
        \label{fig:annotated_grid_additional_003}
    \end{subfigure}

    \vspace{0.5em}

    \begin{subfigure}{\textwidth}
        \centering
        \begin{tikzpicture}
            \node[anchor=south west, inner sep=0] (img) at (0,0)
                {\includegraphics[width=0.93\textwidth]{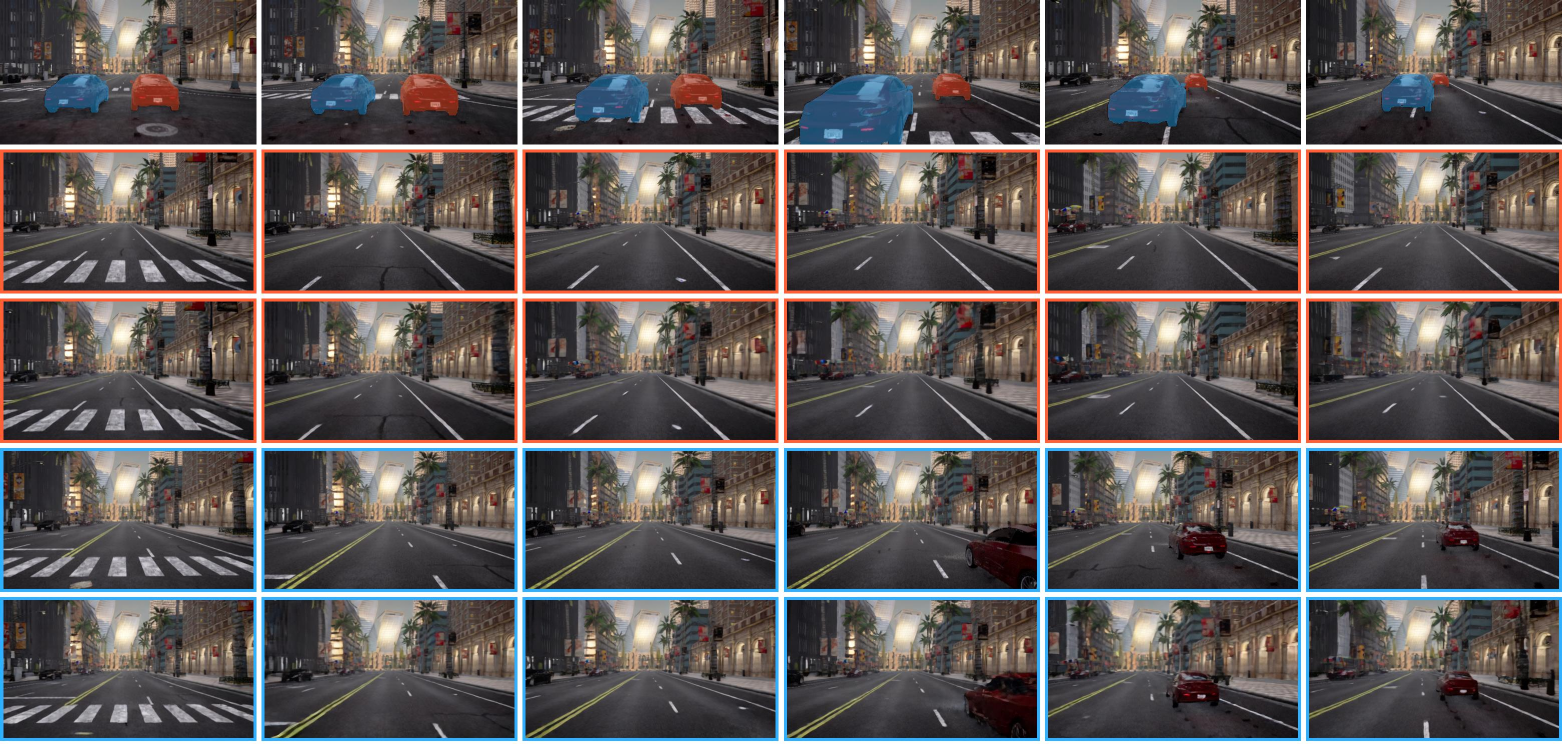}};
            \node[anchor=center, font=\fontsize{5}{6}\selectfont, align=center, rotate=90] at (-0.25, 5.6) {Observation\\(masks overlaid)};
            \node[anchor=center, font=\fontsize{5}{6}\selectfont, align=center, rotate=90] at (-0.25, 4.35) {Ground Truth};
            \node[anchor=center, font=\fontsize{5}{6}\selectfont, align=center, rotate=90] at (-0.25, 3.15) {Mirrored};
            \node[anchor=center, font=\fontsize{5}{6}\selectfont, align=center, rotate=90] at (-0.25, 1.90) {Ground Truth};
            \node[anchor=center, font=\fontsize{5}{6}\selectfont, align=center, rotate=90] at (-0.25, 0.65) {Mirrored};
        \end{tikzpicture}
        \caption{}
        \label{fig:annotated_grid_additional_004}
    \end{subfigure}

    \caption{Additional demonstrator-specific view transformation results in CARLA Test locations Training town.}
    \label{fig:annotated_grid_additional_combined}
\end{figure*}

\clearpage
\FloatBarrier
\subsection{Minecraft Training-Set Examples.}

\begin{figure*}[th]
    \centering
    \captionsetup[subfigure]{font=small, justification=centering}

    \begin{subfigure}{\textwidth}
        \centering
        \begin{tikzpicture}
            \node[anchor=south west, inner sep=0] (img) at (0,0)
                {\includegraphics[width=0.93\textwidth]{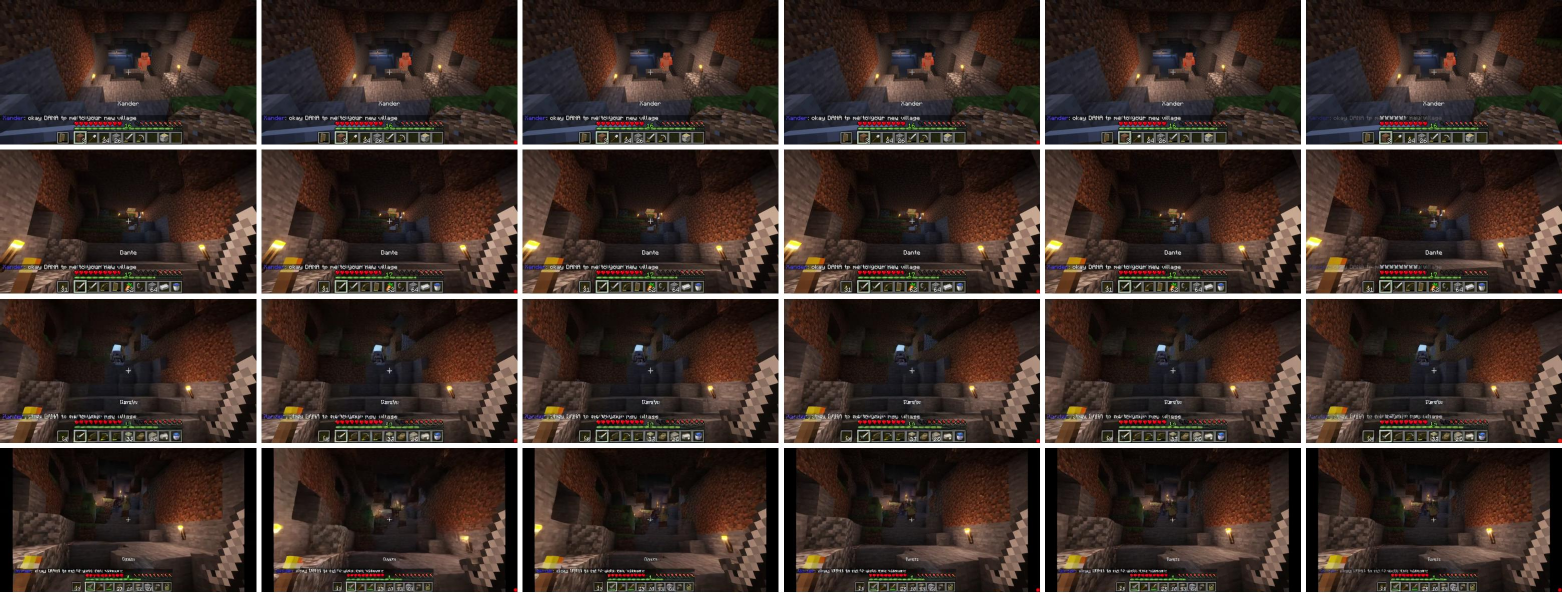}};
            \node[anchor=center, font=\fontsize{5}{6}\selectfont, align=center, rotate=90] at (-0.25, 4.35) {Observation\\(masks overlaid)};
            \node[anchor=center, font=\fontsize{5}{6}\selectfont, align=center, rotate=90] at (-0.25, 3.10) {Ground Truth};
            \node[anchor=center, font=\fontsize{5}{6}\selectfont, align=center, rotate=90] at (-0.25, 1.90) {Mirrored 1};
            \node[anchor=center, font=\fontsize{5}{6}\selectfont, align=center, rotate=90] at (-0.25, 0.65) {Mirrored 2};
        \end{tikzpicture}
        \caption{Mirror video model's generates a cave's texture with plausible torch placements.}
    \end{subfigure}

    \vspace{0.4em}

    \begin{subfigure}{\textwidth}
        \centering
        \begin{tikzpicture}
            \node[anchor=south west, inner sep=0] (img) at (0,0)
                {\includegraphics[width=0.93\textwidth]{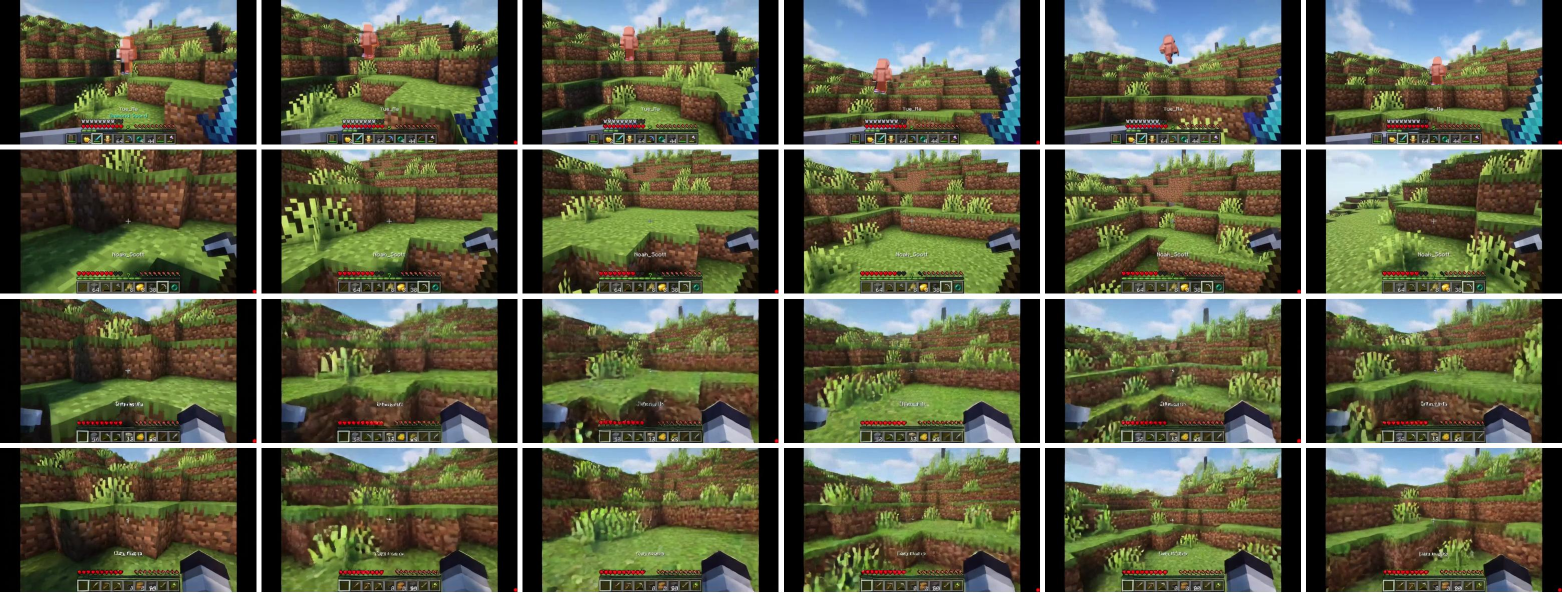}};
            \node[anchor=center, font=\fontsize{5}{6}\selectfont, align=center, rotate=90] at (-0.25, 4.35) {Observation\\(masks overlaid)};
            \node[anchor=center, font=\fontsize{5}{6}\selectfont, align=center, rotate=90] at (-0.25, 3.10) {Ground Truth};
            \node[anchor=center, font=\fontsize{5}{6}\selectfont, align=center, rotate=90] at (-0.25, 1.90) {Mirrored 1};
            \node[anchor=center, font=\fontsize{5}{6}\selectfont, align=center, rotate=90] at (-0.25, 0.65) {Mirrored 2};
        \end{tikzpicture}
        \caption{Mirror video model effectively captures the player jump motion.}
    \end{subfigure}

    \vspace{0.4em}

    \begin{subfigure}{\textwidth}
        \centering
        \begin{tikzpicture}
            \node[anchor=south west, inner sep=0] (img) at (0,0)
                {\includegraphics[width=0.93\textwidth]{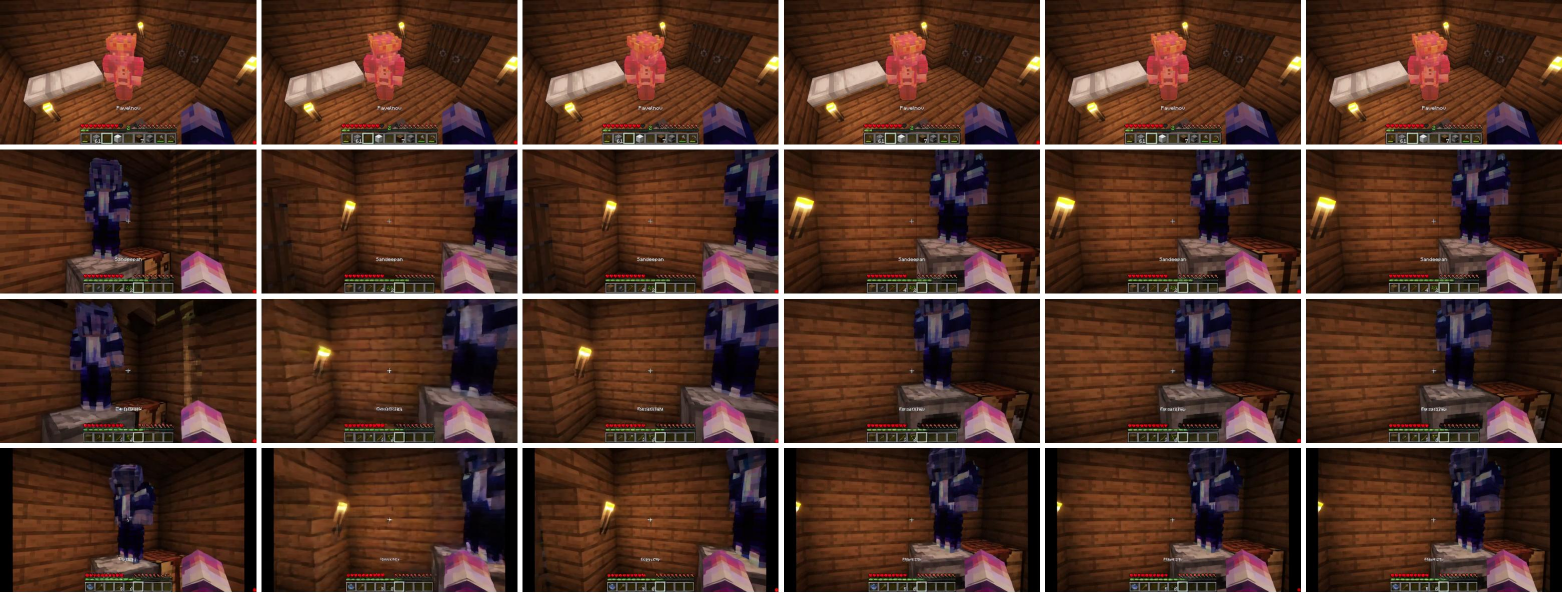}};
            \node[anchor=center, font=\fontsize{5}{6}\selectfont, align=center, rotate=90] at (-0.25, 4.35) {Observation\\(masks overlaid)};
            \node[anchor=center, font=\fontsize{5}{6}\selectfont, align=center, rotate=90] at (-0.25, 3.10) {Ground Truth};
            \node[anchor=center, font=\fontsize{5}{6}\selectfont, align=center, rotate=90] at (-0.25, 1.90) {Mirrored 1};
            \node[anchor=center, font=\fontsize{5}{6}\selectfont, align=center, rotate=90] at (-0.25, 0.65) {Mirrored 2};
        \end{tikzpicture}
        \caption{Mirror video model faithfully renders observed player's camera movement.}
    \end{subfigure}

    \caption{Training set qualitative mirroring results for a scenario the mirror network has been trained on. We select a target agent using orange instance masks. For the target, we show the ground-truth egocentric video and two sampled predictions from the perspective transfer network, demonstrating egocentric generation from a shared observation.}
    \label{fig:plaicraft_train1a}
\end{figure*}
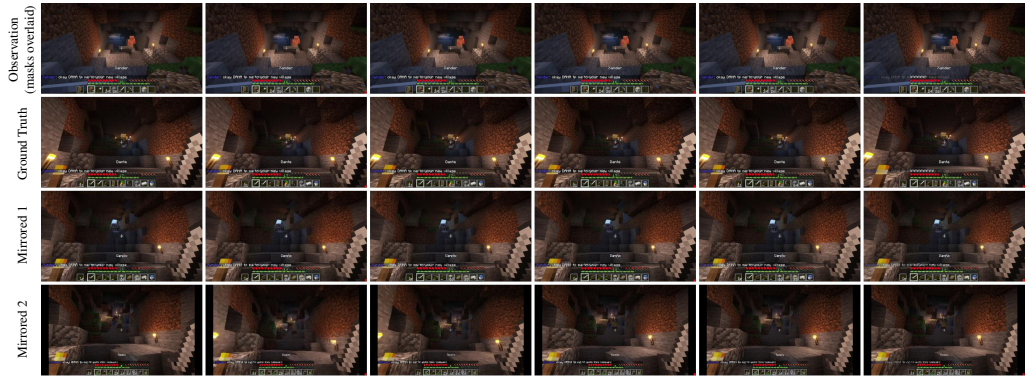
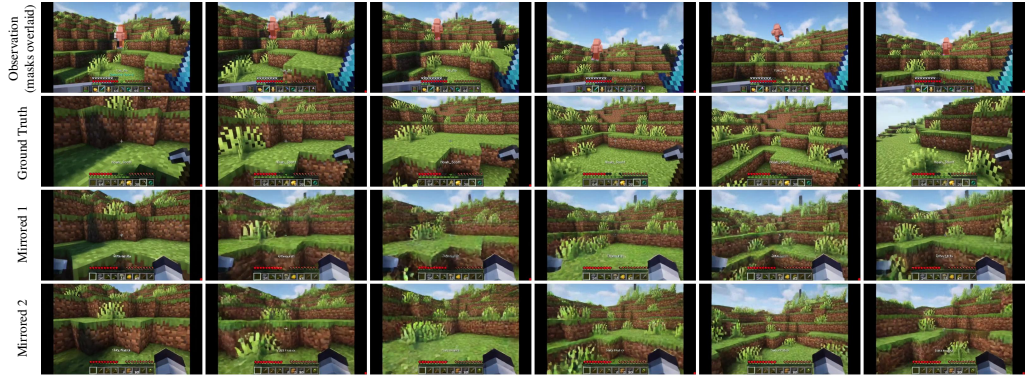
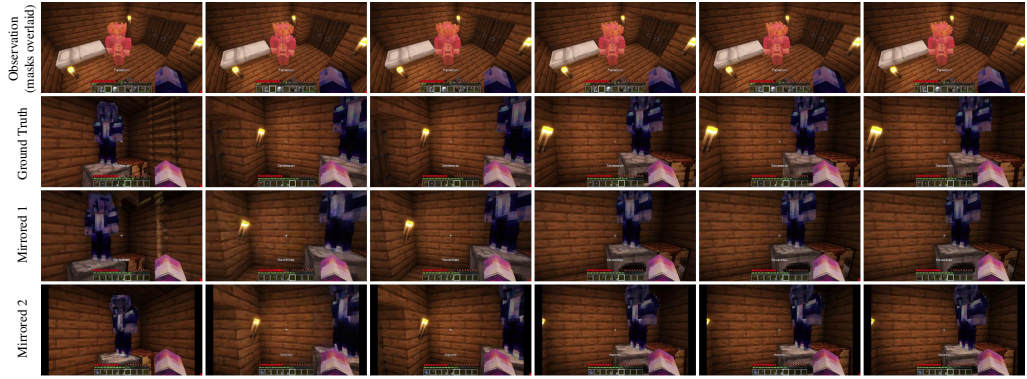

\clearpage

\subsection{Social Impact}

Our work aims to reduce the cost of training embodied policies by reusing observations from a learner in an environment where multiple demonstrators exist and by releasing paired simulation data that can support future research. This could broaden access to mirror-video modeling and data-efficient policy learning for academic groups. At the same time, methods that improve autonomous decision making can create downstream safety and misuse risks if deployed without careful validation. We therefore view our approach as a research tool that should be paired with rigorous testing, uncertainty-aware evaluation.

\newpage

\end{document}